\documentclass{article}

\usepackage{adjustbox}
\usepackage{longtable}

\usepackage{arxiv}

\usepackage{bm}
\usepackage{amssymb,amsthm,mathtools}

\usepackage[utf8]{inputenc} % allow utf-8 input
\usepackage[T1]{fontenc}    % use 8-bit T1 fonts
\usepackage{url}            % simple URL typesetting
\usepackage{booktabs}       % professional-quality tables
\usepackage{amsfonts}       % blackboard math symbols
\usepackage{nicefrac}       % compact symbols for 1/2, etc.
\usepackage{microtype}      % microtypography
\usepackage{multirow}
\usepackage{xcolor}
\usepackage{enumitem}
\usepackage{graphicx}
\usepackage{placeins}
\usepackage{orcidlink}
\usepackage{comment}
\usepackage[most]{tcolorbox}
\usepackage{array}
\newcolumntype{R}[1]{>{\raggedleft\arraybackslash}p{#1}}
\newcolumntype{L}[1]{>{\raggedright\arraybackslash}p{#1}}

\usepackage[numbers]{natbib}
\usepackage{caption}
\usepackage{amsmath}
\usepackage{rotating}
\usepackage[capitalise,noabbrev]{cleveref}

\makeatletter
\renewcommand{\footnotesize}{\@setfontsize\footnotesize{8pt}{10pt}}
\makeatother

\makeatletter
\renewcommand{\scriptsize}{\@setfontsize\scriptsize{7pt}{9pt}}
\makeatother

\definecolor{VUB_blauw}{rgb}{0.1529, 0.2667, 0.5529}
\usepackage{hyperref}       % hyperlinks
\hypersetup{
    colorlinks,%
    citecolor=VUB_blauw,%
    filecolor=VUB_blauw,%
    linkcolor=VUB_blauw,%
    urlcolor=VUB_blauw,%
    pdftitle={Shorter Reasoning, Earlier Answers? An Evaluation of Reasoning Interfaces},%
    pdfauthor={Francesca Carlon, Vincent Ginis, and Andres Algaba}
}

\title{Shorter Reasoning, Earlier Answers? \\ An Evaluation of Reasoning Interfaces}
\runningtitle{Shorter Reasoning, Earlier Answers? An Evaluation of Reasoning Interfaces}
 
\author{
  Francesca Carlon\textsuperscript{1,2,*} \\
  \orcidlinkc{0009-0004-2152-2745} \\ 
  \And
  Vincent Ginis\textsuperscript{1,2,3} \\
  \orcidlinkc{0000-0003-0063-9608} \\
  \And
  Andres Algaba\textsuperscript{1,2} \\
  \orcidlinkc{0000-0002-0532-3066} \\
  \and
  \textsuperscript{1}Data Analytics Lab, Vrije Universiteit Brussel, Pleinlaan 5, 1050 Brussels, Belgium \\
  \textsuperscript{2}imec-SMIT, Vrije Universiteit Brussel, Pleinlaan 9, 1050 Brussels, Belgium \\
  \textsuperscript{3}School of Engineering and Applied Sciences, Harvard University, Cambridge, Massachusetts 02138, USA
}

\begin{document}
\pagenumbering{arabic}

\maketitle
\renewcommand{\thefootnote}{}
\footnotetext{*Corresponding author: \href{mailto:francesca.carlon@vub.be}{francesca.carlon@vub.be} \\}
\renewcommand{\thefootnote}{\arabic{footnote}}
\thispagestyle{plain}

\begin{abstract}
Large language models often reason at length before answering, increasing cost and latency. Prompts and trained settings can shorten this reasoning, but a shorter trace may only show that the model stopped sooner. Here, we evaluate paired runs of the same question at matched reasoning horizons across $198$ GPQA Diamond and $500$ MMLU-Pro questions. We test a numeric/concision prompt that announces a token limit for Qwen3-14B and the trained effort settings of gpt-oss-20b and -120b. The Qwen prompt shortens reasoning traces by $12$--$17\%$, while accuracy changes at matched token limits are small and mixed. A concise/early-answer instruction raises MMLU-Pro accuracy by $3.8$ percentage points at $512$ tokens, including $+2.7$ points when both runs are unfinished. Its gain at $2{,}048$ tokens is uncertain. For gpt-oss, candidate-logit answers from completed low- and medium-effort reasoning are $14.5$--$26.3$ points more accurate than matched-horizon high-effort answers. Most of the $512$-token advantage comes from lower effort finishing earlier, while differences among unfinished runs are smaller and mixed. Wrong early answers often concentrate probability on the chosen option, so earlier stopping does not uniformly improve probability quality. In these tests, a tight deadline can favor lower effort or a concise instruction, whereas allowing high effort to finish can recover higher final accuracy. Evaluations should report correct completion before a deadline, the answer obtained when a run is stopped, differences among unfinished runs, and probability assigned to the correct answer separately.
\end{abstract}


\section{Introduction}

Large language models often solve hard questions by writing intermediate reasoning before giving a final answer, a behavior that began as chain-of-thought prompting and is now trained into models whose reasoning length or effort can be adjusted by the user~\citep{wei2022cot,kojima2022zeroshotcot,jaech2024openai,guo2025deepseek}. Accuracy on competition mathematics, graduate science, and code often grows as models spend more reasoning tokens during generation, sampling, or aggregation~\citep{wang2022selfconsistency,snell2024scaling,muennighoff2025s1simpletesttimescaling,aggarwal2026reasoning}. This reasoning is costly and variable. A single answer can consume tens of thousands of tokens, and additional tokens often add little once the answer has effectively been chosen~\citep{chen2024not,ballon2025relationship,su2025between,sun2025stop}. Controlling that cost is therefore a practical constraint when models operate under deadlines~\citep{dean1988timeplanning,zilberstein1996anytime,russell1991dotherightthing}.

We use the term \emph{reasoning control} for an intervention that can change how much intermediate reasoning a model produces before its final answer. Examples include training that favors shorter reasoning~\citep{aggarwal2025l1,han2025token,arora2025training}, settings such as gpt-oss reasoning effort~\citep{agarwal2025gpt}, methods that steer generation~\citep{li2025steering,sun2025strict,wen2025budgetthinker}, and prompts that request concise reasoning~\citep{renze2024concise,nayab2024concise,xu2025chain}. These controls reach the model through different \emph{reasoning interfaces}: instructions written into the prompt or settings trained into the model. Most studies evaluate these interfaces by plotting final accuracy against the length of the completed trace~\citep{sui2025stop}. That comparison measures the final cost--accuracy tradeoff, but it cannot show whether a control improves the answer after a fixed amount of reasoning or simply makes the model stop sooner. More broadly, reported benchmark scores depend on how much test-time compute an evaluation allows, and recent work calls for protocols that make this allocation explicit~\citep{mcfadyen2026inference}.

A model that stops early has a completed answer at a point where another model may still be working. Comparing those states can answer a useful deployment question, but it does not show that the unfinished reasoning itself is better. Prior work uses stopped trajectories and forced answers to study answer stabilization and early stopping~\citep{ballon2026probing,datta2026decide,wu2025interruptible}. We build on that work by comparing paired runs at matched reasoning horizons and by recording separately whether each run has already finished. At each stopping point, we measure whether the model has finished correctly, what answer an unfinished run gives when stopped, how the runs compare while both are still reasoning, and how much probability the model places on the correct answer. For Qwen3-14B, we compare ordinary generation with generation under the numeric/concision prompt at the same externally imposed point $B$ (\Cref{fig:design}). We separately test a concise/early-answer instruction that asks the model to reason concisely and prioritize an early usable answer. For gpt-oss, we compare a completed low- or medium-effort run with high effort at the point where the lower-effort run chose to stop. We validate the forced-answer measurement against completed answers, controls using the same questions, a scripted positive control, and generated-answer continuations.

The main paired experiments cover Qwen3-14B and gpt-oss-20b/-120b on GPQA Diamond~\citep{rein2024gpqa} and a $500$-item MMLU-Pro subset sampled across subject areas~\citep{wang2024mmlupro}. Qwen3-4B/-8B calibrate the measurement, and Omni-MATH-2~\citep{gao2024omni,ballon2026benchmarks} extends the evaluation to open-ended mathematics. In the tested Qwen3 case, the numeric/concision prompt shortens traces without a consistent same-horizon gain. A nine-arm MMLU-Pro experiment finds an early gain from the concise/early-answer instruction but no clear benefit from matching the stated number to the stopping point. In the tested gpt-oss case, lower effort gives a correct answer earlier mainly because it stops and commits sooner. These within-family case studies separate completion timing from the answer obtained when a run is stopped. They do not provide a causal comparison between prompts and trained effort settings.

\section{Matched-horizon probes separate evaluation points from generation policies}

\begin{figure}[!t]
\centering
\includegraphics[width=\textwidth]{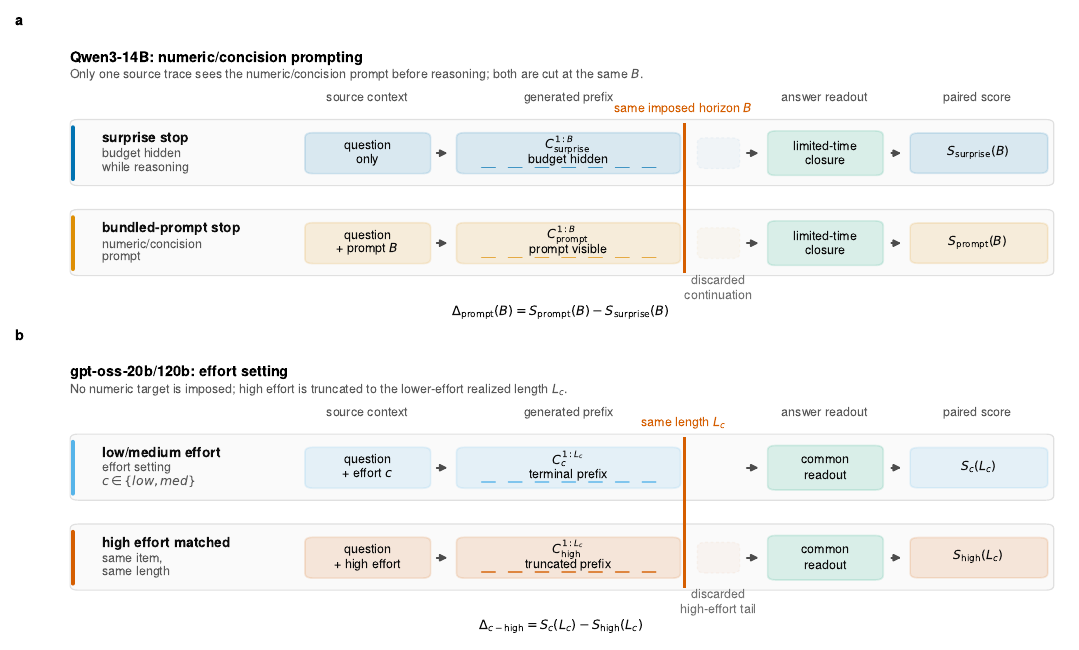}
\caption{\textbf{We compare paired runs at matched reasoning horizons.} \textbf{a}, For Qwen3-14B, the terminal-aware policy contrast evaluates two traces from the same item and generation replicate at the same imposed token horizon $B$: an ordinary trace with the horizon hidden (``surprise stop'') and a trace generated under the textual numeric/concision treatment (``bundled-prompt stop''). Truncated prefixes are read in their policy context, while a trace that finishes before $B$ keeps its natural terminal answer. Common-context and both-active diagnostics hold the answer-time prompt fixed (\Cref{tab:main-estimands}). \textbf{b}, For gpt-oss-20b and gpt-oss-120b, the effort contrast compares a completed low- or medium-effort reasoning prefix with high-effort reasoning at that same itemwise lower-effort length $L_c$ ($c\in\{\mathrm{low},\mathrm{medium}\}$), under a common medium-effort answer context. If high effort finishes before $L_c$, its completed reasoning block is replayed; the high-active subset is diagnostic. Replayed reasoning prefixes exclude the original final-answer tokens.}
\label{fig:design}
\end{figure}

Let $i$ index items, $r$ index independent generation replicates, $z$ index the generation condition, and $L_{ir}^{z}$ denote the condition's terminal reasoning length. We write $Y_{ir}^{z,\rho}(B)\in\{0,1\}$ for correctness when condition $z$ is evaluated at horizon $B$ with readout $\rho$. The standardized prefix-probe readout ($\rho=P$) closes the stopped reasoning block and selects among the candidate answers. The natural-terminal readout ($\rho=N$) uses the answer given when the model finished without interruption. The generated-answer continuation ($\rho=C$) closes an externally stopped reasoning block and greedily generates an unconstrained final-channel answer. These readouts measure which answer is available, not when the model decided. \Cref{fig:design,tab:main-estimands} specify where each readout enters the two case studies.

Token-exact replay and candidate-logit scoring describe different stages of the measurement. Token-exact replay constructs the stopped input directly from stored prompt, reasoning, and closure token IDs, without decoding and retokenizing the reasoning prefix. The candidate-logit probe then scores every valid one-token spelling of each answer letter from full-vocabulary logits and aggregates spellings by letter. We report the total probability mass assigned to valid options before normalization, conditional log loss, multiclass Brier score, and argmax accuracy. These proper scores and valid-option mass diagnose the standardized probe; they do not make it a natural continuation.

We distinguish four readouts. The \emph{policy-context} readout preserves each condition's generation prompt at answer time and measures the full interface-level contrast. The \emph{common-context} readout uses the ordinary base prompt for every Qwen arm or the medium-effort answer context for every gpt-oss arm, replays the stored reasoning IDs, and applies a common closure. Differences at answer time are then carried by the generated prefixes. The \emph{both-active} readout applies the common context only to pairs in which both traces remain unfinished at $B$; it is diagnostic because conditioning on survival changes the compared population. The \emph{behavioral continuation} greedily generates a short final answer from a stopped prefix instead of selecting an option from first-position logits. Completed reasoning blocks are replayed under the common context when estimating prefix-carried differences, whereas the terminal-aware policy readout retains their natural answers. Full closure texts and prompt variants appear in \Cref{tab:prompt-texts,tab:main-estimands,sec:methods}.

For Qwen3-14B, let $h$ denote ordinary generation with the horizon hidden (the \emph{surprise stop}) and $v$ the textual numeric/concision treatment with the horizon visible (the \emph{bundled prompt}). The terminal-aware policy-context readout uses $Y^{z,T_{\mathrm{pol}}}_{ir}(B)=Y^{z,N}_{ir}(L_{ir}^{z})$ when the trace finishes before $B$, and $Y^{z,P_{\mathrm{pol}}}_{ir}(B)$ otherwise. The primary curve is
\[
\tau_{Q,\mathrm{pol}}(B)=\mathbb{E}_{i,r}\!\left[Y_{ir}^{v,T_{\mathrm{pol}}}(B)-Y_{ir}^{h,T_{\mathrm{pol}}}(B)\right].
\]
Both arms therefore face the same imposed horizon, although a naturally completed trace can contain fewer than $B$ tokens. For any tested instruction $a$, we also report correct natural completion by the deadline,
\[
K_{ir}^{a}(B)=\mathbf{1}\!\left\{L_{ir}^{a}\leq B\right\}Y_{ir}^{a,N},
\]
and the both-active common-context contrast
\[
\tau_{Q,\mathrm{BA}}^{a}(B)=\mathbb{E}_{i,r}\!\left[Y_{ir}^{a,P_{\mathrm{com}}}(B)-Y_{ir}^{h,P_{\mathrm{com}}}(B)\mid L_{ir}^{a}>B,\ L_{ir}^{h}>B\right].
\]
The four-stratum decomposition partitions every pair into both active, instruction complete/ordinary active, ordinary complete/instruction active, and both complete. It reports each stratum's frequency, within-stratum contrast, and contribution to the unconditional terminal-aware difference. The literal cap-then-answer experiment and the continuation check of the instruction arms replace candidate-logit scoring with the behavioral readout $C$ at $B\in\{512,2048\}$.

For gpt-oss, let $c\in\{\mathrm{low},\mathrm{medium}\}$ denote the lower effort. The matched horizon is the lower-effort policy's item- and replicate-specific terminal length $L_{ir}^{c}$. Let $R_{ir}^{c}=1$ when both token-exact reasoning prefixes can be reconstructed: high effort either reaches $L_{ir}^{c}$ or has a recoverable terminal boundary before it. For replayable pairs, let $\ell_{ir}^{\mathrm{high},c}=L_{ir}^{c}$ when high effort remains active at the matched horizon and $\ell_{ir}^{\mathrm{high},c}=L_{ir}^{\mathrm{high}}$ when high effort finishes earlier. Under a common medium-effort answer context, define
\[
D_{ir}^{c}=Y_{ir}^{c,P_{\mathrm{com}}}(L_{ir}^{c})-Y_{ir}^{\mathrm{high},P_{\mathrm{com}}}(\ell_{ir}^{\mathrm{high},c}).
\]
This definition selects the first $L_{ir}^{c}$ high-effort reasoning IDs when high effort remains active and the complete high-effort reasoning block when it finishes earlier. The primary all-replayable estimand is
\[
\tau_{G,R}^{c}=\mathbb{E}_{i,r}\!\left[D_{ir}^{c}\mid R_{ir}^{c}=1\right].
\]
The high-active subset, in which high effort has not reached a terminal boundary by $L_{ir}^{c}$, is a diagnostic rather than the primary analysis. Fixed-checkpoint both-active strata address the narrower comparison between two unfinished policies. These conditional estimands are not commensurable with $\tau_{Q,\mathrm{pol}}(B)$.

We retain every matched item--replicate pair with an observed lower-effort terminal outcome. When a high-effort trace cannot be replayed, worst-case bounds set its outcome first to correct and then to incorrect (\Cref{sec:methods}). Greedy final-channel continuations repeat the all-replayable matched-horizon comparison with $\rho=C$ and provide the behavioral check.

We apply these probes to Qwen3-4B, -8B, -14B~\citep{yang2025qwen3} and gpt-oss-20b, -120b~\citep{agarwal2025gpt} on GPQA Diamond ($198$ graduate science questions, four choices)~\citep{rein2024gpqa}, MMLU-Pro ($12{,}032$ questions, ten choices)~\citep{wang2024mmlupro}, and Omni-MATH-2 ($4{,}181$ open-ended problems)~\citep{gao2024omni,ballon2026benchmarks}. The main paired Qwen and gpt-oss comparisons use GPQA Diamond and a category-stratified $500$-item MMLU-Pro subset. Omni-MATH-2 supplies natural full-trace accuracy and imposed- or matched-horizon open-ended comparisons. Full prompts, generation settings, parsing rules, and statistical procedures are in \Cref{sec:methods}.

\begin{figure}[!t]
\centering
\includegraphics[width=\textwidth]{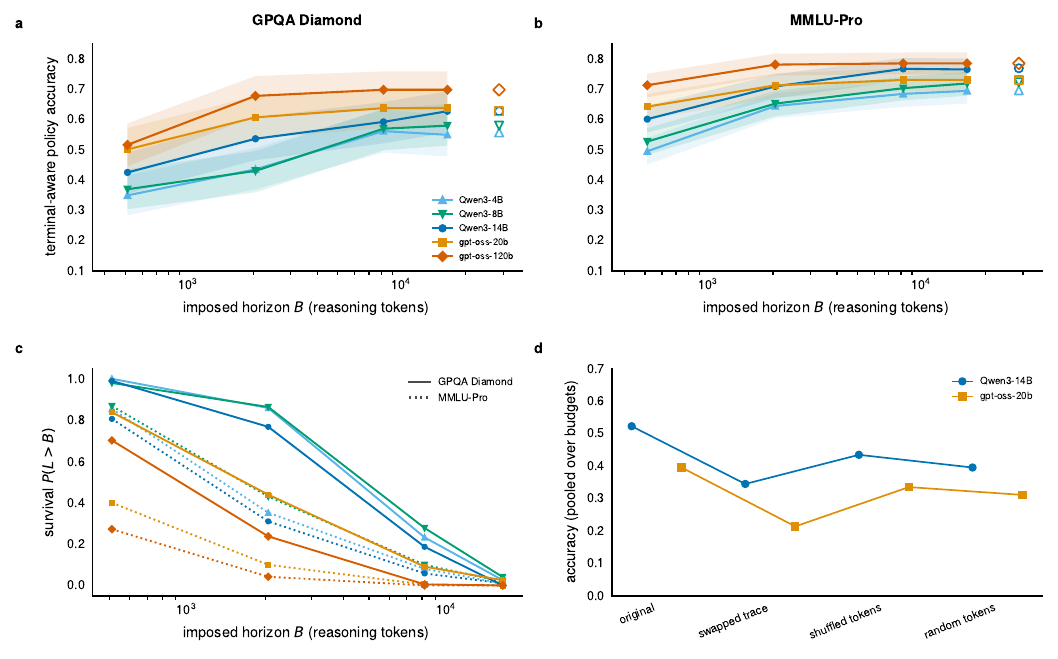}
\caption{\textbf{Longer stopped traces reveal more answer information, and the effect depends on the question-specific reasoning.}
Surprise-stop probes use the full-vocabulary candidate-logit forced-answer readout on GPQA Diamond ($n{=}198$) and the stratified MMLU-Pro subset ($n{=}500$); colors and markers denote models. \textbf{a,b}, Terminal-aware policy accuracy increases with the imposed horizon $B$ and approaches the cohort-matched natural full-trace reference shown at the right of each panel, with $95\%$ item-bootstrap bands. Token positions are paired within model; cross-family positions are descriptive because each model uses its own tokenizer. \textbf{c}, Survival $P(L>B)$ falls sharply with $B$, so late-budget survivor accuracy describes a selected set of long-running items. \textbf{d}, Length-matched prefix controls under the same candidate-logit readout, available for Qwen3-14B and gpt-oss-20b, show that recovered accuracy comes from the item-specific reasoning prefix: original prefixes beat swapped-trace, shuffled-token, and random-token controls on the common four-arm item--budget cohort pooled over budgets. \Cref{fig:truncation} reports the returned-top-$20$ readout comparison.}
\label{fig:fixed-stop-validation}
\end{figure}

\section{Stopped reasoning retains question-specific answer information}

Stopped prefixes reveal increasingly accurate, question-specific answers as the imposed horizon grows (\Cref{fig:fixed-stop-validation}a,b,d). For Qwen3-14B on MMLU-Pro, terminal-aware policy accuracy rises from $50.4\%$ without reasoning to $60.0\%$, $70.8\%$, and $76.6\%$ at $B=512$, $2{,}048$, and $8{,}192$, approaching the cohort-matched full-trace accuracy of $76.8\%$ ($384/500$; \Cref{tab:calibration-survival}). Across all five models, saturation follows natural reasoning length: the shorter gpt-oss traces plateau by $B=2{,}048$, while Qwen remains on the rising part of the curve and smaller models show a few late-horizon reversals or near-ties (\Cref{fig:length-ecdf}). The calibration curves and natural references share one natural-trace cohort. The Qwen prompt comparison below uses separately generated, replicate-matched cohorts (\Cref{sec:methods}).

Survival falls as terminal-aware accuracy rises, making survivor-only accuracy a selected comparison (\Cref{fig:fixed-stop-validation}c and \Cref{tab:calibration-survival}). The model's decision to continue reasoning changes which items remain at a large horizon, so survivor-only accuracy is flat or non-monotone even as the terminal-aware curve improves. At $B=8{,}192$, only $18.7\%$ of Qwen3-14B and $9.1\%$ of gpt-oss-20b GPQA trajectories are still reasoning. We therefore report survival with every horizon curve rather than conflating the evaluation point with the selected set of items that reaches it.

The candidate-logit probe also reproduces completed answers without replaying final-channel text. A separate validation cohort yields $98.3$--$100.0\%$ agreement between replayed-prefix and parsed natural answers across its model--benchmark--policy combinations (\Cref{tab:exact-readout-agreement}). A single-run check also finds no final-channel delimiter in $2{,}788$ terminal gpt-oss replay prefixes (\Cref{tab:terminal-prefix-leak-audit}). These cohorts validate terminal replay but do not exhaustively check the three-replicate token-exact analysis. The fixed-stop and prefix-control probes have no errors or missing valid answer-token IDs. Answers asserted inside the reasoning remain part of the measured behavior.

Length-matched controls confirm that the recovered answer depends on the question-specific prefix (\Cref{fig:fixed-stop-validation}d). Swapping in reasoning from another question reduces accuracy by $18.1$ points for gpt-oss-20b and $17.8$ points for Qwen3-14B, while shuffled-token and random-token controls fall between the swap and original prefixes (\Cref{fig:controls}). These checks support comparisons between generated prefixes at matched evaluation points.

\begin{figure}[!p]
\centering
\includegraphics[width=\textwidth]{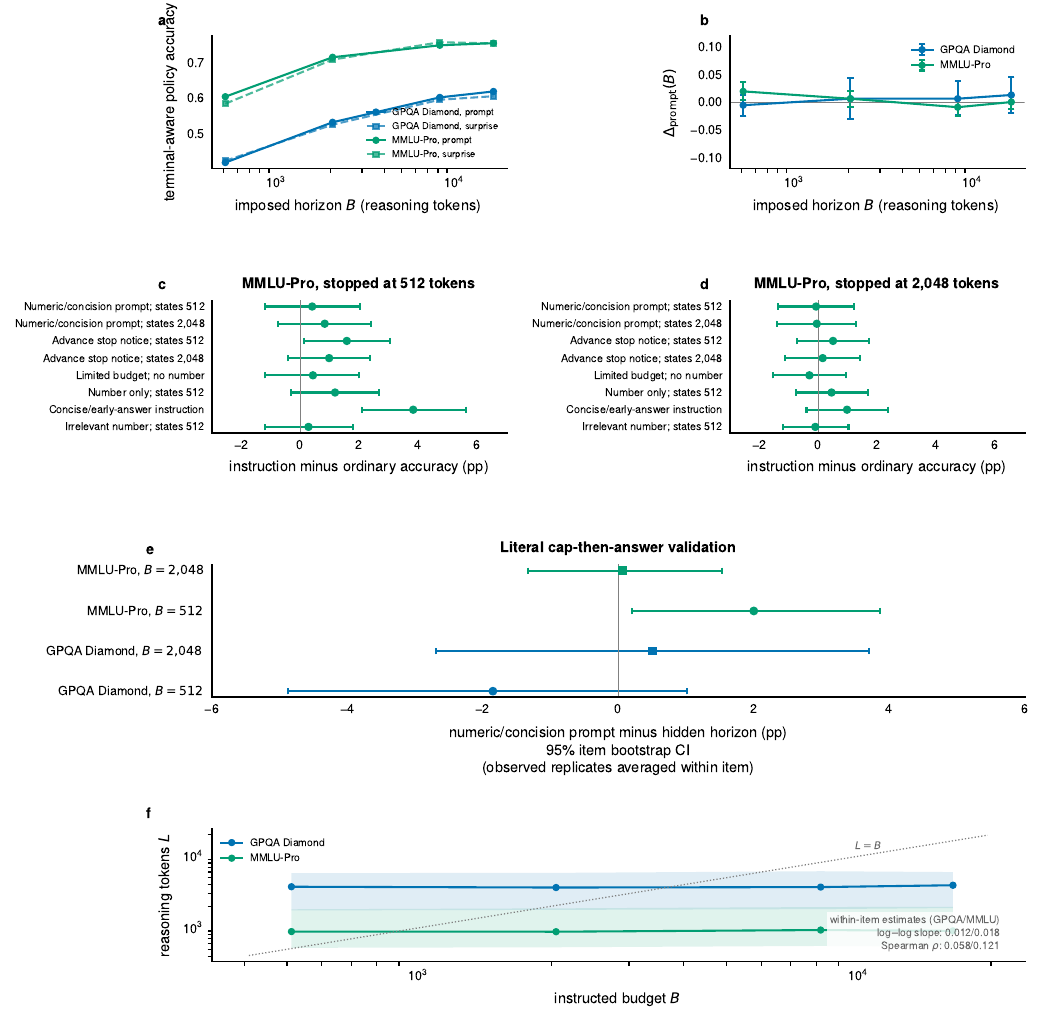}
\caption{\textbf{The numeric/concision prompt shortens Qwen reasoning without a consistent early accuracy gain.} \textbf{a}, Terminal-aware policy accuracy for the broader numeric/concision prompt (solid) and surprise stop (dashed), pairing three explicit generation replicates on GPQA Diamond ($n=198$) and MMLU-Pro ($n=500$) with token-exact replay and candidate-logit scoring. \textbf{b}, The corresponding contrast $\Delta_{\mathrm{prompt}}(B)$. Primary $95\%$ intervals average the three paired replicate outcomes within item and bootstrap items; two-way item $\times$ replicate and crossed-intercept intervals are sensitivity analyses. \textbf{c,d}, MMLU-Pro mechanism controls at imposed stops of $512$ and $2{,}048$ tokens. Points are instruction minus ordinary accuracy, with pointwise $95\%$ intervals from a nested bootstrap over $500$ items and eight paired replicates per item. Generation and probing use stored token IDs. \Cref{tab:qwen-mechanism-audited} reports the instruction grid and matching-number test, \Cref{tab:qwen-mechanism-decomposition} reports completion and both-active results, and \Cref{tab:qwen-mechanism-validation} reports multiplicity-adjusted, direct-contrast, and generated-answer checks. \textbf{e}, Literal cap-then-answer validation at external caps $B=512$ and $2{,}048$, following Qwen's documented two-stage procedure. Points are accuracy differences between the numeric/concision-prompt and hidden-horizon arms; primary bars condition on the three observed trajectory sets and bootstrap items (\Cref{tab:reviewer-qwen-primary}). \textbf{f}, Median realized reasoning length with interquartile bands versus the instructed budget. Realized length changes little with the announced budget: within-item slopes of $\log L$ on $\log B$ are $0.012$ on GPQA Diamond and $0.018$ on MMLU-Pro, with Spearman correlations of $0.058$ and $0.121$ (\Cref{tab:reviewer-qwen-length}).}
\label{fig:budget}
\end{figure}

\section{The numeric/concision prompt shortens Qwen reasoning without a consistent early accuracy gain}

The numeric/concision prompt shortens typical Qwen3-14B reasoning by $12$--$17\%$ without a consistent accuracy gain at the same imposed horizon (\Cref{fig:budget}a,b,f). Across the $32$-fold range from $B=512$ to $16{,}384$, the treated median changes only from $3{,}740.5$ to $3{,}925$ tokens on GPQA Diamond and remains between $881$ and $929$ on MMLU-Pro. The within-item slopes of $\log L$ on $\log B$ are $0.012$ and $0.018$, with rank correlations of $0.058$ and $0.121$ (\Cref{tab:reviewer-qwen-length}). GPQA adherence rises from $1.9\%$ at the tightest budget to $99.2\%$ at the largest, showing that the prompt changes typical length but weakly targets the announced count. The $12$--$17\%$ contrast concerns geometric-mean reasoning length, not arithmetic-mean token cost. The candidate-logit comparison is a controlled analogue of Qwen's documented external-stop-then-answer procedure~\citep{qwen2025thinkingbudget,yang2025qwen3}. It evaluates the externally stopped prefix with deterministic candidate-logit scoring, while the literal generated-answer validation implements the two-stage answer continuation. The prompts and closures are specified in \Cref{sec:methods}.

The same-horizon accuracy differences remain small and inconsistent across the main curve (\Cref{fig:budget}a,b and \Cref{tab:reviewer-qwen-primary}). GPQA Diamond estimates range from $-0.5$ to $+1.3$ points, while the four MMLU-Pro estimates are $+2.0$, $+0.7$, $-0.9$, and $+0.1$ points. Only the MMLU-Pro $B=512$ estimate has both a pointwise interval ($[+0.4,+3.7]$) and simultaneous band ($[+0.1,+3.9]$) above zero. Termination-stratum contributions are also small and benchmark-dependent (\Cref{tab:extended-qwen-strata}). Replicate-sensitive, answer-context, proper-score, and literal cap-then-answer analyses show no consistent gain across benchmarks and horizons (\Cref{tab:reviewer-seed-sensitivity,tab:canonical-readout,tab:reviewer-qwen-scores,tab:extended-context-closure}). Dense trajectory probes show positive log-odds AUC shifts but no clear stable-correct gain (\Cref{tab:extended-answer-availability}). The proper-score table stops at $B=2{,}048$ because the jointly active sets at $B=8{,}192/16{,}384$ contain only $25/1$ GPQA Diamond items and $19/2$ MMLU-Pro items. The companion data release contains those sparse estimates.

A concise/early-answer instruction produces a clear early gain, whereas matching the announced number to the stop does not (\Cref{fig:budget}c,d). The nine-arm experiment uses $500$ MMLU-Pro items with eight replicates per item. At $B=512$, the concise/early-answer instruction raises terminal-aware accuracy from $59.7\%$ to $63.5\%$ ($+3.8$ points, $95\%$ CI $[+2.1,+5.6]$) and reduces median reasoning length from $1{,}083$ to $679$ tokens (\Cref{tab:qwen-mechanism-audited}). Its simultaneous interval over all $16$ instruction--horizon contrasts is $[+1.7,+6.0]$ points, and direct comparisons place it $+2.3$ to $+3.4$ points above the numeric and exact-stop prompts at this deadline (\Cref{tab:qwen-mechanism-validation}). At $B=2{,}048$, its $+1.0$-point accuracy difference is unclear ($[-0.4,+2.4]$). Matching the announced and imposed horizons gives $-0.2$ points ($[-1.0,+0.6]$) for the numeric/concision prompt and $+0.1$ points ($[-0.6,+0.8]$) for advance exact-stop notice. Both intervals lie within about one percentage point of zero, although this is a descriptive comparison rather than an equivalence test.

The $512$-token concise/early-answer gain combines earlier completion with a change in unfinished prefixes (\Cref{tab:qwen-mechanism-decomposition}). Correct natural completion rises from $18.2\%$ to $31.9\%$ ($+13.8$ points), while replaying all prefixes under the ordinary prompt retains a $+3.1$-point accuracy difference ($[+1.5,+4.8]$). Among the $2{,}476$ pairs in which both traces remain active, the common-context difference is $+2.7$ points ($[+0.6,+5.0]$). At $B=2{,}048$, correct completion remains $+7.6$ points higher, but the both-active difference is $+0.5$ points ($[-4.8,+5.5]$; $644$ pairs) and the pooled candidate-logit gain is unclear. Greedy answer continuations retain positive concise/early-answer point estimates at both stops, including differences of $+4.1$ points in the policy context and $+3.5$ under the common context at $B=512$ (\Cref{tab:qwen-mechanism-validation}).

The literal cap-then-answer validation of the numeric/concision prompt reproduces the benchmark split but no broad gain (\Cref{fig:budget}e and \Cref{tab:reviewer-qwen-primary}). A text-reconstructed four-budget crossed analysis shows no diagonal matching-number pattern, and a separate four-horizon concise/early-answer comparison likewise shows no broad advantage (\Cref{fig:qwen-crossed-budget-stop,tab:extended-qwen-scope-detail}). In a pooled MMLU-Pro $B=512$ decomposition, category differences range from $-7.1$ to $+7.8$ points (\Cref{tab:extended-qwen-scope-detail}). A single-rollout Qwen3-8B comparison also shows no broad same-horizon gain (\Cref{tab:extended-scope}). A $200$-item open-ended Omni-MATH-2 comparison finds differences of $+3.5$, $+2.0$, and $-4.1$ points across the three tested horizons, with every interval spanning zero (\Cref{tab:open-ended-validation}). A scripted control moves from $25.0\%$ to $100.0\%$ at $B=64$ and reaches $100.0\%$ in both arms by $B=512$, confirming that the readout detects a known early-answer policy (\Cref{tab:positive-control}).

A text-reconstructed MMLU-Pro subset replication broadens the item sample, but its saved Qwen records lack generation seeds and completion token IDs and therefore do not enter the token-exact primary analysis. On a second $500$-item subset, the numeric/concision contrast is $+2.5$ points at $B=512$ ($[+0.4,+4.8]$), and pooling the original and independent subsets gives $+2.1$ points ($[+0.6,+3.6]$). On that independent subset at $B=2{,}048$, the numeric/concision estimate is $-0.2$ points ($[-2.6,+2.2]$), whereas the concise/early-answer estimate is $+2.7$ points ($[+0.5,+5.0]$; \Cref{tab:qwen-mmlu-b512-independent-subset}). A separate text-reconstructed, eight-instruction prompt-control comparison at $B=512$ likewise shows small, mixed GPQA estimates and larger early MMLU-Pro point estimates, with the concise/early-answer instruction again largest. Its clearest pattern is instruction-dependent reasoning length rather than a consistent accuracy gain (\Cref{tab:qwen-prompt-controls}).

Two diagnostic analyses ask what remains after an early answer has formed. Continuing an incorrect Qwen prefix raises rescue from $20.0$--$30.1\%$ under immediate answering to $26.5$--$38.2\%$ with a long hidden continuation, but $35.8$--$55.8\%$ of continuations remain anchored to the original wrong answer and stating the remaining budget gives no consistent benefit (\Cref{tab:rescue}). Hidden-state readouts recover the exact announced budget at the prompt boundary with balanced accuracy $0.938$ on GPQA Diamond and $0.905$ on MMLU-Pro, then mostly fall to label-shuffle levels after reasoning begins. Binary bundled-prompt membership remains modestly decodable at most post-start token positions, with balanced accuracy from $0.588$ to $0.700$ (\Cref{tab:activation-readout}). These analyses characterize answer rescue, anchoring, and persistence of prompt information; they are not estimates of the matched-horizon effects.

\section{Lower gpt-oss effort improves early accuracy mainly by stopping sooner}

\begin{figure}[!tp]
\centering
\includegraphics[width=\textwidth]{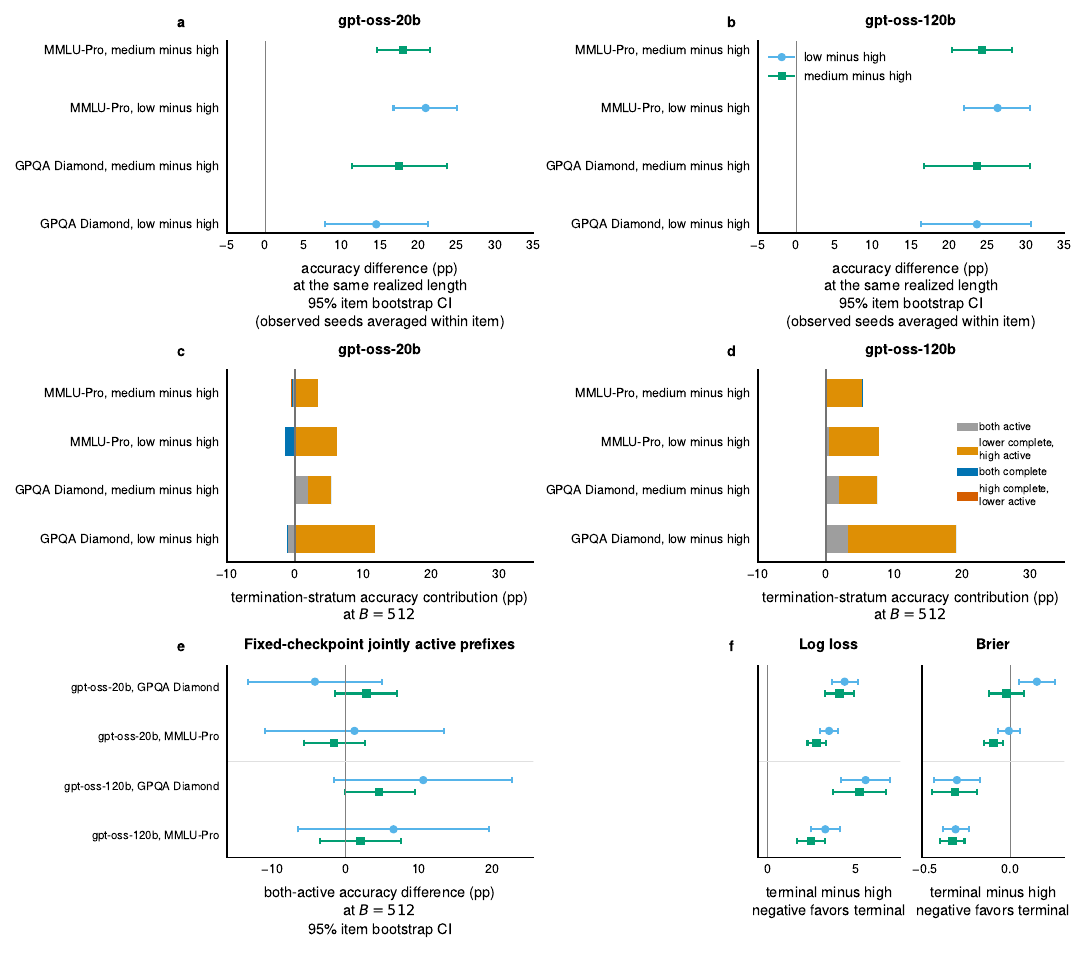}
\caption{\textbf{Lower gpt-oss effort improves early accuracy mainly by stopping sooner, while proper scores qualify the accuracy gain.} \textbf{a,b}, Matched-horizon accuracy gaps for gpt-oss-20b and gpt-oss-120b on GPQA Diamond and MMLU-Pro, pairing each item within three explicit generation replicates. Each completed low- or medium-effort prefix is read under a common medium-effort final-answer context and compared with every reconstructable high-effort prefix at that itemwise horizon. The primary analysis includes early high-effort completions; the high-active subset is reported separately. Primary $95\%$ intervals average the three paired replicate outcomes within item and bootstrap items. \textbf{c,d}, Fixed-checkpoint $B=512$ decomposition by termination stratum. Stacked bars show each stratum's signed contribution to the lower-effort-minus-high accuracy gap; the stratum legend in \textbf{d} applies to both panels. \textbf{e}, Within-stratum accuracy differences for the pairs in which both traces remain active at $B=512$. \textbf{f}, Conditional log-loss and multiclass Brier-score contrasts at the policy-chosen matched horizon. Values are the completed lower-effort readout minus the matched-horizon high-effort readout, so negative proper-score differences favor lower effort. Matched-horizon accuracy, the high-active diagnostic, and generated-answer validation are in \Cref{tab:reviewer-gptoss-exact}; fixed-checkpoint strata are in \Cref{tab:reviewer-gptoss-strata}; and proper scores with their correct-option probability tails are in \Cref{tab:reviewer-gptoss-scores}. The descriptive accuracy--median-length plot is in \Cref{fig:effort-frontier}.}
\label{fig:effort}
\end{figure}

At the low- or medium-effort stopping point, candidate-logit accuracy from the completed lower-effort reasoning block exceeds that from the matched-horizon high-effort prefix in all eight model--benchmark--effort comparisons (\Cref{fig:effort}a,b). The primary comparison replays each completed lower-effort block and its corresponding reconstructable high-effort prefix under a common medium-effort answer context; if high effort finishes before the matched horizon, its completed reasoning block is replayed. Because lower effort chooses the horizon, the result includes its decision to stop and does not establish a uniform advantage among unfinished prefixes. The high-active subset and the fixed $B=512$ termination strata isolate those narrower comparisons. Full estimands and replay rules are in \Cref{tab:main-estimands,sec:methods}. Medium is the default effort setting, and none of the settings specifies a target length~\citep{agarwal2025gpt}.

At the lower-effort terminal horizon, candidate-logit accuracy differences range from $+14.5$ to $+26.3$ points, and every primary interval excludes zero (\Cref{fig:effort}a,b and \Cref{tab:reviewer-gptoss-exact}). The high-active diagnostic remains positive in all eight comparisons, and all eligible outcome pairs were replayable (nonreplayable $=0$; arm-level accuracies and coverage are in \Cref{tab:extended-gptoss-absolute}). Greedy answer continuations preserve the ordering without option normalization, with differences of $+15.5$ to $+25.9$ points and intervals excluding zero throughout (\Cref{tab:reviewer-gptoss-exact}).

Earlier completion explains most of the gpt-oss advantage at the fixed $B=512$ checkpoint (\Cref{fig:effort}c--e). The lower-complete/high-active stratum contributes $+3.4$ to $+15.8$ points and is the largest positive component in all eight comparisons. Contributions from pairs that are both active are smaller and mixed ($-1.0$ to $+3.2$ points), and none of their within-stratum accuracy intervals excludes zero. Both-complete contributions range from $-1.5$ to $+0.1$ points (\Cref{tab:reviewer-gptoss-strata}). Although the lower-complete/high-active share varies from $13.5\%$ to $73.7\%$, its contribution remains positive. Thus, at $B=512$, lower effort's earlier completion accounts for most of the fixed-checkpoint gap.

Proper scores qualify the accuracy gain (\Cref{fig:effort}f). Brier estimates favor the terminal lower-effort arm in seven of eight comparisons, with intervals excluding zero in five, whereas conditional log loss favors the high-effort prefix in all eight, with all intervals excluding zero (\Cref{tab:reviewer-gptoss-scores}). This divergence is concentrated in the lower tail of correct-option probability. Across comparisons, $20.9\%$--$36.0\%$ of probed terminal answers assign it less than $10^{-4}$, compared with $0.2\%$--$11.7\%$ of matched high-effort prefixes (\Cref{tab:reviewer-gptoss-scores}). An incorrect completed trace can concentrate mass on its chosen letter, producing near-zero correct-option probabilities that dominate unbounded log loss; the corresponding incorrect-answer score tails are in \Cref{tab:extended-gptoss-score-tails}. Brier score caps these tails, whereas argmax accuracy ignores confidence.

Replicate-specific, two-way, and crossed-intercept calculations agree with the primary finite-run result (\Cref{tab:reviewer-seed-sensitivity}). Answer-context, source-context, and closure checks at the reported checkpoints do not reverse the family-level pattern (\Cref{tab:canonical-readout,tab:extended-context-closure}). In model-specific paired cohorts of 475 items for gpt-oss-20b and 490 for gpt-oss-120b from a difficulty-stratified 500-item Omni-MATH-2 draw, completed lower-effort answer accuracy exceeds matched-horizon high-effort answer accuracy by $+33.5$ to $+42.0$ points across four comparisons; fixed-checkpoint comparisons are reported in \Cref{tab:extended-scope}, but the single rollout does not establish that the estimate generalizes across sampled generations (\Cref{tab:open-ended-validation}). Dense probes place the lower-effort answer earlier in the trajectory: terminal-aware AUC differences are positive in all eight gpt-oss comparisons, whereas jointly active early-prefix differences are much smaller (\Cref{tab:extended-answer-availability,fig:answer-availability}). Surface diagnostics show more digits and operators, fewer explicit alternative-strategy cues, and effort classification accuracy of $0.70$ from the first $64$ tokens (\Cref{fig:structural,fig:prefix-classifiers}). These item-conditional diagnostics are not used to compare effect magnitudes across the two case-study estimands.

Adding numeric-budget wording to the gpt-oss effort instruction produces no broad $B=512$ advantage over effort alone (\Cref{tab:gptoss-numeric-effort}). Ten of the twelve intervals span zero. The gpt-oss-20b medium-effort difference on GPQA Diamond is $-5.6$ points ($[-10.6,-0.5]$); the gpt-oss-20b high-effort difference on MMLU-Pro is $+2.7$ points, with a rounded lower endpoint of zero ($[+0.0,+5.5]$). The effort-only baseline ends at $B=512$, leaving longer announced-budget contrasts unmatched.

High effort can nevertheless be more accurate when allowed to finish (\Cref{fig:effort-frontier}). The completed-effort view describes accuracy against median trace length rather than arithmetic-mean token use or expected API cost; it provides completed-trace context but is not the matched-horizon estimand. At the same fixed $B=512$ checkpoint, low- or medium-minus-high differences range from $+2.8$ to $+19.0$ points (\Cref{fig:effort-fixed,tab:reviewer-gptoss-strata}). Effort changes both when the model finishes and what it answers at a given checkpoint.

\section{Discussion}

Matched-horizon comparisons reveal two distinct ways to answer earlier: a control can make the model finish sooner, change the answer available in an unfinished prefix, or do both. The Qwen3-14B numeric/concision prompt mainly reduces typical reasoning length, whereas the concise/early-answer instruction improves the early answer through both earlier completion and a change in unfinished prefixes. Lower gpt-oss effort produces a substantially better answer at its chosen stopping point, but most of its $512$-token advantage comes from having finished while high effort is still reasoning. Differences between jointly active gpt-oss prefixes are smaller and depend on the model and task, and high effort can be more accurate once allowed to finish.

Reasoning-interface evaluations should therefore separate correct completion by a deadline, the answer obtained by stopping an unfinished run, differences between jointly active runs, and probability assigned to the correct answer. The gpt-oss results show how earlier completion can improve the first two outcomes, the Qwen concise/early-answer result shows that an instruction can also change unfinished reasoning, and the gpt-oss log-loss result shows that a better selected answer need not imply uniformly better probability quality.

Matched-horizon evaluation connects language-model reasoning to work on anytime algorithms, deadline-aware planning, and rational metareasoning~\citep{dean1988timeplanning,zilberstein1996anytime,russell1991dotherightthing}. Those traditions separate the value of an intermediate answer from the policy that allocates computation before a deadline. A shorter final trace does not establish that the model planned around a deadline, and a better answer at lower terminal cost does not imply continuous remaining-budget tracking.

The same distinction may matter when reasoning traces are used to study behavior under resource pressure~\citep{baker2025monitoring,korbak2025chainthoughtmonitorabilitynew}, although the present tasks contain no competing objective or safety constraint and do not test behavioral misalignment.

\paragraph{Limitations.} Three limitations bound our conclusions. First, primary experiments cover two model families and two multiple-choice benchmarks with noncomparable prompt and effort interventions. We therefore restrict claims within families; single-rollout Qwen3-8B and Omni-MATH-2 estimates describe only the sampled trajectories. Second, three generation replicates limit stochastic inference; text-reconstructed Qwen subset and prompt-control cohorts lack token IDs and verified seeds. We report replicate-specific, two-way, and crossed-intercept sensitivity analyses; these cohorts inform subset replication and prompt patterns, not primary token-exact estimates. Third, forced-answer readout affects unfinished-prefix accuracy, and near-zero correct-option probabilities affect gpt-oss log loss. We validate against completed answers and continuations and report valid-option mass, Brier scores, medians, tails, and a scripted positive control. Conclusions remain specific to the tested models, tasks, horizons, and answer procedures.

\clearpage
\section*{Acknowledgements}
This research was supported by funding from the Flemish Government under the ``Onderzoeksprogramma Artifici\"ele Intelligentie (AI) Vlaanderen'' program.
Andres Algaba acknowledges support from the Francqui Foundation (Belgium) through a Francqui Start-Up Grant and a fellowship from the Research Foundation Flanders (FWO) under Grant No.1286924N.
Vincent Ginis acknowledges support from Research Foundation Flanders under Grant No.G032822N and G0K9322N.
The computational resources and services used in this work were provided by the VSC (Flemish Supercomputer Center), funded by the Research Foundation Flanders (FWO) and the Flemish Government - department WEWIS.

\section*{Author contributions}
A.A. and F.C. conceived the study and designed the experiments. A.A. implemented the probing and generation code and performed the analysis. F.C., V.G., and A.A. interpreted the results and wrote the manuscript.

\section*{Data and code availability}
The code for this publication is publicly available at \url{https://github.com/AndresAlgaba/shorter_reasoning}. \\
Data associated with this study are available in a public repository at \url{https://doi.org/10.5281/zenodo.21773770}.

The GPQA Diamond dataset~\cite{rein2024gpqa} is available at \url{https://huggingface.co/datasets/Idavidrein/gpqa}. \\
The MMLU-Pro dataset~\cite{wang2024mmlupro} is available at \url{https://huggingface.co/datasets/TIGER-Lab/MMLU-Pro}. \\
The Omni-MATH-2 dataset~\cite{ballon2026benchmarks} is available at \url{https://huggingface.co/datasets/martheballon/Omni-MATH-2}.

The Qwen3 model family~\cite{yang2025qwen3} is available at \url{https://huggingface.co/collections/Qwen/qwen3}. \\
The gpt-oss model family~\cite{agarwal2025gpt} is available at \url{https://huggingface.co/collections/openai/gpt-oss}.

We used Python 3.13.1 across separate environments. Data processing, analysis, and visualization used \textit{pandas} 3.0.0, \textit{pyarrow} 23.0.0, \textit{scikit-learn} 1.7.2, \textit{SciPy} 1.17.0, \textit{statsmodels} 0.14.6, and \textit{matplotlib} 3.10.8. Model generation used \textit{vLLM} 0.12.0 with \textit{flashinfer-python} 0.5.3; local candidate-logit readouts used \textit{PyTorch} 2.11.0 and \textit{Transformers} 5.8.0.

\clearpage
\bibliographystyle{unsrtnat}
\bibliography{references}

\clearpage
\appendix
\makeatletter
\@addtoreset{equation}{section}
\makeatother
\renewcommand{\theequation}{\arabic{equation}}
\renewcommand{\theHequation}{app.\thesection.\arabic{equation}}

\setcounter{figure}{0}
\renewcommand{\thefigure}{A\arabic{figure}}
\renewcommand{\theHfigure}{A.\arabic{figure}}
\setcounter{table}{0}
\renewcommand{\thetable}{A\arabic{table}}
\renewcommand{\theHtable}{A.\arabic{table}}

\section{Methods}
\label{sec:methods}

\subsection{Datasets and sampled items}
We use GPQA Diamond~\citep{rein2024gpqa}, $198$ expert-validated graduate-level multiple-choice questions in biology, chemistry, and physics with four choices (A--D); MMLU-Pro~\citep{wang2024mmlupro}, $12{,}032$ questions across $14$ categories with ten choices (A--J); and Omni-MATH-2~\citep{gao2024omni}, $4{,}181$ open-ended problems with free-form answers, introduced by \citet{ballon2026benchmarks}. The Qwen numeric/concision experiments use the full GPQA Diamond set and a $500$-item MMLU-Pro subset drawn by proportional stratified sampling over the $14$ categories with seed $0$, so that every budget condition is evaluated on the identical items. The nine-arm mechanism experiment reuses this MMLU-Pro subset.
\subsection{Models, prompts, and generation}
We evaluate Qwen3-4B, -8B, -14B~\citep{yang2025qwen3} with a $32{,}768$-token context, and gpt-oss-20b, -120b~\citep{agarwal2025gpt} with a $131{,}072$-token context, all served locally from pinned Hugging Face snapshots. Qwen models run in bfloat16 without weight quantization; gpt-oss models use their released MXFP4 quantization.

Full reasoning traces are generated in thinking mode with model-recommended sampling: Qwen3 uses temperature $0.6$, top-$p$ $0.95$, and top-$k$ $20$ with up to $28{,}000$ reasoning tokens; gpt-oss uses temperature $1.0$, top-$p$ $1.0$, and top-$k$ $50$ with up to $126{,}000$ tokens. The main Qwen budget curve and gpt-oss experiments use three independently seeded global generation replicates. We store full completion token IDs and seeds, require every planned generation to be present, and reject any pair of global replicates whose complete per-item token-sequence collections are identical within a model--benchmark--condition combination. All $23{,}034$ planned generations are present, no two global replicate collections are identical, and the largest incidental item-level exact-match rate within a combination is $0.6\%$.

The primary candidate-logit readouts and controls use local Transformers/PyTorch forward passes rather than vLLM's returned log probabilities. Effort-conditioned gpt-oss traces are generated through the chat template with thinking enabled and the reasoning-effort setting low, medium, or high. In the released template this renders \texttt{Reasoning: low|medium|high} before the user message and assistant generation prompt. Prompted Qwen3 traces append the tested textual instruction to the system prompt and otherwise leave generation unconstrained, so the external horizon $B$ is applied identically to prompted and unprompted trajectories at probe time. The returned-top-$20$ readout is retained only for the fixed-stop calibration (\Cref{fig:truncation}).

The numeric/concision prompt is present during treated generation and absent during ordinary generation. Prompt-control generations replace it with the alternatives in \Cref{tab:prompt-texts}, using the same imposed horizon and the same ordinary-generation baseline within that control cohort. The policy-context readout reconstructs each arm from its stored source-prompt IDs, so the generation-time instruction remains visible when an active prefix is probed. The common-context readout instead uses the ordinary base-prompt IDs for every Qwen arm, replays the arm's exact reasoning IDs, and appends the same neutral closure. Completed reasoning blocks are also replayed under this common context; their original final answers are retained only in the terminal-aware policy-context estimand. Budget-present controls add the numeric/concision prompt only at answer time while holding the source prefix fixed. The token horizon is defined with the source model's tokenizer, so absolute horizons are paired within model and remain descriptive across model families.

\subsection{Trace parsing and forced-answer probes}
Qwen3 reasoning is the token sequence between the thinking tags. gpt-oss reasoning is the analysis-channel token sequence, including the channel marker needed to form a valid replay. For the primary token-exact paths, we locate the terminal boundary in the stored completion IDs, exclude all following final-answer IDs, and concatenate the chosen prompt IDs, the first $B$ reasoning IDs or the complete reasoning block, and the closure IDs. No generated reasoning prefix is decoded and retokenized. We verify the replayed prompt tokens against the prompts used at generation, check boundary and special-token placement, and record hashes of the source completion and the complete replay input.

For gpt-oss at the lower-effort matched horizon $L_c$, a high-effort trace is replayable when it contains at least $L_c$ stored reasoning IDs, whether or not it later reaches a terminal boundary, or when a terminal boundary identifies a completed reasoning block before $L_c$. A stored sequence that ends before $L_c$ without such a boundary is retained among the matched pairs but has an unavailable high-effort outcome. This rule includes capped high-effort traces whenever their stored reasoning IDs reach the matched horizon. The primary analysis uses all replayable pairs; the high-active diagnostic retains only the first case.

The nine-arm Qwen mechanism experiment follows the same token-exact construction at $B\in\{512,2{,}048\}$. All $36{,}000$ generations and $72{,}000$ policy-context replays are present, with no probe errors or context truncations. Cohorts without stored token IDs use text reconstruction: the reasoning text is tokenized with the source model, cut to the requested horizon, decoded, and embedded in the probe prompt. They do not supply primary token-exact evidence.

The candidate-logit probe contains the single-letter system instruction, question, replayed reasoning prefix, and a family-specific suffix that closes reasoning. For gpt-oss, the suffix closes the analysis channel and opens the final channel. One deterministic forward pass extracts full-vocabulary logits at the first answer position. The valid set contains every supported one-token bare-letter and leading-space spelling; spellings for the same answer are aggregated by log-sum-exp. If $p_y$ is the resulting unnormalized probability for option $y$, we report valid-option mass $M=\sum_{y\in\mathcal{Y}}p_y$ and the conditional distribution $q_y=p_y/M$. The predicted answer is $\arg\max_y q_y$, conditional log loss is $-\log q_{y^\star}$, and multiclass Brier score is $\sum_y(q_y-\mathbf{1}\{y=y^\star\})^2$. Lower log loss and Brier score are better. Qwen policy-context proper-score contrasts are restricted to paired active prefixes because a retained natural terminal answer has no associated probe distribution. Common-context terminal replay supplies probe distributions for completed reasoning blocks as well.

Each candidate-logit probe records the valid answer-token IDs, missing IDs, coverage, and any probe error. Terminal-prefix agreement is reported in \Cref{tab:exact-readout-agreement}. In a separate single-run cohort, none of the $2{,}788$ terminal gpt-oss replay prefixes contains a final-channel delimiter (\Cref{tab:terminal-prefix-leak-audit}); this check does not exhaustively cover the three-replicate primary cohort. The returned-top-$20$ calibration is incomplete by construction and supports no primary estimate (\Cref{fig:truncation}). For the nine Qwen instruction arms and the central gpt-oss comparisons, we also append the appropriate final-channel closure to the exact stored prefix and greedily generate up to $16$ unconstrained answer tokens. Missing or malformed answer letters count as incorrect and as parse failures. These continuations decode answers deterministically from independently sampled reasoning prefixes, so the reasoning replicate is the stochastic unit. Terminal-prefix readouts reproduce the naturally sampled final answer for $98.3$--$100.0\%$ of completed prefixes in the terminal-agreement check.

The Omni-MATH-2 analyses use free-form answers judged with \texttt{gpt-5-mini-2025-08-07}, following \citet{ballon2026benchmarks}. The Qwen comparison uses a $200$-item subset and includes item--horizon comparisons for which both prompt and surprise answers were judged. The gpt-oss comparison draws $500$ items using Omni-MATH's native difficulty metadata~\citep{gao2024omni}; the analyzed model-specific paired cohorts contain 475 items for gpt-oss-20b and 490 for gpt-oss-120b. All gpt-oss prefixes are replayed under a common medium-effort final-answer context, and the analysis includes all $15{,}440$ judged answers across the matched-horizon and fixed-checkpoint comparisons. These comparisons use one rollout per condition and therefore do not characterize variation across sampled generations. An independent full-trace judging pass agrees on $94$--$99\%$ of items across the five models. Structural-marker annotations use the same snapshot with the released JSON-schema prompt.

\subsection{Readout contexts and answer availability}
We run policy- and common-context readouts to separate the full interface contrast from differences carried by the generated reasoning prefix. The Qwen policy-context readout preserves the stored condition-specific prompt. The Qwen common-context readout uses the ordinary base prompt and neutral closure for every arm and replays completed as well as active reasoning blocks. A paired answer-context control then reintroduces the numeric/concision prompt only at answer time while holding the bundled-prompt prefix fixed. For gpt-oss, the primary matched-horizon readout uses a common medium-effort answer context for every low-, medium-, and high-source prefix. Additional controls cross the source effort with the effort setting used only at readout. We treat candidate-logit scoring as an option-normalized, deterministic paired instrument for answer availability at a standardized prefix. Greedy final-answer continuations provide the corresponding behavioral readout.

A source-context branch readout reconstructs the source prompt and reasoning prefix at the answer point instead of placing the prefix in a neutral context. Qwen surprise-stop traces keep the budget absent, Qwen bundled-prompt traces retain the numeric-budget instruction, and gpt-oss traces retain the source effort setting. This text-based recomputation approximates the trajectory's original prompt context rather than recovering the generation-time KV cache.

To test when answers become available in the trajectory, we run a dense readout over the generated prefixes at checkpoints $0,64,\ldots,512$. For gpt-oss, the source prefixes come from low, medium, and high effort traces, and all are read with a common medium-effort readout. For Qwen3-14B, the source prefixes compare bundled-prompt and surprise-stop traces at the $B=512$ condition, and the readout uses a neutral budget-free closure. At each checkpoint we extract candidate-letter probabilities from full-vocabulary next-token logits and compute the gold-answer log-odds, $\log p(y^\star)-\log\sum_{y\ne y^\star}p(y)$. Natural termination is treated as an absorbing state. For a trace that terminates at length $L_{ic}<512$, we carry its terminal forced-answer distribution forward, so that $q_{ic}(t)=q_{ic}(L_{ic})$ for every $t\geq L_{ic}$. The resulting terminal-aware answer-availability AUC is a token-normalized trapezoidal average that includes both evidence accumulated before termination and the practical consequence of having completed the answer early. We also report retrospective stable correctness, which counts a checkpoint as correct only if the predicted answer remains correct at all later checkpoints through $512$.

As a sensitivity analysis, we recompute the terminal-aware answer-availability AUC over $0$--$128$ tokens after retaining only items for which both compared traces are still reasoning after $128$ tokens ($L>128$). We call this an active-pair sensitivity analysis rather than a separately identified causal estimate, because conditioning on survival changes the target comparison.

\subsection{Literal Qwen cap-then-answer validation}
The literal Qwen validation follows the documented two-stage thinking-budget shape while retaining the paired item--replicate design of the candidate-logit analysis. The design crosses GPQA Diamond ($198$ items) and the primary MMLU-Pro subset ($500$ items) with $B\in\{512,2{,}048\}$, hidden/surprise and numeric/concision-visible arms, and the same three generation replicates. In the first call, vLLM generates the reasoning block with the Qwen sampling settings and an external maximum of $B$ tokens, stopping early if \texttt{</think>} appears. If generation hits the cap, we append the Qwen limited-time closure and close the thinking block; otherwise we retain the naturally closed state. In the second call, the model greedily generates up to $16$ answer tokens. Malformed or missing choice labels are parse failures and count as incorrect. We store the full reasoning- and answer-stage token IDs, hash the reasoning and answer stages together, and reject a benchmark--arm--budget combination if two complete replicate collections are record-identical across its items. Accuracy contrasts pair arms within item, replicate, and budget. Primary intervals average the three observed replicate outcomes within item and bootstrap items; replicate-specific, two-way item $\times$ replicate, and crossed-intercept calculations are sensitivities. Candidate-logit deltas are included only as a readout comparison.

\subsection{Scripted positive control}
To check that the same-imposed-horizon readout detects a known early-answer trajectory, we construct a scripted synthetic positive control with $96$ balanced four-choice items. The question contains no answer information. All options are generic labels A--D, and the gold label is assigned by a deterministic balanced key. The long policy uses a neutral reasoning prefix in which the candidate-answer line appears after the first $64$ source tokens but before $512$ source tokens. The short policy uses a prefix beginning with \texttt{candidate answer: X}, where \texttt{X} is the gold label. A short $B=512$ policy reuses the long trace, so the positive control should be large at $64$ tokens and vanish once both traces reveal the candidate answer.

We generate probes directly from these scripted reasoning prefixes, truncate them with the Qwen tokenizer, and apply the same candidate-logit answer-token readout used for the paired multiple-choice contrasts and controls. Accuracy moves from $25.0\%$ for the long policy to $100.0\%$ for the short policy at $B=64$; both reach $100.0\%$ at $B=512$ (\Cref{tab:positive-control}). This positive control validates the truncation and readout on a known text-level policy difference. It is not evidence that Qwen3-14B or gpt-oss spontaneously condition on exact numeric budgets.

\subsection{Trajectory diagnostics and controls}
To test whether a condition is legible in the early reasoning, we train a logistic-regression classifier to predict the condition from the first $k$ reasoning tokens of a trace, for $k \in \{64, 128, 256, 512, 1024, 2048\}$. The input is the surface text of the first $k$ generated reasoning tokens and nothing else: no question, no answer, no trace length, and no model identity. Traces shorter than $k$ are excluded at that $k$ rather than padded, so termination or residual length is not provided as an implicit feature. We use two feature sets, TF-IDF over word $n$-grams ($1$--$3$) and character $n$-grams ($3$--$5$), and a set of length-normalized style counts (backtracking and verification cue words, punctuation, operator and digit density, answer-letter mentions, mean word length). The label is the effort setting (low, medium, high) for gpt-oss and the bundled-prompt condition (bundled prompt versus surprise-stop) for Qwen3. We evaluate with $5$-fold cross-validation grouped by item, so the same question never appears in both train and test, and compare against a label-shuffle baseline of $20$ permutations.

We analyze the generated reasoning text rather than treating prompt-boundary hidden-state decodability as evidence that the reasoning policy changed. The effort or budget instruction is present in the prompt, whereas the surface text is generated by the policy itself. We therefore use text classifiers for the early-policy diagnostic.

To characterize how the early reasoning differs across conditions, we compute length-normalized structural markers on the first $\min(256,L)$ reasoning tokens of each trace: the position of the first computational cue, the position of the first answer assertion, the rates of backtracking, verification, and alternative-strategy cue phrases per thousand words, characters per word, and digit and operator density. Raw trace length is not among the reported markers. Markers are compared between paired conditions with bootstrap intervals over items on the same GPQA Diamond and stratified MMLU-Pro analysis subsets used for the main probe contrasts. These are deterministic lexical patterns and capture surface form rather than meaning. To validate them, we annotate a stratified sample of $250$ trace excerpts ($50$ per condition) with GPT-5 mini for the same constructs and correlate the cue-based markers with the annotations ($n{=}204$ matched). The backtracking and alternative-strategy markers agree with the annotations (Spearman $\rho{=}0.32$ and $0.19$, respectively; both $p<0.01$), whereas the verification marker is weaker and not significant (point-biserial $r{=}0.11$, $p=0.109$), and the two position-based markers, restatement length and first-answer commitment, do not validate. We base our structural interpretation on the digit and operator density and the alternative-strategy marker, which are either objective character statistics or annotation-supported.

To test whether the recovered accuracy is instance-specific, we re-probe each surprise-stop prefix against three length-matched controls at the same budget $B$: a \emph{random} prefix of $B$ tokens sampled uniformly from the vocabulary excluding special tokens; a \emph{swap} prefix taken as the first $B$ reasoning tokens of a different question's trace; and a \emph{shuffle} prefix that permutes the first $B$ original reasoning tokens. Swap donors must be distinct and contain at least $B$ source tokens. We exclude $31$ swap prefixes whose donor traces are shorter than $B$ and restrict all pooled contrasts to the remaining common four-arm item--budget cohort. The controls isolate, respectively, context length and position, coherent but irrelevant reasoning, and token identity without order.

\subsection{Statistical inference}
The primary Qwen curve and gpt-oss contrasts use three independently sampled global generation replicates. Conditions are paired within item and replicate. Primary intervals condition on the three trajectory sets observed in this experiment: we first average the three paired replicate outcomes within item and then draw $10{,}000$ bootstrap resamples of items. Bootstrap resampling uses NumPy seed $0$ by default, and every reported interval uses a fixed, recorded resampling seed. We report each replicate separately with an item bootstrap. A paired two-way item $\times$ replicate bootstrap and a crossed item--replicate variance decomposition are sensitivity analyses. With only three global replicate levels, those sensitivities do not provide precise population-over-seed inference.

For the model-based sensitivity analysis, we fit the paired differences with crossed item and replicate random intercepts. Variance components are estimated by method of moments on complete item--replicate matrices; uncertainty for the grand mean uses the item, replicate, and residual components with a Satterthwaite $t$ approximation. With three global replicates, this calculation is a diagnostic of sensitivity to trajectory-set variation rather than a precise variance-component analysis.

The nine-arm Qwen mechanism experiment uses eight item-specific replicates for each of $500$ MMLU-Pro items. A deterministic hash of the item ID, replicate number, and base seed defines one generation seed for each item--replicate pair; the same seed is reused across all nine prompt arms, and all $4{,}000$ item--replicate seeds are distinct. The arms are hidden horizon; numeric/concision prompts announcing $512$ or $2{,}048$ tokens; advance exact-stop notices announcing the same two horizons; generic limited-budget wording; number-only wording; a concise/early-answer instruction; and an irrelevant-number placebo. Primary intervals use a nested paired bootstrap that first resamples items and then resamples the eight paired replicates within each selected item. The arm--horizon family is covered by simultaneous intervals obtained from the maximum absolute deviation across the treatment-versus-ordinary contrasts in each bootstrap draw. Direct prompt-to-prompt contrasts use the same paired draws. The direct diagonal statistic averages the advantage of announcing $512$ rather than $2{,}048$ tokens when the imposed stop is $512$ with the reverse contrast when the imposed stop is $2{,}048$. A positive value would indicate that matching the announced and imposed horizons helps.

For the three-replicate Qwen accuracy curve, each item's paired outcomes are first averaged over the observed replicates. The primary bootstrap resamples items and preserves each selected item's complete budget curve. The simultaneous $95\%$ band uses the $95$th percentile of the maximum absolute deviation from the estimated curve. A scalar summary integrates the accuracy difference over $\log_2 B$ with trapezoidal weighting and divides by the observed log-budget span; its percentile interval is computed from the same item-bootstrap draws. A two-way item $\times$ replicate version of both summaries is retained as a sensitivity analysis. The $\pm3$ percentage-point margin is a descriptive scale, not a preregistered utility threshold or an equivalence test.

For the gpt-oss matched-horizon comparison, we retain every matched item--replicate pair with an observed lower-effort terminal outcome and bound the effect of unreplayable high-effort traces. Let $\mathcal{O}_c$ and $\mathcal{M}_c$ denote the replayable and missing-high sets, let $N_c=|\mathcal{O}_c|+|\mathcal{M}_c|$, and write the observed completed lower-effort outcome as $Y_{ir}^{c,N}$. Worst-case binary bounds set every missing high-effort outcome first to correct and then to incorrect:
\[
\underline{\tau}_{G}^{c}=\frac{\sum_{(i,r)\in\mathcal{O}_c}D_{ir}^{c}+\sum_{(i,r)\in\mathcal{M}_c}(Y_{ir}^{c,N}-1)}{N_c},\qquad
\overline{\tau}_{G}^{c}=\frac{\sum_{(i,r)\in\mathcal{O}_c}D_{ir}^{c}+\sum_{(i,r)\in\mathcal{M}_c}Y_{ir}^{c,N}}{N_c}.
\]

Several gpt-oss and Qwen run pairs have duplicate decoded response strings. Because their token IDs and seeds are unavailable, we cannot determine whether the underlying token sequences are identical or reconstruct the cause; these runs are used only for run-level diagnostics. Prompt-control, readout-context, crossed-horizon, and subset-replication cohorts without stored token IDs or verified seeds are excluded from the token-exact primary estimates. The Omni-MATH-2 comparisons use one rollout per condition.

Single-run binary matched-pair checks use item-bootstrap intervals. Where McNemar's test is reported, it is exact when the discordant count is below $25$ and otherwise uses the continuity-corrected $\chi^2$ approximation~\citep{McNemar_1947}. Repeated-budget prefix-control summaries first average paired outcomes over budgets within item and then use an item-cluster bootstrap. Every budget curve reports the total item count, the number surviving to $B$, the survival rate, and both survivor-only and terminal-aware policy accuracy.

\subsection{Replication, interface comparisons, and diagnostic analyses}

\subsubsection{Text-reconstructed Qwen subset replication and prompt comparisons}
The text-reconstructed MMLU-Pro subset replication compares the $500$-item subset used in the main design with a second category-stratified $500$-item subset that excludes those items. Both subsets compare the numeric/concision and surprise-stop conditions at $B=512$ over three nominal runs, and the independent subset also compares the numeric/concision and concise/early-answer conditions with the same surprise-stop cohort at $B=2{,}048$. Accuracy and survival use the candidate-logit readout. Intervals resample items and nominal runs in a paired two-way bootstrap. Generation seeds and completion token IDs are unavailable for these cohorts, so the estimates describe the observed nominal runs and do not support generalization across generation seeds. Only aggregate estimates and intervals are available for the two $B=2{,}048$ rows, so they cannot be regenerated independently and are treated as descriptive evidence (\Cref{tab:qwen-mmlu-b512-independent-subset}).

The text-reconstructed prompt-control comparison evaluates eight Qwen3-14B instructions against the same surprise-stop cohort at $B=512$ on GPQA Diamond and MMLU-Pro. Each estimate averages three nominal runs within item before an item-cluster bootstrap. \Cref{tab:qwen-prompt-controls} reports terminal-aware accuracy and median full-trace length; paired log-length contrasts are included in the companion data release. Completion token IDs and verified generation seeds are unavailable for these cohorts.

\subsubsection{Numeric wording with gpt-oss effort}
The within-model gpt-oss comparison crosses the numeric-budget instruction with low, medium, and high effort; $B\in\{512,2{,}048,8{,}192,16{,}384\}$; both gpt-oss model sizes; and GPQA Diamond or MMLU-Pro. The directly matched comparison adds the numeric wording to each effort setting and evaluates both conditions at $B=512$ (\Cref{tab:gptoss-numeric-effort}). Longer announced-budget comparisons lack a directly matched effort-only baseline because the effort-only comparison covers only $B=512$, and low-effort traces often terminate before later checkpoints.

\subsubsection{Wrong-prefix continuation}
The wrong-prefix continuation analysis selects natural Qwen3-14B traces with an incorrect parsed final answer, retains the first $B$ source-tokenized reasoning tokens, and presents that prefix with the same question. The answer-now condition closes reasoning immediately with the forced-answer suffix. Continuation conditions instead allow $R$ additional reasoning tokens before the forced answer, with the remaining budget either hidden or stated in the prompt. Both conditions use the returned-top-$20$ readout, whose incomplete option coverage applies symmetrically. We evaluate $B\in\{512,2{,}048,8{,}192\}$ and $R\in\{512,2{,}048,8{,}192,32{,}768\}$ on GPQA Diamond and the stratified MMLU-Pro subset. The displayed $R=32{,}768$ conditions are limited by the remaining context window (\Cref{tab:rescue}).

\subsubsection{Activation readout of the announced budget}
The activation analysis tests whether Qwen3-14B hidden states retain the announced numeric budget. For each item with one natural trace and all four numeric-budget traces, we feed the prompt and the first $k\in\{0,64,256,512,1024,2048\}$ generated reasoning tokens. We random-project hidden states to $256$ dimensions at layers $0$, $10$, $20$, $30$, and $39$, then fit logistic-regression readouts with item-grouped cross-validation. One readout predicts whether the trace uses the numeric/concision prompt; the other predicts which of the four announced budgets generated a treated trace. Each is compared with $20$ label-shuffle baselines. A focused check uses $100$ shuffles and $C\in\{0.1,1,10\}$ for the post-start exact-budget comparisons (\Cref{tab:activation-readout}).

\clearpage
\section{Figures}

\begin{figure}[!htbp]
\centering
\includegraphics[width=\textwidth]{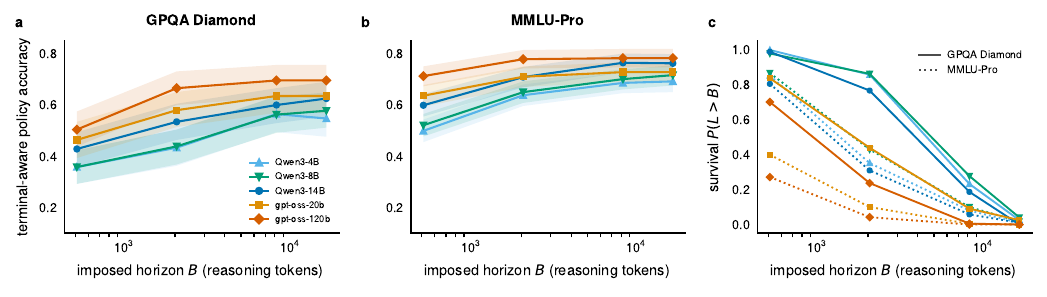}
\caption{\textbf{A returned-top-$20$ readout reproduces the qualitative fixed-stop pattern.}
Terminal-aware policy accuracy (\textbf{a}, GPQA Diamond; \textbf{b}, MMLU-Pro) and survival (\textbf{c}) for surprise-stop probes under a returned-top-$20$ readout, which takes the argmax over the option letters present in the returned top-$20$ log-probabilities and has incomplete option-letter coverage (\Cref{sec:methods}). The full-vocabulary candidate-logit readout (\Cref{fig:fixed-stop-validation}) is used for the main fixed-stop calibration and all primary multiple-choice estimates.}
\label{fig:truncation}
\end{figure}

\begin{figure}[!htbp]
\centering
\includegraphics[width=\textwidth]{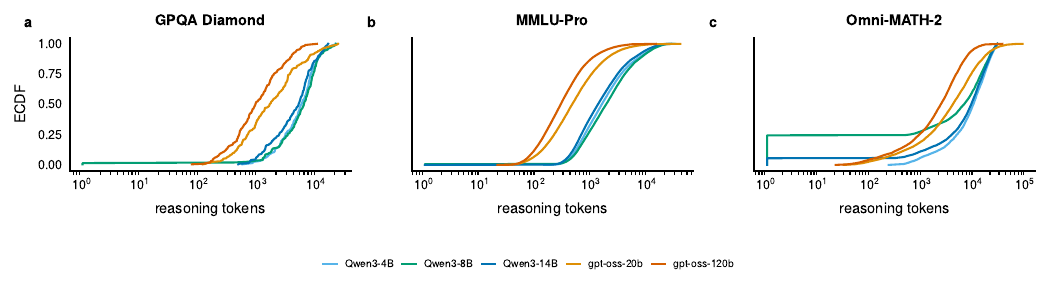}
\caption{\textbf{Natural reasoning-length distributions by model and benchmark.} Empirical cumulative distribution of reasoning-trace token length for the five models on GPQA Diamond, MMLU-Pro, and Omni-MATH-2. Natural length varies severalfold across models and benchmarks, which sets where the imposed-horizon curves of \Cref{fig:fixed-stop-validation} saturate. Panel \textbf{c} includes $980/4{,}011$ Qwen3-8B and $223/4{,}039$ Qwen3-14B responses that immediately close an empty \texttt{<think></think>} block. We count each empty block as one source-model token and exclude the following final-answer text from reasoning length.}
\label{fig:length-ecdf}
\end{figure}

\begin{figure}[!htbp]
\centering
\includegraphics[width=\textwidth]{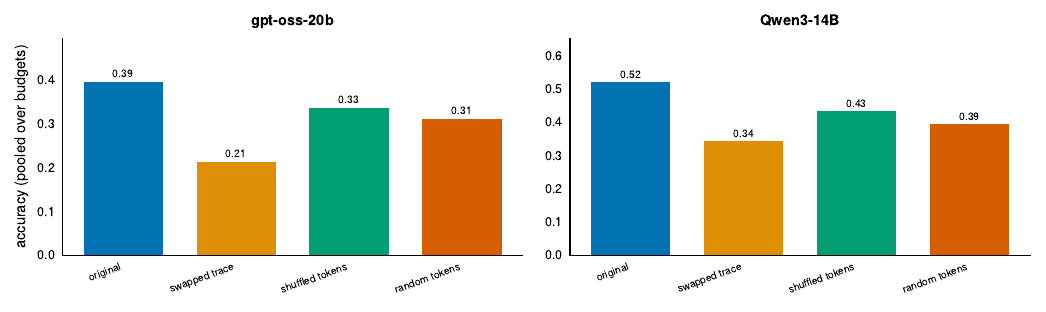}
\caption{\textbf{Truncation-recovered accuracy is instance-specific: the original prefix beats length-matched controls.} Accuracy of the surprise-stop probe under the original budget-$B$ prefix and three length-matched controls, pooled over budgets on the common four-arm item--budget cohort, for gpt-oss-20b and Qwen3-14B on GPQA Diamond and MMLU-Pro. The swapped-trace control (coherent reasoning from a different question) is the weakest, below the shuffled-token and random-token controls, and all sit below the original. Inference averages paired original-minus-control outcomes over budgets within item and uses an item-cluster bootstrap plus item-level sign-flip randomization test.}
\label{fig:controls}
\end{figure}

\begin{figure}[!htbp]
\centering
\includegraphics[width=\textwidth]{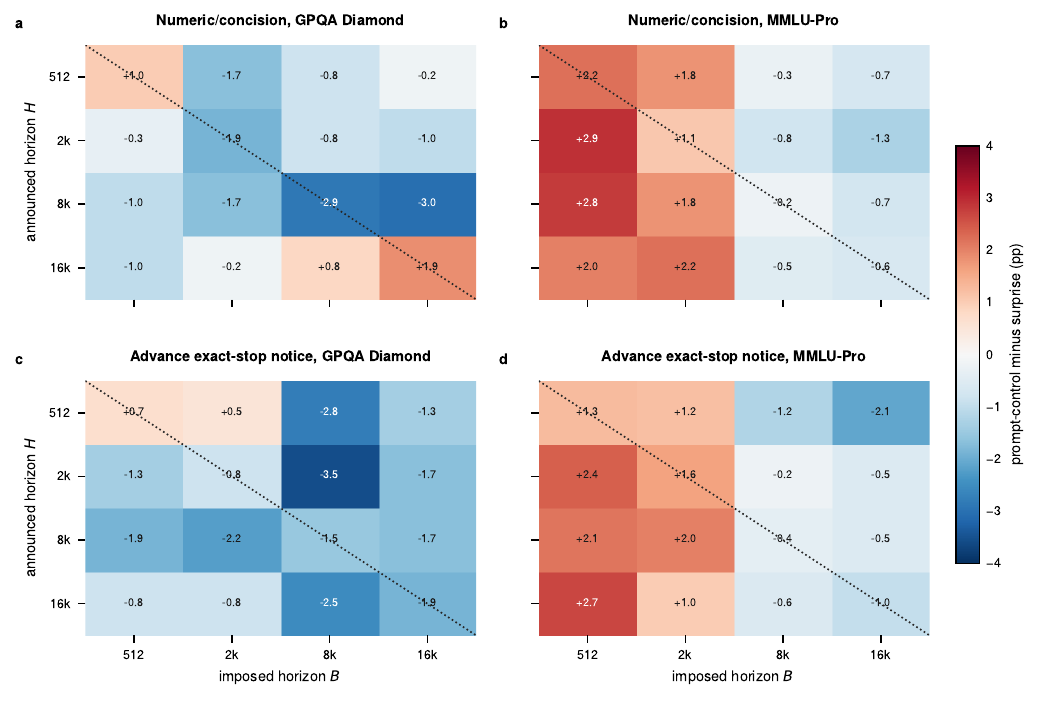}
\caption{\textbf{Crossing announced and imposed horizons shows no diagonal matching-number pattern.} Heatmaps show prompt-control minus surprise-stop accuracy deltas in percentage points for Qwen3-14B under the candidate-logit readout. Rows vary the announced horizon $H$ in the prompt, columns vary the imposed horizon $B$, and dotted diagonals mark $H=B$. \textbf{a,b}, Numeric/concision prompt for GPQA Diamond and MMLU-Pro. \textbf{c,d}, Advance exact-stop notice for the same benchmarks. MMLU-Pro positives concentrate at early imposed horizons, while GPQA remains small or negative across the grid. Completion token IDs and verified generation seeds are unavailable for these four-budget and GPQA cohorts; the paired MMLU-Pro experiment likewise finds no clear matching-number benefit at two imposed stops (\Cref{tab:qwen-mechanism-audited}).}
\label{fig:qwen-crossed-budget-stop}
\end{figure}

\begin{figure}[!htbp]
\centering
\includegraphics[width=\textwidth]{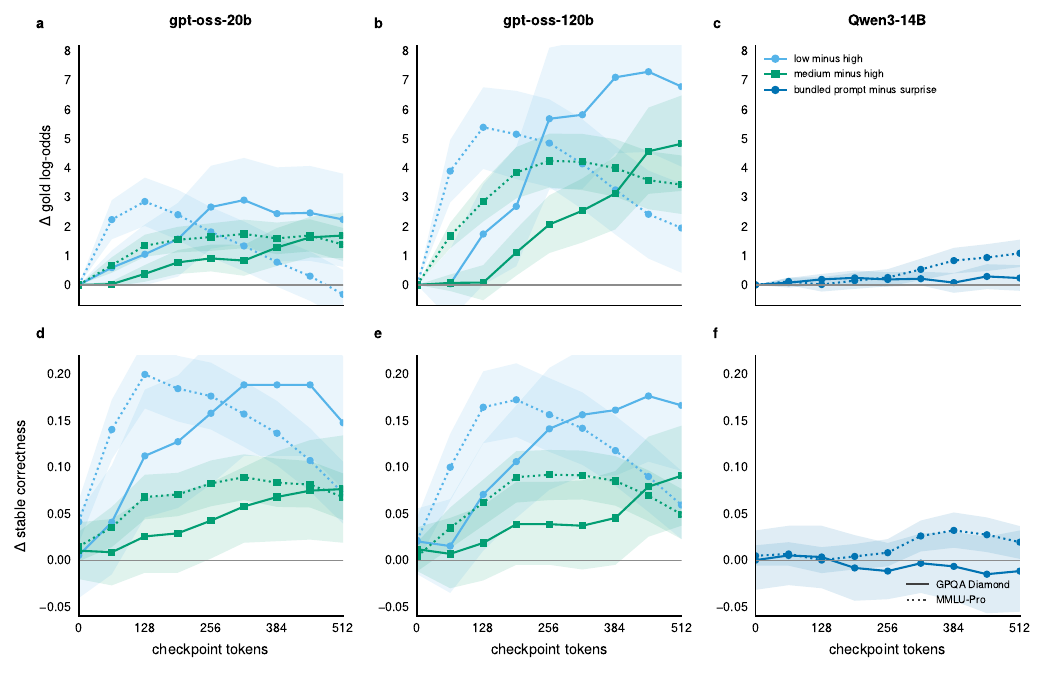}
\caption{\textbf{Shorter gpt-oss effort settings make answers available earlier under the candidate-logit readout.} Curves show paired condition differences at dense checkpoints through $B=512$; solid lines are GPQA Diamond, dotted lines are MMLU-Pro, and shaded bands are pointwise $95\%$ paired-bootstrap intervals over items. \textbf{a--c}, Difference in option-normalized gold-answer log-odds for gpt-oss-20b, gpt-oss-120b, and Qwen3-14B, computed from full-vocabulary next-token logits over every valid answer-token ID. \textbf{d--f}, Difference in retrospective stable correctness, where a checkpoint is counted correct only if all later checkpoints through $512$ remain correct. The gpt-oss contrasts compare low or medium source prefixes against high source prefixes under a medium-effort readout; Qwen compares bundled-prompt and surprise prefixes under a neutral budget-free closure.}
\label{fig:answer-availability}
\end{figure}

\begin{figure}[!htbp]
\centering
\includegraphics[width=\textwidth]{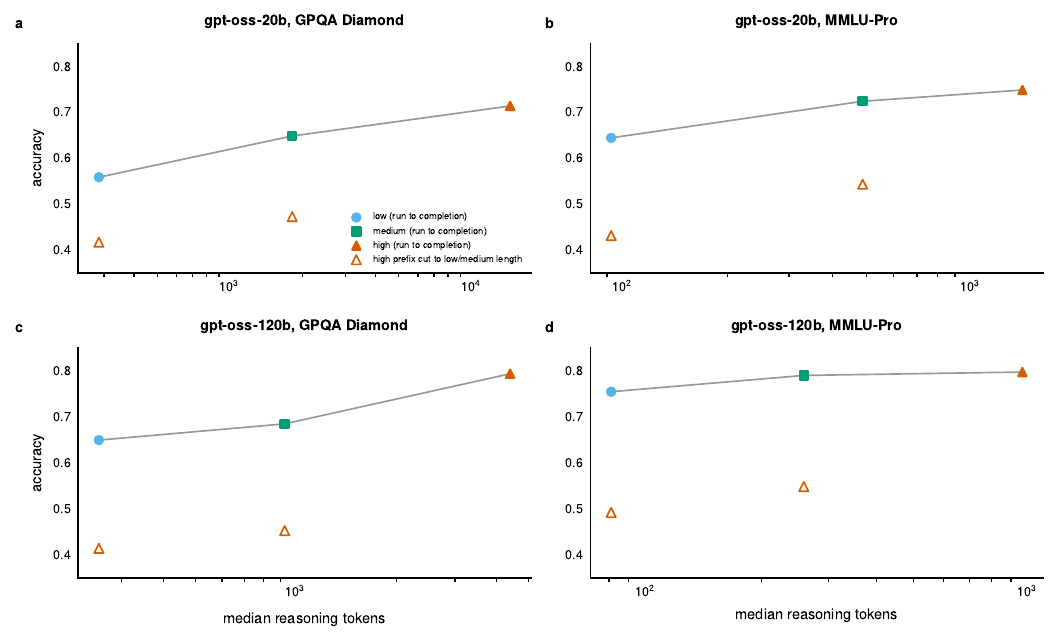}
\caption{\textbf{Completed gpt-oss effort settings trade accuracy against median trace length.} Results for gpt-oss-20b (\textbf{a,b}) and gpt-oss-120b (\textbf{c,d}) on GPQA Diamond and the stratified MMLU-Pro subset pool three generation replicates. When run to completion, the three effort settings define the observed relationship, while high-effort readouts at the itemwise low- or medium-effort terminal horizon sit below it; a completed high-effort reasoning block is replayed when it finishes before that horizon. Because the horizontal axis uses medians, this figure describes typical trace length rather than arithmetic-mean token use or expected API cost.}
\label{fig:effort-frontier}
\end{figure}

\begin{figure}[!htbp]
\centering
\includegraphics[width=\textwidth]{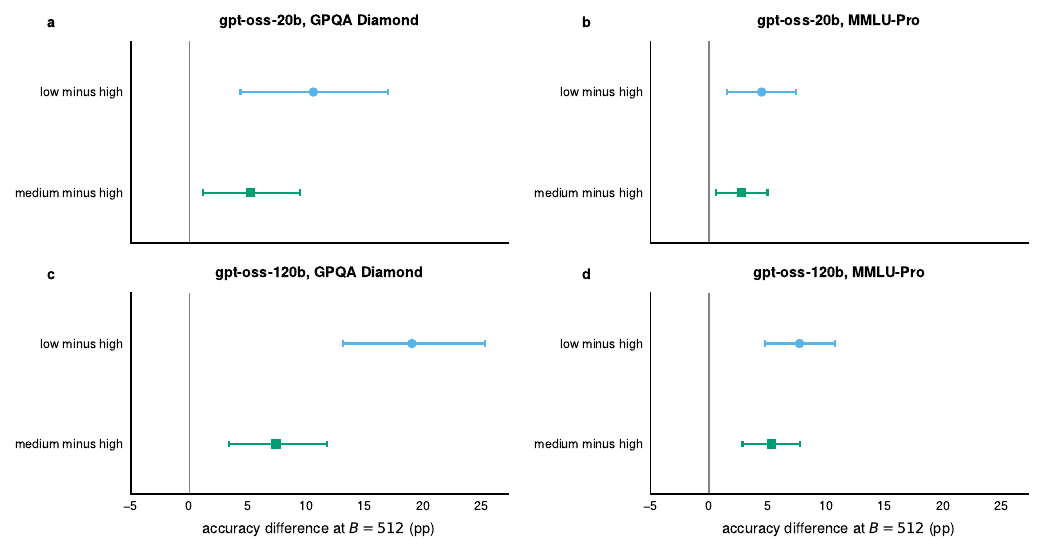}
\caption{\textbf{Lower gpt-oss effort remains more accurate at the same 512-token checkpoint.} Accuracy differences compare low or medium effort with high effort for gpt-oss-20b (\textbf{a,b}) and gpt-oss-120b (\textbf{c,d}) on GPQA Diamond and MMLU-Pro. Naturally stopped traces retain their final answer. Primary $95\%$ intervals average the three paired replicate outcomes within item and bootstrap items. Positive values favor the lower-effort setting.}
\label{fig:effort-fixed}
\end{figure}

\begin{figure}[t]
\centering
\includegraphics[width=\textwidth]{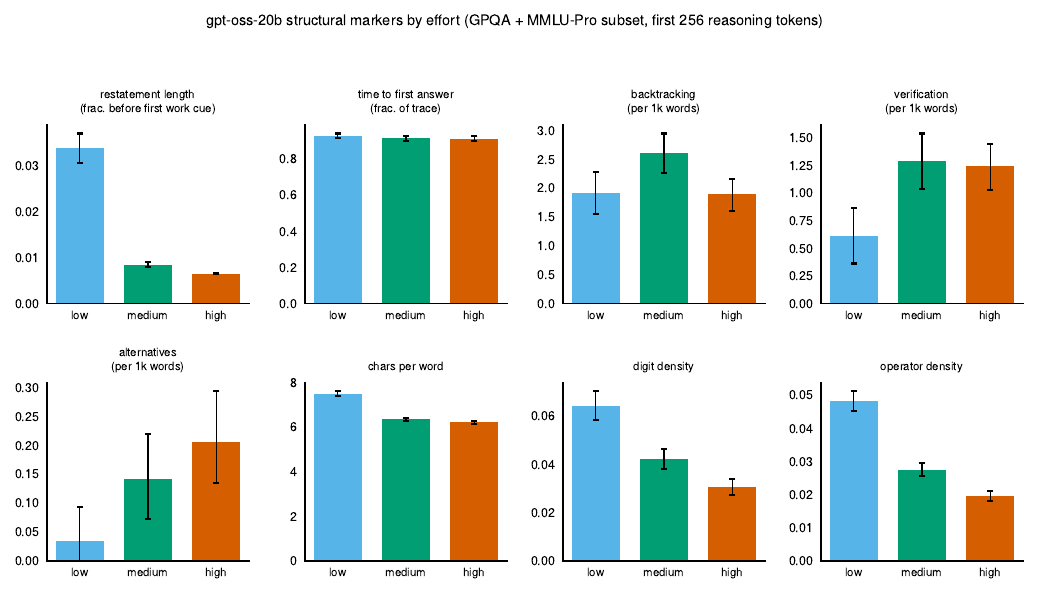}
\caption{\textbf{Structural markers of the first $256$ reasoning tokens by gpt-oss-20b effort setting.} Each panel is one length-normalized marker, averaged over GPQA Diamond and the stratified MMLU-Pro subset with $95\%$ bootstrap intervals over items, for low (blue), medium (green), and high (orange) effort. Digit density, operator density, and characters per word decrease monotonically from low to high effort, while the explicit alternative-strategy marker rate increases. Together, these surface measures show that shorter-effort traces allocate more of their opening tokens to calculation-like text and contain fewer explicit alternatives. The time-to-first-answer marker is near-saturated, and neither position-based marker validates against GPT-5 mini annotations; the backtracking cue rate agrees with the annotations but is not monotone across effort, and the verification marker does not validate in the agreement analysis on this subset. We interpret the density and alternative-strategy markers, which are objective or annotation-supported (\Cref{sec:methods}).}
\label{fig:structural}
\end{figure}

\begin{figure}[t]
\centering
\includegraphics[width=\textwidth]{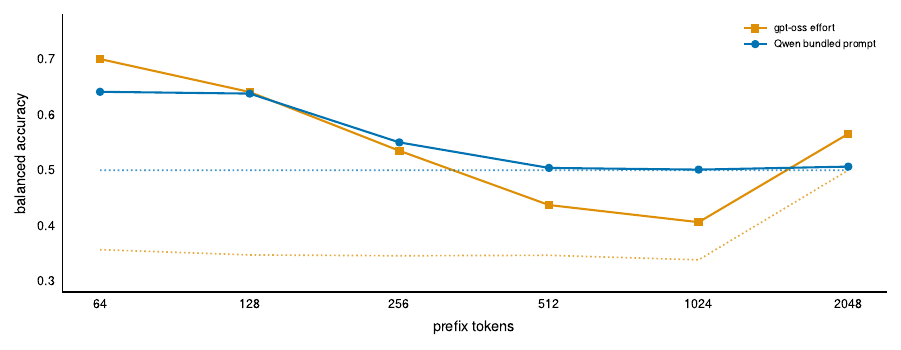}
\caption{\textbf{Condition classifiers distinguish gpt-oss effort more strongly than Qwen bundled prompting.} Balanced accuracy of a classifier predicting the condition from the first $k$ reasoning tokens, for gpt-oss effort (three-class) and Qwen3-14B bundled-prompt versus surprise-stop (two-class), with dotted label-shuffle baselines. Traces shorter than $k$ are excluded at that prefix length, so the classifier is a diagnostic of condition legibility among surviving traces rather than a causal estimate. The effort setting is strongly legible from the opening tokens; the bundled-prompt signal is weaker and decays to chance by $512$ tokens.}
\label{fig:prefix-classifiers}
\end{figure}

The appendix figures are organized by the question each analysis addresses. \Crefrange{fig:truncation}{fig:controls} characterize readout sensitivity and natural trace lengths and test whether recovered information depends on the question-specific prefix. \Crefrange{fig:qwen-crossed-budget-stop}{fig:answer-availability} test horizon matching and show when answer evidence becomes available. \Crefrange{fig:effort-frontier}{fig:effort-fixed} compare completed-trace and fixed-checkpoint gpt-oss accuracy, while \Crefrange{fig:structural}{fig:prefix-classifiers} characterize how effort and prompting alter early traces. The crossed-horizon Qwen cohorts in \Cref{fig:qwen-crossed-budget-stop} lack stored completion token IDs and verified generation seeds. They provide a horizon-pattern check and do not enter the token-exact effect estimates.

\clearpage
\section{Tables}

The appendix tables report the fine-grained estimates needed to check the primary matched-horizon results. \Crefrange{tab:main-estimands}{tab:exact-readout-agreement} define the estimands and interventions and validate the fixed-stop readouts, while \Crefrange{tab:reviewer-qwen-primary}{tab:reviewer-gptoss-strata} report the Qwen and gpt-oss estimates and sensitivity analyses. \Crefrange{tab:extended-answer-availability}{tab:extended-gptoss-score-tails} give dense trajectory, termination-stratum, answer-context, arm-level, and score-tail detail; \Cref{tab:open-ended-validation,tab:extended-scope} report single-rollout comparisons for Qwen3-8B and Omni-MATH-2 and characterize sampled trajectories. Qwen subset-replication and prompt-control results from cohorts without verified token IDs or seeds are in \Cref{tab:extended-qwen-scope-detail,tab:qwen-mmlu-b512-independent-subset,tab:qwen-prompt-controls}; readout validations, wrong-prefix continuation, activation readouts, and the prompt--effort interaction are in \Cref{tab:positive-control,tab:terminal-prefix-leak-audit,tab:rescue,tab:activation-readout,tab:gptoss-numeric-effort}. These tables broaden item or prompt coverage and address measurement or diagnostic questions; the primary matched-horizon claims rest on the token-exact experiments above.

\begin{table}[!htbp]
\centering
\caption{\textbf{We evaluate each condition at a common stopping point and separate completion from prefix quality.} The terminal-aware readout retains a natural answer after completion and probes an unfinished prefix in its source context. Common-context replay applies the same answer context to both conditions; both-active comparisons further restrict the analysis to jointly unfinished pairs. Behavioral continuations generate an unconstrained short answer from the stopped prefix.}
\label{tab:main-estimands}
\small
\setlength{\tabcolsep}{4pt}
\begin{tabular}{L{0.21\textwidth} L{0.23\textwidth} L{0.29\textwidth} L{0.17\textwidth}}
\toprule
Comparison & Where the comparison stops & How the answer is measured & Main result \\
\midrule
Qwen textual treatment: visible numeric/concision prompt ($v$) vs ordinary generation ($h$) & Common imposed horizon $B$; retain either arm's natural answer if it completed before $B$ & Terminal-aware policy-context readout; $\tau_{Q,\mathrm{pol}}(B)=\mathbb{E}[Y^{v,T_{\mathrm{pol}}}(B)-Y^{h,T_{\mathrm{pol}}}(B)]$ & Shorter traces; no broad accuracy gain at the same imposed horizon \\
Qwen prefix diagnostics: tested instruction ($a$) vs ordinary generation ($h$) & Common imposed horizon $B$; replay both prefixes under the ordinary base prompt & Correct completion by $B$, common-context terminal replay, both-active contrast, and four-stratum decomposition & Separates stopping and commitment from differences between unfinished prefixes \\
Qwen literal cap: visible numeric/concision prompt vs hidden horizon & Common external cap $B\in\{512,2048\}$; both arms then receive an answer continuation & Greedy generated-answer continuation; $\mathbb{E}[Y^{v,C}(B)-Y^{h,C}(B)]$ & Benchmark split checked without option normalization \\
gpt-oss effort: completed $c\in\{\mathrm{low},\mathrm{medium}\}$ vs high effort & Itemwise lower-effort terminal length $L_{ir}^{c}$; replay every reconstructable high prefix, including early completions & All-replayable common-context probe; $\tau_{G,R}^{c}=\mathbb{E}[D^c\mid R^c=1]$, with worst-case bounds & Lower effort reaches correct terminal answers earlier \\
gpt-oss active diagnostic & The same matched pairs restricted to high effort still active at $L_{ir}^{c}$ & Common-context candidate-logit probe & Separates the policy-cost result from early high-effort completion \\
gpt-oss continuation check: the same matched prefixes & The same all-replayable $L_{ir}^{c}$ pairs & Greedy unconstrained final-channel answer; $\mathbb{E}[Y^{c,C}-Y^{\mathrm{high},C}\mid R^c=1]$ & Tests whether the ordering survives a generated-answer continuation \\
\bottomrule
\end{tabular}
\end{table}

\begin{table}[!htbp]
\centering
\caption{\textbf{Verbatim generation and readout interventions.} Braces indicate the interpolated token budget.}
\label{tab:prompt-texts}
\small
\begin{adjustbox}{max width=\textwidth}
\begin{tabular}{L{2.8cm} L{2.2cm} L{8.0cm}}
\toprule
Condition & Message role & Exact text or template \\
\midrule
Qwen numeric/concision prompt & system, appended to base instruction & \texttt{You have at most \{B\} tokens inside the <think>...</think> reasoning block. Use a strategy appropriate for this budget. When you have enough evidence, close the reasoning block and answer with only the final answer.} \\
Qwen limited-budget control & system, appended to base instruction & \texttt{You have a limited token budget inside the <think>...</think> reasoning block. Use a strategy appropriate for this limit. When you have enough evidence, close the reasoning block and answer with only the final answer.} \\
Qwen number-only control & system, appended to base instruction & \texttt{You have at most \{B\} tokens inside the <think>...</think> reasoning block.} \\
Qwen advance exact-stop control & system, appended to base instruction & \texttt{Your <think>...</think> reasoning block will be stopped after exactly \{B\} tokens.} \\
Qwen approximate-budget control & system, appended to base instruction & \texttt{Aim to use approximately \{B\} tokens inside the <think>...</think> reasoning block.} \\
Qwen irrelevant-number control & system, appended to base instruction & \texttt{The control number for this problem is \{B\}. Do not treat this number as a token budget.} \\
Qwen concise/early-answer control & system, appended to base instruction & \texttt{Use a very concise reasoning strategy. Prioritize reaching a usable answer quickly.} \\
Qwen thorough semantic control & system, appended to base instruction & \texttt{Use a thorough reasoning strategy. Explore alternatives and check your answer carefully before finalizing.} \\
Qwen limited-time closure & continuation & \texttt{Considering the limited time by the user, I have to give the solution based on the thinking directly now. </think>} \\
Qwen paraphrased closure & continuation & \texttt{Based on my reasoning so far, my answer is: </think>} \\
Qwen neutral closure & continuation & \texttt{</think>} \\
Qwen budget-present readout & system, answer-time only & \texttt{You have at most \{B\} tokens inside the <think>...</think> reasoning block. Use a strategy appropriate for this budget. When you have enough evidence, close the reasoning block and answer with only the final answer.} \\
Qwen answer instruction & system & \texttt{Solve the following problem. Please make sure that your response only consists of a single letter corresponding to the correct answer choice. Do not include anything else in your final response.} \\
gpt-oss effort & chat-template kwarg/system scaffold & \texttt{reasoning\_effort = low | medium | high}, rendered by the snapshot chat template as \texttt{Reasoning: low|medium|high} in the system message before the developer/user messages \\
\bottomrule
\end{tabular}
\end{adjustbox}
\end{table}

\begin{table*}[t]
\centering
\caption{\textbf{Fixed-stop calibration separates completed answers from the selected set of surviving traces.} The full-vocabulary candidate-logit readout is evaluated at $B\in\{0,512,2{,}048,8{,}192,16{,}384\}$. Survival is $P(L>B)$, survivor-only accuracy conditions on $L>B$, and terminal-aware accuracy retains the natural answer after completion. The natural full-trace reference uses the same cohort. Rate columns report percentages with $95\%$ item-bootstrap intervals. Accuracy omits missing parsed outcomes; $n$ retains the full cohort.}
\label{tab:calibration-survival}
\scriptsize
\setlength{\tabcolsep}{3pt}
\begin{adjustbox}{max width=\textwidth}
\begin{tabular}{L{1.65cm} L{1.65cm} r r r r r r}
\toprule
Model & Benchmark & $B$ & $n$ & Survival (\%) & Survivor-only accuracy (\%) & Terminal-aware policy accuracy (\%) & Natural full-trace accuracy (\%) \\
\midrule
Qwen3-4B & GPQA Diamond & 0 & 198 & $100.0$ $[100.0,\,100.0]$ & $31.8$ $[25.3,\,38.4]$ & $31.8$ $[25.3,\,38.4]$ & $55.6$ $[48.5,\,62.2]$ \\
Qwen3-4B & GPQA Diamond & 512 & 198 & $100.0$ $[100.0,\,100.0]$ & $34.8$ $[28.3,\,41.4]$ & $34.8$ $[28.3,\,41.4]$ & $55.6$ $[48.5,\,62.2]$ \\
Qwen3-4B & GPQA Diamond & 2{,}048 & 198 & $85.9$ $[80.8,\,90.4]$ & $37.6$ $[30.6,\,44.7]$ & $43.4$ $[36.9,\,50.5]$ & $55.6$ $[48.5,\,62.2]$ \\
Qwen3-4B & GPQA Diamond & 8{,}192 & 198 & $23.2$ $[17.7,\,29.3]$ & $50.0$ $[34.8,\,65.2]$ & $56.1$ $[49.5,\,62.6]$ & $55.6$ $[48.5,\,62.2]$ \\
Qwen3-4B & GPQA Diamond & 16{,}384 & 198 & $2.5$ $[0.5,\,5.1]$ & $40.0$ $[0.0,\,80.0]$ & $54.8$ $[47.7,\,61.9]$ & $55.6$ $[48.5,\,62.2]$ \\
\addlinespace
Qwen3-4B & MMLU-Pro & 0 & 500 & $100.0$ $[100.0,\,100.0]$ & $39.0$ $[34.8,\,43.2]$ & $39.0$ $[34.8,\,43.2]$ & $69.5$ $[65.5,\,73.5]$ \\
Qwen3-4B & MMLU-Pro & 512 & 500 & $85.4$ $[82.2,\,88.4]$ & $43.3$ $[38.9,\,48.0]$ & $49.4$ $[45.2,\,53.8]$ & $69.5$ $[65.5,\,73.5]$ \\
Qwen3-4B & MMLU-Pro & 2{,}048 & 500 & $35.2$ $[31.0,\,39.4]$ & $44.3$ $[36.9,\,51.7]$ & $64.3$ $[60.1,\,68.5]$ & $69.5$ $[65.5,\,73.5]$ \\
Qwen3-4B & MMLU-Pro & 8{,}192 & 500 & $8.0$ $[5.6,\,10.4]$ & $27.5$ $[15.0,\,42.5]$ & $68.3$ $[64.3,\,72.3]$ & $69.5$ $[65.5,\,73.5]$ \\
Qwen3-4B & MMLU-Pro & 16{,}384 & 500 & $1.0$ $[0.2,\,2.0]$ & $0.0$ $[0.0,\,0.0]$ & $69.3$ $[65.3,\,73.3]$ & $69.5$ $[65.5,\,73.5]$ \\
\addlinespace
Qwen3-8B & GPQA Diamond & 0 & 198 & $100.0$ $[100.0,\,100.0]$ & $30.8$ $[24.2,\,37.4]$ & $30.8$ $[24.2,\,37.4]$ & $57.9$ $[50.8,\,64.5]$ \\
Qwen3-8B & GPQA Diamond & 512 & 198 & $98.0$ $[96.0,\,99.5]$ & $35.6$ $[28.9,\,42.3]$ & $36.9$ $[30.3,\,43.4]$ & $57.9$ $[50.8,\,64.5]$ \\
Qwen3-8B & GPQA Diamond & 2{,}048 & 198 & $86.4$ $[81.3,\,90.9]$ & $35.7$ $[28.7,\,42.7]$ & $42.9$ $[35.9,\,49.5]$ & $57.9$ $[50.8,\,64.5]$ \\
Qwen3-8B & GPQA Diamond & 8{,}192 & 198 & $27.8$ $[21.7,\,34.3]$ & $30.9$ $[18.2,\,43.6]$ & $56.9$ $[49.7,\,63.5]$ & $57.9$ $[50.8,\,64.5]$ \\
Qwen3-8B & GPQA Diamond & 16{,}384 & 198 & $4.0$ $[1.5,\,7.1]$ & $37.5$ $[0.0,\,75.0]$ & $57.9$ $[50.8,\,64.5]$ & $57.9$ $[50.8,\,64.5]$ \\
\addlinespace
Qwen3-8B & MMLU-Pro & 0 & 500 & $100.0$ $[100.0,\,100.0]$ & $43.0$ $[38.8,\,47.4]$ & $43.0$ $[38.8,\,47.4]$ & $72.2$ $[68.1,\,76.0]$ \\
Qwen3-8B & MMLU-Pro & 512 & 500 & $86.8$ $[83.8,\,89.8]$ & $46.5$ $[41.7,\,51.2]$ & $52.6$ $[48.2,\,57.0]$ & $72.2$ $[68.1,\,76.0]$ \\
Qwen3-8B & MMLU-Pro & 2{,}048 & 500 & $43.0$ $[38.8,\,47.4]$ & $43.7$ $[37.2,\,50.2]$ & $65.1$ $[60.9,\,69.3]$ & $72.2$ $[68.1,\,76.0]$ \\
Qwen3-8B & MMLU-Pro & 8{,}192 & 500 & $10.0$ $[7.4,\,12.8]$ & $34.0$ $[22.0,\,48.0]$ & $70.2$ $[66.2,\,74.2]$ & $72.2$ $[68.1,\,76.0]$ \\
Qwen3-8B & MMLU-Pro & 16{,}384 & 500 & $1.8$ $[0.8,\,3.0]$ & $33.3$ $[0.0,\,66.7]$ & $71.8$ $[67.7,\,75.8]$ & $72.2$ $[68.1,\,76.0]$ \\
\addlinespace
Qwen3-14B & GPQA Diamond & 0 & 198 & $100.0$ $[100.0,\,100.0]$ & $37.4$ $[30.8,\,43.9]$ & $37.4$ $[30.8,\,43.9]$ & $62.6$ $[55.6,\,69.2]$ \\
Qwen3-14B & GPQA Diamond & 512 & 198 & $99.0$ $[97.5,\,100.0]$ & $41.8$ $[35.2,\,49.0]$ & $42.4$ $[35.8,\,49.5]$ & $62.6$ $[55.6,\,69.2]$ \\
Qwen3-14B & GPQA Diamond & 2{,}048 & 198 & $76.8$ $[70.7,\,82.3]$ & $44.1$ $[36.2,\,52.0]$ & $53.5$ $[46.5,\,60.6]$ & $62.6$ $[55.6,\,69.2]$ \\
Qwen3-14B & GPQA Diamond & 8{,}192 & 198 & $18.7$ $[13.6,\,24.2]$ & $29.7$ $[16.2,\,45.9]$ & $59.1$ $[52.5,\,66.2]$ & $62.6$ $[55.6,\,69.2]$ \\
Qwen3-14B & GPQA Diamond & 16{,}384 & 198 & $0.0$ $[0.0,\,0.0]$ & \textemdash & $62.6$ $[56.1,\,69.2]$ & $62.6$ $[55.6,\,69.2]$ \\
\addlinespace
Qwen3-14B & MMLU-Pro & 0 & 500 & $100.0$ $[100.0,\,100.0]$ & $50.4$ $[46.0,\,54.8]$ & $50.4$ $[46.0,\,54.8]$ & $76.8$ $[73.0,\,80.4]$ \\
Qwen3-14B & MMLU-Pro & 512 & 500 & $80.6$ $[77.0,\,84.0]$ & $52.4$ $[47.4,\,57.1]$ & $60.0$ $[55.8,\,64.4]$ & $76.8$ $[73.0,\,80.4]$ \\
Qwen3-14B & MMLU-Pro & 2{,}048 & 500 & $31.0$ $[27.0,\,35.0]$ & $50.3$ $[42.6,\,58.1]$ & $70.8$ $[66.8,\,74.8]$ & $76.8$ $[73.0,\,80.4]$ \\
Qwen3-14B & MMLU-Pro & 8{,}192 & 500 & $5.8$ $[3.8,\,8.0]$ & $51.7$ $[34.5,\,69.0]$ & $76.6$ $[72.8,\,80.2]$ & $76.8$ $[73.0,\,80.4]$ \\
Qwen3-14B & MMLU-Pro & 16{,}384 & 500 & $1.0$ $[0.2,\,2.0]$ & $0.0$ $[0.0,\,0.0]$ & $76.4$ $[72.8,\,80.2]$ & $76.8$ $[73.0,\,80.4]$ \\
\addlinespace
gpt-oss-20b & GPQA Diamond & 0 & 198 & $100.0$ $[100.0,\,100.0]$ & $40.9$ $[33.8,\,48.0]$ & $40.9$ $[33.8,\,48.0]$ & $62.6$ $[55.6,\,69.2]$ \\
gpt-oss-20b & GPQA Diamond & 512 & 198 & $83.8$ $[78.3,\,88.9]$ & $42.2$ $[34.3,\,49.4]$ & $50.0$ $[42.9,\,57.1]$ & $62.6$ $[55.6,\,69.2]$ \\
gpt-oss-20b & GPQA Diamond & 2{,}048 & 198 & $43.9$ $[37.4,\,51.0]$ & $36.8$ $[26.4,\,47.1]$ & $60.6$ $[53.5,\,67.2]$ & $62.6$ $[55.6,\,69.2]$ \\
gpt-oss-20b & GPQA Diamond & 8{,}192 & 198 & $9.1$ $[5.6,\,13.1]$ & $38.9$ $[16.7,\,61.1]$ & $63.6$ $[56.6,\,70.2]$ & $62.6$ $[55.6,\,69.2]$ \\
gpt-oss-20b & GPQA Diamond & 16{,}384 & 198 & $2.5$ $[0.5,\,5.1]$ & $60.0$ $[20.0,\,100.0]$ & $63.6$ $[57.1,\,70.2]$ & $62.6$ $[55.6,\,69.2]$ \\
\addlinespace
gpt-oss-20b & MMLU-Pro & 0 & 500 & $100.0$ $[100.0,\,100.0]$ & $43.4$ $[39.2,\,47.8]$ & $43.4$ $[39.2,\,47.8]$ & $72.9$ $[68.9,\,76.8]$ \\
gpt-oss-20b & MMLU-Pro & 512 & 500 & $40.0$ $[35.8,\,44.2]$ & $35.5$ $[29.0,\,42.0]$ & $64.1$ $[59.9,\,68.3]$ & $72.9$ $[68.9,\,76.8]$ \\
gpt-oss-20b & MMLU-Pro & 2{,}048 & 500 & $10.0$ $[7.4,\,12.6]$ & $24.0$ $[12.0,\,36.0]$ & $71.1$ $[67.1,\,75.2]$ & $72.9$ $[68.9,\,76.8]$ \\
gpt-oss-20b & MMLU-Pro & 8{,}192 & 500 & $0.4$ $[0.0,\,1.0]$ & $0.0$ $[0.0,\,0.0]$ & $72.9$ $[69.1,\,76.8]$ & $72.9$ $[68.9,\,76.8]$ \\
gpt-oss-20b & MMLU-Pro & 16{,}384 & 500 & $0.0$ $[0.0,\,0.0]$ & \textemdash & $72.9$ $[69.1,\,76.8]$ & $72.9$ $[68.9,\,76.8]$ \\
\addlinespace
gpt-oss-120b & GPQA Diamond & 0 & 198 & $100.0$ $[100.0,\,100.0]$ & $38.9$ $[32.3,\,46.0]$ & $38.9$ $[32.3,\,46.0]$ & $69.7$ $[63.1,\,75.8]$ \\
gpt-oss-120b & GPQA Diamond & 512 & 198 & $70.2$ $[63.6,\,76.8]$ & $36.0$ $[28.1,\,43.9]$ & $51.5$ $[44.4,\,58.6]$ & $69.7$ $[63.1,\,75.8]$ \\
gpt-oss-120b & GPQA Diamond & 2{,}048 & 198 & $23.7$ $[18.2,\,29.8]$ & $40.4$ $[25.5,\,53.2]$ & $67.7$ $[61.1,\,74.2]$ & $69.7$ $[63.1,\,75.8]$ \\
gpt-oss-120b & GPQA Diamond & 8{,}192 & 198 & $0.5$ $[0.0,\,1.5]$ & $0.0$ $[0.0,\,0.0]$ & $69.7$ $[63.1,\,75.8]$ & $69.7$ $[63.1,\,75.8]$ \\
gpt-oss-120b & GPQA Diamond & 16{,}384 & 198 & $0.0$ $[0.0,\,0.0]$ & \textemdash & $69.7$ $[63.1,\,75.8]$ & $69.7$ $[63.1,\,75.8]$ \\
\addlinespace
gpt-oss-120b & MMLU-Pro & 0 & 500 & $100.0$ $[100.0,\,100.0]$ & $48.4$ $[44.0,\,52.8]$ & $48.4$ $[44.0,\,52.8]$ & $78.4$ $[74.8,\,81.8]$ \\
gpt-oss-120b & MMLU-Pro & 512 & 500 & $27.2$ $[23.4,\,31.2]$ & $40.4$ $[32.4,\,48.5]$ & $71.2$ $[67.2,\,75.2]$ & $78.4$ $[74.8,\,81.8]$ \\
gpt-oss-120b & MMLU-Pro & 2{,}048 & 500 & $4.2$ $[2.6,\,6.0]$ & $42.9$ $[23.8,\,66.7]$ & $78.0$ $[74.4,\,81.6]$ & $78.4$ $[74.8,\,81.8]$ \\
gpt-oss-120b & MMLU-Pro & 8{,}192 & 500 & $0.0$ $[0.0,\,0.0]$ & \textemdash & $78.4$ $[74.8,\,82.0]$ & $78.4$ $[74.8,\,81.8]$ \\
gpt-oss-120b & MMLU-Pro & 16{,}384 & 500 & $0.0$ $[0.0,\,0.0]$ & \textemdash & $78.4$ $[74.8,\,82.0]$ & $78.4$ $[74.8,\,81.8]$ \\
\bottomrule
\end{tabular}
\end{adjustbox}
\end{table*}

\begin{table}[t]
\centering
\caption{\textbf{Completed-prefix readouts agree with natural answers in $98.3$--$100.0\%$ of cases.} Each row compares the candidate-logit answer from a replayed completed reasoning prefix with the parsed natural answer from the same trace. This separate cohort validates terminal replay; it does not assess coverage of the three-replicate token-exact analysis.}
\label{tab:exact-readout-agreement}
\small
\begin{adjustbox}{max width=\textwidth}
\begin{tabular}{L{2.1cm} L{2.1cm} L{3.7cm} r r r}
\toprule
Model & Benchmark & Source policy & Compared & Agree & Agreement \\
\midrule
gpt-oss-120b & GPQA Diamond & gpt-oss low effort & $198$ & $198$ & 100.0\% \\
gpt-oss-120b & GPQA Diamond & gpt-oss medium effort & $198$ & $198$ & 100.0\% \\
gpt-oss-120b & MMLU-Pro & gpt-oss low effort & $499$ & $499$ & 100.0\% \\
gpt-oss-120b & MMLU-Pro & gpt-oss medium effort & $500$ & $500$ & 100.0\% \\
gpt-oss-20b & GPQA Diamond & gpt-oss low effort & $196$ & $196$ & 100.0\% \\
gpt-oss-20b & GPQA Diamond & gpt-oss medium effort & $196$ & $196$ & 100.0\% \\
gpt-oss-20b & MMLU-Pro & gpt-oss low effort & $496$ & $494$ & 99.6\% \\
gpt-oss-20b & MMLU-Pro & gpt-oss medium effort & $499$ & $499$ & 100.0\% \\
Qwen3-14B & GPQA Diamond & Qwen numeric/concision prompt & $1{,}317$ & $1{,}315$ & 99.8\% \\
Qwen3-14B & GPQA Diamond & Qwen surprise & $588$ & $586$ & 99.7\% \\
Qwen3-14B & MMLU-Pro & Qwen numeric/concision prompt & $4{,}451$ & $4{,}447$ & 99.9\% \\
Qwen3-14B & MMLU-Pro & Qwen surprise & $1{,}489$ & $1{,}463$ & 98.3\% \\
\bottomrule
\end{tabular}
\end{adjustbox}
\end{table}

\begin{table}[t]
\centering
\caption{\textbf{Open-ended continuations show no consistent Qwen prompt gain and a gpt-oss matched-horizon advantage.} GPT-5 mini judges free-form Omni-MATH-2 answers. Qwen rows compare numeric/concision-prompt and surprise-stop prefixes at the same imposed horizon on a $200$-item sample; each row includes only item--horizon comparisons for which both answers were judged. gpt-oss rows compare a completed lower-effort reasoning block with high effort at the itemwise lower-effort terminal horizon, replaying a completed high-effort reasoning block when it finishes before that horizon. Every reasoning block is read under a common medium-effort final-answer context. The one-rollout, difficulty-stratified $500$-item draw yields model-specific paired cohorts of 475 items for gpt-oss-20b and 490 for gpt-oss-120b. The analysis includes all $15{,}440$ judged gpt-oss answers across the matched-horizon and fixed-checkpoint comparisons. Intervals bootstrap items.}
\label{tab:open-ended-validation}
\small
\begin{adjustbox}{max width=\textwidth}
\begin{tabular}{L{1.8cm} L{1.8cm} L{3.3cm} r r r r L{2.2cm}}
\toprule
Comparison & Model & Contrast & Stop & $n$ & Treated/lower acc. & Comparator acc. & $\Delta$ [95\% CI] \\
\midrule
Qwen prompt & Qwen3-14B & prompt minus surprise & $512$ & $170$ & $22.9\%$ & $19.4\%$ & $+3.5$ $[-1.2,+8.2]$ \\
Qwen prompt & Qwen3-14B & prompt minus surprise & $2{,}048$ & $150$ & $33.3\%$ & $31.3\%$ & $+2.0$ $[-3.3,+7.3]$ \\
Qwen prompt & Qwen3-14B & prompt minus surprise & $8{,}192$ & $73$ & $26.0\%$ & $30.1\%$ & $-4.1$ $[-12.3,+4.1]$ \\
\addlinespace
gpt-oss effort & gpt-oss-20b & low minus high prefix & matched & $475$ & $47.8\%$ & $14.3\%$ & $+33.5$ $[+28.8,+37.9]$ \\
gpt-oss effort & gpt-oss-20b & medium minus high prefix & matched & $475$ & $61.5\%$ & $27.6\%$ & $+33.9$ $[+28.8,+38.9]$ \\
gpt-oss effort & gpt-oss-120b & low minus high prefix & matched & $490$ & $61.2\%$ & $19.2\%$ & $+42.0$ $[+37.3,+46.7]$ \\
gpt-oss effort & gpt-oss-120b & medium minus high prefix & matched & $490$ & $70.6\%$ & $33.1\%$ & $+37.6$ $[+32.4,+42.7]$ \\
\bottomrule
\end{tabular}
\end{adjustbox}
\end{table}

\begin{table*}[t]
\centering
\caption{\textbf{The numeric/concision prompt does not consistently improve Qwen accuracy at the same horizon.} Panel A compares the numeric/concision (visible-horizon) and ordinary (hidden-horizon) conditions at the same imposed horizon using token-exact replay and the terminal-aware policy readout for all $198$ GPQA Diamond and $500$ MMLU-Pro items across three generation replicates; $n$ gives pointwise/complete-curve item counts. $\Delta$ is visible minus hidden accuracy; intervals and simultaneous bands average replicates within item before bootstrapping items. AUC integrates $\Delta$ over $\log_2 B$. Panel B applies the literal cap-then-answer continuation and reports parse coverage as visible/hidden.}
\label{tab:reviewer-qwen-primary}
\small
\begin{adjustbox}{max width=\textwidth}
\begin{tabular}{L{2.0cm} L{2.0cm} r r r r r L{2.8cm} L{2.3cm} L{2.8cm}}
\toprule
\multicolumn{10}{l}{\textbf{Panel A: terminal-aware policy readout}} \\
Model & Benchmark & $B$ & $n$ & Replicates & Visible acc. & Hidden acc. & $\Delta$ [95\% CI] & Simultaneous 95\% band & AUC [95\% CI] \\
\midrule
Qwen3-14B & GPQA Diamond & 512 & 198/198 & 3 & $41.8\%$ & $42.3\%$ & $-0.5$ $[-2.5,\,+1.3]$ & $[-4.7,\,+3.7]$ & $+0.5$ $[-1.6,\,+2.7]$ \\
Qwen3-14B & GPQA Diamond & 2{,}048 & 198/198 & 3 & $53.2\%$ & $52.5\%$ & $+0.7$ $[-3.0,\,+4.4]$ & $[-3.5,\,+4.9]$ & \textemdash \\
Qwen3-14B & GPQA Diamond & 8{,}192 & 198/198 & 3 & $60.3\%$ & $59.6\%$ & $+0.7$ $[-2.4,\,+3.9]$ & $[-3.5,\,+4.9]$ & \textemdash \\
Qwen3-14B & GPQA Diamond & 16{,}384 & 198/198 & 3 & $62.0\%$ & $60.6\%$ & $+1.3$ $[-1.9,\,+4.5]$ & $[-2.9,\,+5.6]$ & \textemdash \\
Qwen3-14B & MMLU-Pro & 512 & 500/500 & 3 & $60.5\%$ & $58.5\%$ & $+2.0$ $[+0.4,\,+3.7]$ & $[+0.1,\,+3.9]$ & $+0.4$ $[-0.7,\,+1.5]$ \\
Qwen3-14B & MMLU-Pro & 2{,}048 & 500/500 & 3 & $71.7\%$ & $71.0\%$ & $+0.7$ $[-0.9,\,+2.1]$ & $[-1.3,\,+2.6]$ & \textemdash \\
Qwen3-14B & MMLU-Pro & 8{,}192 & 500/500 & 3 & $75.1\%$ & $75.9\%$ & $-0.9$ $[-2.5,\,+0.7]$ & $[-2.8,\,+1.1]$ & \textemdash \\
Qwen3-14B & MMLU-Pro & 16{,}384 & 500/500 & 3 & $75.7\%$ & $75.6\%$ & $+0.1$ $[-1.2,\,+1.3]$ & $[-1.9,\,+2.0]$ & \textemdash \\
\addlinespace
\multicolumn{10}{l}{\textbf{Panel B: literal cap-then-answer continuation}} \\
Model & Benchmark & $B$ & $n$ & Replicates & Visible acc. & Hidden acc. & $\Delta$ [95\% CI] & Parse coverage V/H & Policy-readout $\Delta$ \\
\midrule
Qwen3-14B & GPQA Diamond & 512 & 198 & 3 & $39.9\%$ & $41.8\%$ & $-1.9$ $[-4.9,\,+1.0]$ & $100.0\%$/$100.0\%$ & $-0.5$ \\
Qwen3-14B & GPQA Diamond & 2{,}048 & 198 & 3 & $53.9\%$ & $53.4\%$ & $+0.5$ $[-2.7,\,+3.7]$ & $99.8\%$/$99.8\%$ & $+0.7$ \\
Qwen3-14B & MMLU-Pro & 512 & 500 & 3 & $60.7\%$ & $58.7\%$ & $+2.0$ $[+0.2,\,+3.9]$ & $97.1\%$/$96.7\%$ & $+2.0$ \\
Qwen3-14B & MMLU-Pro & 2{,}048 & 500 & 3 & $71.4\%$ & $71.3\%$ & $+0.1$ $[-1.3,\,+1.5]$ & $99.5\%$/$97.5\%$ & $+0.7$ \\
\bottomrule
\end{tabular}
\end{adjustbox}
\end{table*}

\begin{table}[t]
\centering
\caption{\textbf{At the 512-token stop, the concise/early-answer instruction gives the largest accuracy gain; matching the stated number to the stop shows no clear benefit.} \textbf{a}, Qwen3-14B results on the $500$-item MMLU-Pro subset with eight paired replicates per item. ``Finished'' is the share of runs that produce a natural answer by $B$; otherwise, accuracy uses the exact stored prefix and standardized forced-answer readout. Differences are relative to no advance notice, with nested item--replicate bootstrap intervals. \textbf{b}, Direct tests of whether matching the stated number to the imposed stop helps.}
\label{tab:qwen-mechanism-audited}
\small
\textbf{a} Nine prompt instructions\\[2pt]
\begin{adjustbox}{max width=\textwidth}
\begin{tabular}{L{3.1cm} r r r r L{2.5cm} r r L{2.5cm}}
\toprule
& & & \multicolumn{3}{c}{Imposed stop $B=512$} & \multicolumn{3}{c}{Imposed stop $B=2{,}048$} \\
\cmidrule(lr){4-6} \cmidrule(lr){7-9}
Prompt instruction & Number stated $H$ & Median length & Finished & Accuracy & Difference [95\% CI] & Finished & Accuracy & Difference [95\% CI] \\
\midrule
No advance notice & \textemdash & $1{,}083$ & $20.3\%$ & $59.7\%$ & \textemdash & $72.1\%$ & $72.1\%$ & \textemdash \\
Numeric/concision prompt & $512$ & $900$ & $25.0\%$ & $60.1\%$ & $+0.4$ $[-1.2,\,+2.0]$ & $77.6\%$ & $72.0\%$ & $-0.1$ $[-1.4,\,+1.2]$ \\
Numeric/concision prompt & $2{,}048$ & $921$ & $23.5\%$ & $60.5\%$ & $+0.8$ $[-0.8,\,+2.4]$ & $77.1\%$ & $72.1\%$ & $-0.1$ $[-1.4,\,+1.3]$ \\
Advance notice of exact stop & $512$ & $857$ & $25.4\%$ & $61.3\%$ & $+1.6$ $[+0.1,\,+3.1]$ & $78.4\%$ & $72.6\%$ & $+0.5$ $[-0.7,\,+1.7]$ \\
Advance notice of exact stop & $2{,}048$ & $900$ & $23.0\%$ & $60.7\%$ & $+1.0$ $[-0.4,\,+2.4]$ & $77.3\%$ & $72.3\%$ & $+0.2$ $[-1.1,\,+1.4]$ \\
Limited budget (no number) & \textemdash & $874$ & $25.0\%$ & $60.1\%$ & $+0.4$ $[-1.2,\,+2.0]$ & $77.1\%$ & $71.8\%$ & $-0.3$ $[-1.6,\,+1.0]$ \\
Number only & $512$ & $924$ & $24.1\%$ & $60.9\%$ & $+1.2$ $[-0.3,\,+2.7]$ & $77.1\%$ & $72.6\%$ & $+0.5$ $[-0.8,\,+1.7]$ \\
Concise/early-answer instruction & \textemdash & $679$ & $36.4\%$ & $63.5\%$ & $+3.8$ $[+2.1,\,+5.6]$ & $82.3\%$ & $73.1\%$ & $+1.0$ $[-0.4,\,+2.4]$ \\
Irrelevant number & $512$ & $1{,}020$ & $21.2\%$ & $60.0\%$ & $+0.3$ $[-1.2,\,+1.8]$ & $73.0\%$ & $72.0\%$ & $-0.1$ $[-1.2,\,+1.0]$ \\
\bottomrule
\end{tabular}
\end{adjustbox}

\vspace{4pt}
\textbf{b} Direct test of matching the stated number to the stop\\[2pt]
\begin{tabular}{L{5.0cm} r L{2.3cm}}
\toprule
Prompt instruction & Match advantage (pp) & 95\% CI \\
\midrule
Numeric/concision prompt & $-0.2$ & $[-1.0,\,+0.6]$ \\
Advance notice of exact stop & $+0.1$ & $[-0.6,\,+0.8]$ \\
\bottomrule
\end{tabular}
\end{table}
\begin{table}[t]
\centering
\caption{\textbf{At $B=512$, the concise/early-answer contrast combines earlier completion with better unfinished prefixes.} \textbf{a}, Correct completion, all-pair accuracy, and both-active accuracy for concise/early-answer versus ordinary generation. Both-active rows condition on pairs in which neither trace has finished. \textbf{b}, The same $4{,}000$ item--replicate pairs partitioned into four termination strata under terminal-aware policy scoring and common replay. Contributions sum to the corresponding all-pair difference. Intervals use a nested item--replicate bootstrap.}
\label{tab:qwen-mechanism-decomposition}
\scriptsize
\textbf{a} Arm-level estimands\\[2pt]
\begin{adjustbox}{max width=\textwidth}
\begin{tabular}{L{2.8cm} L{2.6cm} r r r r r L{2.8cm}}
\toprule
Estimand & Readout context & $B$ & Items & Pairs & Ordinary & Concise/early-answer & $\Delta$ [95\% CI] \\
\midrule
Correct natural completion & Readout-context invariant & $512$ & $500$ & $4{,}000$ & $18.2\%$ & $31.9\%$ & $+13.8$ $[+11.6,\,+16.0]$ \\
Correct natural completion & Readout-context invariant & $2{,}048$ & $500$ & $4{,}000$ & $58.0\%$ & $65.6\%$ & $+7.6$ $[+5.8,\,+9.5]$ \\
\addlinespace
All-pair accuracy & Terminal-aware policy & $512$ & $500$ & $4{,}000$ & $59.7\%$ & $63.5\%$ & $+3.8$ $[+2.1,\,+5.6]$ \\
All-pair accuracy & Terminal-aware policy & $2{,}048$ & $500$ & $4{,}000$ & $72.1\%$ & $73.1\%$ & $+1.0$ $[-0.4,\,+2.4]$ \\
All-pair accuracy & Common replay context & $512$ & $500$ & $4{,}000$ & $59.8\%$ & $62.9\%$ & $+3.1$ $[+1.5,\,+4.8]$ \\
All-pair accuracy & Common replay context & $2{,}048$ & $500$ & $4{,}000$ & $72.0\%$ & $73.0\%$ & $+1.0$ $[-0.4,\,+2.4]$ \\
\addlinespace
Both-active accuracy & Terminal-aware policy & $512$ & $389$ & $2{,}476$ & $45.2\%$ & $49.2\%$ & $+4.0$ $[+1.6,\,+6.5]$ \\
Both-active accuracy & Terminal-aware policy & $2{,}048$ & $132$ & $644$ & $40.2\%$ & $41.0\%$ & $+0.8$ $[-4.7,\,+6.0]$ \\
Both-active accuracy & Common replay context & $512$ & $389$ & $2{,}476$ & $45.4\%$ & $48.2\%$ & $+2.7$ $[+0.6,\,+5.0]$ \\
Both-active accuracy & Common replay context & $2{,}048$ & $132$ & $644$ & $40.1\%$ & $40.5\%$ & $+0.5$ $[-4.8,\,+5.5]$ \\
\bottomrule
\end{tabular}
\end{adjustbox}

\vspace{4pt}
\textbf{b} Termination-stratum decomposition\\[2pt]
\begin{adjustbox}{max width=\textwidth}
\begin{tabular}{L{2.2cm} r L{3.8cm} r L{3.2cm} L{3.2cm}}
\toprule
Readout context & $B$ & Termination stratum & Share & Within-stratum $\Delta$ [95\% CI] & Contribution [95\% CI] \\
\midrule
Terminal-aware policy & $512$ & Neither trace finished & $61.9\%$ & $+4.0$ $[+1.6,\,+6.5]$ & $+2.5$ $[+1.0,\,+4.0]$ \\
Terminal-aware policy & $512$ & Concise/early-answer finished; ordinary active & $17.8\%$ & $+8.6$ $[+4.9,\,+12.6]$ & $+1.5$ $[+0.9,\,+2.3]$ \\
Terminal-aware policy & $512$ & Ordinary finished; concise/early-answer active & $1.7\%$ & $-7.5$ $[-20.3,\,+3.8]$ & $-0.1$ $[-0.4,\,+0.1]$ \\
Terminal-aware policy & $512$ & Both traces finished & $18.7\%$ & $-0.3$ $[-1.4,\,+0.8]$ & $-0.1$ $[-0.3,\,+0.2]$ \\
\addlinespace
Terminal-aware policy & $2{,}048$ & Neither trace finished & $16.1\%$ & $+0.8$ $[-4.7,\,+6.0]$ & $+0.1$ $[-0.8,\,+1.0]$ \\
Terminal-aware policy & $2{,}048$ & Concise/early-answer finished; ordinary active & $11.8\%$ & $+6.2$ $[+0.2,\,+12.1]$ & $+0.7$ $[+0.0,\,+1.4]$ \\
Terminal-aware policy & $2{,}048$ & Ordinary finished; concise/early-answer active & $1.7\%$ & $-16.7$ $[-34.3,\,+1.3]$ & $-0.3$ $[-0.6,\,+0.0]$ \\
Terminal-aware policy & $2{,}048$ & Both traces finished & $70.5\%$ & $+0.6$ $[-0.4,\,+1.6]$ & $+0.4$ $[-0.3,\,+1.2]$ \\
\addlinespace
Common replay context & $512$ & Neither trace finished & $61.9\%$ & $+2.7$ $[+0.6,\,+5.0]$ & $+1.7$ $[+0.4,\,+3.1]$ \\
Common replay context & $512$ & Concise/early-answer finished; ordinary active & $17.8\%$ & $+8.6$ $[+4.9,\,+12.6]$ & $+1.5$ $[+0.9,\,+2.3]$ \\
Common replay context & $512$ & Ordinary finished; concise/early-answer active & $1.7\%$ & $-6.0$ $[-19.4,\,+6.5]$ & $-0.1$ $[-0.3,\,+0.1]$ \\
Common replay context & $512$ & Both traces finished & $18.7\%$ & $-0.3$ $[-1.4,\,+0.8]$ & $-0.1$ $[-0.3,\,+0.2]$ \\
\addlinespace
Common replay context & $2{,}048$ & Neither trace finished & $16.1\%$ & $+0.5$ $[-4.8,\,+5.5]$ & $+0.1$ $[-0.8,\,+0.9]$ \\
Common replay context & $2{,}048$ & Concise/early-answer finished; ordinary active & $11.8\%$ & $+6.2$ $[+0.2,\,+12.1]$ & $+0.7$ $[+0.0,\,+1.4]$ \\
Common replay context & $2{,}048$ & Ordinary finished; concise/early-answer active & $1.7\%$ & $-16.7$ $[-34.5,\,+1.4]$ & $-0.3$ $[-0.6,\,+0.0]$ \\
Common replay context & $2{,}048$ & Both traces finished & $70.5\%$ & $+0.6$ $[-0.4,\,+1.7]$ & $+0.5$ $[-0.3,\,+1.2]$ \\
\bottomrule
\end{tabular}
\end{adjustbox}
\end{table}
\begin{table}[t]
\centering
\caption{\textbf{The 512-token concise/early-answer advantage persists under multiplicity control, direct prompt comparisons, and generated-answer continuation.} \textbf{a}, Concise/early-answer minus ordinary generation with simultaneous $95\%$ intervals over the $16$ instruction--deadline contrasts in each readout context. \textbf{b}, Direct comparisons at $B=512$. \textbf{c}, Greedy generated-answer continuations from the stored prefixes. Panels \textbf{b} and \textbf{c} use pointwise nested item--replicate bootstrap intervals; each cell contains $500$ items and eight paired replicates per item.}
\label{tab:qwen-mechanism-validation}
\scriptsize
\textbf{a} Familywise simultaneous inference\\[2pt]
\begin{tabular}{L{3.0cm} r r L{2.8cm}}
\toprule
Readout context & $B$ & $\Delta$ (pp) & Simultaneous 95\% CI \\
\midrule
Terminal-aware policy & $512$ & $+3.8$ & $[+1.7,\,+6.0]$ \\
Terminal-aware policy & $2{,}048$ & $+1.0$ & $[-1.2,\,+3.1]$ \\
\addlinespace
Common replay context & $512$ & $+3.1$ & $[+1.1,\,+5.1]$ \\
Common replay context & $2{,}048$ & $+1.0$ & $[-1.0,\,+3.0]$ \\
\bottomrule
\end{tabular}

\vspace{4pt}
\textbf{b} Direct prompt contrasts at $B=512$\\[2pt]
\begin{adjustbox}{max width=\textwidth}
\begin{tabular}{L{2.5cm} L{4.8cm} r r L{3.2cm}}
\toprule
Readout context & Comparator & Concise/early-answer acc. & Comparator acc. & $\Delta$ [95\% CI] \\
\midrule
Terminal-aware policy & Numeric/concision prompt, $H=512$ & $63.5\%$ & $60.1\%$ & $+3.4$ $[+1.7,\,+5.2]$ \\
Terminal-aware policy & Numeric/concision prompt, $H=2{,}048$ & $63.5\%$ & $60.5\%$ & $+3.0$ $[+1.3,\,+4.8]$ \\
Terminal-aware policy & Advance notice of exact stop, $H=512$ & $63.5\%$ & $61.3\%$ & $+2.3$ $[+0.5,\,+4.0]$ \\
\addlinespace
Common replay context & Numeric/concision prompt, $H=512$ & $62.9\%$ & $60.7\%$ & $+2.2$ $[+0.7,\,+3.8]$ \\
Common replay context & Numeric/concision prompt, $H=2{,}048$ & $62.9\%$ & $60.6\%$ & $+2.4$ $[+0.9,\,+3.9]$ \\
Common replay context & Advance notice of exact stop, $H=512$ & $62.9\%$ & $60.6\%$ & $+2.3$ $[+0.8,\,+3.9]$ \\
\bottomrule
\end{tabular}
\end{adjustbox}

\vspace{4pt}
\textbf{c} Greedy generated-answer continuation\\[2pt]
\begin{tabular}{L{3.0cm} r r r L{3.2cm}}
\toprule
Readout context & $B$ & Concise/early-answer acc. & Ordinary acc. & $\Delta$ [95\% CI] \\
\midrule
Terminal-aware policy & $512$ & $63.2\%$ & $59.1\%$ & $+4.1$ $[+2.4,\,+5.9]$ \\
Terminal-aware policy & $2{,}048$ & $73.1\%$ & $70.9\%$ & $+2.2$ $[+0.7,\,+3.7]$ \\
\addlinespace
Common replay context & $512$ & $62.6\%$ & $59.1\%$ & $+3.5$ $[+1.9,\,+5.2]$ \\
Common replay context & $2{,}048$ & $72.8\%$ & $70.9\%$ & $+1.9$ $[+0.4,\,+3.4]$ \\
\bottomrule
\end{tabular}
\end{table}

\begin{table*}[t]
\centering
\caption{\textbf{The numeric/concision prompt shortens Qwen reasoning but weakly tracks the announced token count.} \textbf{a}, Budget-specific reasoning length and adherence. Visible denotes the numeric/concision condition and hidden denotes matched ordinary generation. $L/B$ and $P(L\leq B)$ use visible-condition lengths; the log-length contrast is $\log L_\mathrm{visible}-\log L_\mathrm{hidden}$. \textbf{b}, Responsiveness across the four announced budgets. The slope and Spearman correlation are computed within each item--replicate trajectory. Intervals first average the three observed generation replicates within item and then bootstrap items.}
\label{tab:reviewer-qwen-length}
\small
\begin{adjustbox}{max width=\textwidth}
\begin{tabular}{L{1.9cm} L{1.9cm} r r r r r r r L{3.0cm}}
\toprule
\multicolumn{10}{l}{\textbf{a} Budget-specific length and adherence} \\
Model & Benchmark & $B$ & Items & Item--replicate pairs & Visible median $L$ & Hidden median $L$ & Median $L/B$ & $P(L\leq B)$ & Mean $\Delta\log L$ [95\% CI] \\
\midrule
Qwen3-14B & GPQA Diamond & 512 & 198 & 594 & $3{,}740.5$ & $4{,}486.5$ & $7.31$ & $1.9\%$ & $-0.179$ $[-0.215,\,-0.144]$ \\
Qwen3-14B & GPQA Diamond & 2{,}048 & 198 & 594 & $3{,}647.0$ & $4{,}486.5$ & $1.78$ & $29.6\%$ & $-0.176$ $[-0.211,\,-0.142]$ \\
Qwen3-14B & GPQA Diamond & 8{,}192 & 198 & 594 & $3{,}698.5$ & $4{,}486.5$ & $0.45$ & $90.6\%$ & $-0.159$ $[-0.196,\,-0.121]$ \\
Qwen3-14B & GPQA Diamond & 16{,}384 & 198 & 594 & $3{,}925.0$ & $4{,}486.5$ & $0.24$ & $99.2\%$ & $-0.133$ $[-0.168,\,-0.099]$ \\
Qwen3-14B & MMLU-Pro & 512 & 500 & 1{,}500 & $884.5$ & $1{,}094.0$ & $1.73$ & $23.6\%$ & $-0.189$ $[-0.214,\,-0.165]$ \\
Qwen3-14B & MMLU-Pro & 2{,}048 & 500 & 1{,}500 & $880.5$ & $1{,}094.0$ & $0.43$ & $77.5\%$ & $-0.170$ $[-0.195,\,-0.145]$ \\
Qwen3-14B & MMLU-Pro & 8{,}192 & 500 & 1{,}500 & $929.0$ & $1{,}094.0$ & $0.11$ & $96.4\%$ & $-0.134$ $[-0.159,\,-0.109]$ \\
Qwen3-14B & MMLU-Pro & 16{,}384 & 500 & 1{,}500 & $905.0$ & $1{,}094.0$ & $0.06$ & $99.6\%$ & $-0.130$ $[-0.154,\,-0.106]$ \\
\bottomrule
\end{tabular}
\end{adjustbox}
\vspace{0.75em}
\begin{adjustbox}{max width=\textwidth}
\begin{tabular}{L{2.0cm} L{2.0cm} L{5.2cm} r L{3.0cm}}
\toprule
\multicolumn{5}{l}{\textbf{b} Across-budget within-item responsiveness} \\
Model & Benchmark & Metric & Items & Estimate [95\% CI] \\
\midrule
Qwen3-14B & GPQA Diamond & Within-item slope of $\log L$ on $\log B$ & 198 & $+0.012$ $[+0.004,\,+0.021]$ \\
Qwen3-14B & GPQA Diamond & Within-item Spearman $\rho(B,L)$ & 198 & $+0.058$ $[+0.013,\,+0.103]$ \\
Qwen3-14B & MMLU-Pro & Within-item slope of $\log L$ on $\log B$ & 500 & $+0.018$ $[+0.013,\,+0.024]$ \\
Qwen3-14B & MMLU-Pro & Within-item Spearman $\rho(B,L)$ & 500 & $+0.121$ $[+0.091,\,+0.150]$ \\
\bottomrule
\end{tabular}
\end{adjustbox}
\end{table*}

\FloatBarrier
\begin{center}
\scriptsize
\renewcommand{\arraystretch}{0.95}
\setlength{\tabcolsep}{1pt}
\begin{longtable}{L{1.5cm} L{1.5cm} L{1.5cm} L{1.8cm} L{2.2cm} r L{2.25cm} L{2.25cm} L{2.25cm}}
\caption{\textbf{Replicate-specific estimates and crossed-intercept analyses preserve the absence of a broad Qwen gain and the positive gpt-oss effort differences.} Pooled rows first average the three observed generation replicates within item and then bootstrap items. The three named replicate rows report separate paired item-bootstrap estimates and 95\% intervals. Two-way intervals resample items and replicates. Crossed intervals use method-of-moments item and replicate random intercepts with a Satterthwaite approximation on complete item--replicate matrices, so crossed point estimates can differ from the primary $\Delta$ when pairs are ineligible. Qwen effects are visible minus hidden; gpt-oss effects are the completed lower-effort readout minus the matched-horizon high-effort readout. All effects are percentage points.}
\label{tab:reviewer-seed-sensitivity}\\
\toprule
Study & Model & Benchmark & Comparison & Replicate & Items & Item-bootstrap $\Delta$ [95\% CI] & Two-way item$\times$replicate $\Delta$ [95\% CI] & Crossed $\Delta$ [95\% CI] \\
\midrule
\endfirsthead
\caption[]{\textbf{Replicate-specific estimates and crossed-intercept analyses.} Continued.}\\
\toprule
Study & Model & Benchmark & Comparison & Replicate & Items & Item-bootstrap $\Delta$ [95\% CI] & Two-way item$\times$replicate $\Delta$ [95\% CI] & Crossed $\Delta$ [95\% CI] \\
\midrule
\endhead
\midrule
\multicolumn{9}{r}{Continued on next page}\\
\endfoot
\bottomrule
\endlastfoot
Qwen textual & Qwen3-14B & GPQA Diamond & Exact, $B=512$ & Pooled (3) & 198 & $-0.5$ $[-2.5,\,+1.3]$ & $-0.5$ $[-3.4,\,+2.2]$ & $-0.5$ $[-2.7,\,+1.7]$ \\
 &  &  &  & Run 41; seed 74{,}017 & 198 & $-1.0$ $[-5.1,\,+3.0]$ & \textemdash & \textemdash \\
 &  &  &  & Run 42; seed 74{,}029 & 198 & $+0.5$ $[-3.0,\,+4.0]$ & \textemdash & \textemdash \\
 &  &  &  & Run 43; seed 74{,}047 & 198 & $-1.0$ $[-4.5,\,+2.5]$ & \textemdash & \textemdash \\
Qwen textual & Qwen3-14B & GPQA Diamond & Exact, $B=2{,}048$ & Pooled (3) & 198 & $+0.7$ $[-3.0,\,+4.4]$ & $+0.7$ $[-4.0,\,+5.7]$ & $+0.7$ $[-3.3,\,+4.7]$ \\
 &  &  &  & Run 41; seed 74{,}017 & 198 & $-1.5$ $[-6.6,\,+3.5]$ & \textemdash & \textemdash \\
 &  &  &  & Run 42; seed 74{,}029 & 198 & $+0.0$ $[-5.1,\,+5.1]$ & \textemdash & \textemdash \\
 &  &  &  & Run 43; seed 74{,}047 & 198 & $+3.5$ $[-2.0,\,+9.6]$ & \textemdash & \textemdash \\
Qwen textual & Qwen3-14B & GPQA Diamond & Exact, $B=8{,}192$ & Pooled (3) & 198 & $+0.7$ $[-2.4,\,+3.9]$ & $+0.7$ $[-3.4,\,+5.1]$ & $+0.7$ $[-2.5,\,+3.8]$ \\
 &  &  &  & Run 41; seed 74{,}017 & 198 & $+1.0$ $[-4.5,\,+6.6]$ & \textemdash & \textemdash \\
 &  &  &  & Run 42; seed 74{,}029 & 198 & $-0.5$ $[-5.1,\,+4.0]$ & \textemdash & \textemdash \\
 &  &  &  & Run 43; seed 74{,}047 & 198 & $+1.5$ $[-4.0,\,+7.1]$ & \textemdash & \textemdash \\
Qwen textual & Qwen3-14B & GPQA Diamond & Exact, $B=16{,}384$ & Pooled (3) & 198 & $+1.3$ $[-1.9,\,+4.5]$ & $+1.3$ $[-2.9,\,+5.7]$ & $+1.3$ $[-1.8,\,+4.5]$ \\
 &  &  &  & Run 41; seed 74{,}017 & 198 & $+3.0$ $[-2.0,\,+8.1]$ & \textemdash & \textemdash \\
 &  &  &  & Run 42; seed 74{,}029 & 198 & $+0.0$ $[-5.1,\,+5.1]$ & \textemdash & \textemdash \\
 &  &  &  & Run 43; seed 74{,}047 & 198 & $+1.0$ $[-4.5,\,+6.6]$ & \textemdash & \textemdash \\
Qwen textual & Qwen3-14B & MMLU-Pro & Exact, $B=512$ & Pooled (3) & 500 & $+2.0$ $[+0.4,\,+3.7]$ & $+2.0$ $[-0.1,\,+4.1]$ & $+2.0$ $[+0.3,\,+3.7]$ \\
 &  &  &  & Run 41; seed 74{,}017 & 500 & $+2.6$ $[+0.2,\,+5.2]$ & \textemdash & \textemdash \\
 &  &  &  & Run 42; seed 74{,}029 & 500 & $+1.6$ $[-1.0,\,+4.2]$ & \textemdash & \textemdash \\
 &  &  &  & Run 43; seed 74{,}047 & 500 & $+1.8$ $[-1.0,\,+4.6]$ & \textemdash & \textemdash \\
Qwen textual & Qwen3-14B & MMLU-Pro & Exact, $B=2{,}048$ & Pooled (3) & 500 & $+0.7$ $[-0.9,\,+2.1]$ & $+0.7$ $[-1.6,\,+2.9]$ & $+0.7$ $[-1.1,\,+2.5]$ \\
 &  &  &  & Run 41; seed 74{,}017 & 500 & $+0.8$ $[-1.8,\,+3.4]$ & \textemdash & \textemdash \\
 &  &  &  & Run 42; seed 74{,}029 & 500 & $+2.0$ $[-0.2,\,+4.4]$ & \textemdash & \textemdash \\
 &  &  &  & Run 43; seed 74{,}047 & 500 & $-0.8$ $[-3.0,\,+1.6]$ & \textemdash & \textemdash \\
Qwen textual & Qwen3-14B & MMLU-Pro & Exact, $B=8{,}192$ & Pooled (3) & 500 & $-0.9$ $[-2.5,\,+0.7]$ & $-0.9$ $[-3.1,\,+1.5]$ & $-0.9$ $[-2.5,\,+0.7]$ \\
 &  &  &  & Run 41; seed 74{,}017 & 500 & $-1.0$ $[-3.6,\,+1.6]$ & \textemdash & \textemdash \\
 &  &  &  & Run 42; seed 74{,}029 & 500 & $+0.4$ $[-2.2,\,+3.0]$ & \textemdash & \textemdash \\
 &  &  &  & Run 43; seed 74{,}047 & 500 & $-2.0$ $[-4.6,\,+0.4]$ & \textemdash & \textemdash \\
Qwen textual & Qwen3-14B & MMLU-Pro & Exact, $B=16{,}384$ & Pooled (3) & 500 & $+0.1$ $[-1.2,\,+1.3]$ & $+0.1$ $[-2.3,\,+2.7]$ & $+0.1$ $[-2.5,\,+2.6]$ \\
 &  &  &  & Run 41; seed 74{,}017 & 500 & $-1.0$ $[-3.6,\,+1.6]$ & \textemdash & \textemdash \\
 &  &  &  & Run 42; seed 74{,}029 & 500 & $+2.2$ $[-0.4,\,+4.8]$ & \textemdash & \textemdash \\
 &  &  &  & Run 43; seed 74{,}047 & 500 & $-1.0$ $[-3.0,\,+1.0]$ & \textemdash & \textemdash \\
\addlinespace
Qwen literal & Qwen3-14B & GPQA Diamond & Continuation, $B=512$ & Pooled (3) & 198 & $-1.9$ $[-4.9,\,+1.0]$ & $-1.9$ $[-5.4,\,+1.7]$ & $-1.9$ $[-4.8,\,+1.1]$ \\
 &  &  &  & Run 41; seed 74{,}017 & 198 & $-3.0$ $[-6.6,\,+0.5]$ & \textemdash & \textemdash \\
 &  &  &  & Run 42; seed 74{,}029 & 198 & $-1.5$ $[-6.1,\,+3.0]$ & \textemdash & \textemdash \\
 &  &  &  & Run 43; seed 74{,}047 & 198 & $-1.0$ $[-5.1,\,+3.0]$ & \textemdash & \textemdash \\
Qwen literal & Qwen3-14B & GPQA Diamond & Continuation, $B=2{,}048$ & Pooled (3) & 198 & $+0.5$ $[-2.7,\,+3.7]$ & $+0.5$ $[-3.9,\,+5.4]$ & $+0.5$ $[-3.4,\,+4.5]$ \\
 &  &  &  & Run 41; seed 74{,}017 & 198 & $-1.5$ $[-6.1,\,+3.0]$ & \textemdash & \textemdash \\
 &  &  &  & Run 42; seed 74{,}029 & 198 & $-0.5$ $[-5.1,\,+4.0]$ & \textemdash & \textemdash \\
 &  &  &  & Run 43; seed 74{,}047 & 198 & $+3.5$ $[-1.5,\,+8.6]$ & \textemdash & \textemdash \\
Qwen literal & Qwen3-14B & MMLU-Pro & Continuation, $B=512$ & Pooled (3) & 500 & $+2.0$ $[+0.2,\,+3.9]$ & $+2.0$ $[-0.4,\,+4.5]$ & $+2.0$ $[+0.2,\,+3.8]$ \\
 &  &  &  & Run 41; seed 74{,}017 & 500 & $+2.0$ $[-0.6,\,+4.6]$ & \textemdash & \textemdash \\
 &  &  &  & Run 42; seed 74{,}029 & 500 & $+0.8$ $[-2.0,\,+3.6]$ & \textemdash & \textemdash \\
 &  &  &  & Run 43; seed 74{,}047 & 500 & $+3.2$ $[+0.2,\,+6.2]$ & \textemdash & \textemdash \\
Qwen literal & Qwen3-14B & MMLU-Pro & Continuation, $B=2{,}048$ & Pooled (3) & 500 & $+0.1$ $[-1.3,\,+1.5]$ & $+0.1$ $[-2.3,\,+2.4]$ & $+0.1$ $[-2.1,\,+2.3]$ \\
 &  &  &  & Run 41; seed 74{,}017 & 500 & $-1.0$ $[-3.4,\,+1.6]$ & \textemdash & \textemdash \\
 &  &  &  & Run 42; seed 74{,}029 & 500 & $-0.8$ $[-3.2,\,+1.6]$ & \textemdash & \textemdash \\
 &  &  &  & Run 43; seed 74{,}047 & 500 & $+2.0$ $[-0.2,\,+4.4]$ & \textemdash & \textemdash \\
\addlinespace
gpt-oss exact & gpt-oss-20b & GPQA Diamond & low & Pooled (3) & 198 & $+14.5$ $[+7.7,\,+21.2]$ & $+14.5$ $[+6.1,\,+22.7]$ & $+14.5$ $[+6.9,\,+22.0]$ \\
 &  &  &  & Run 41; seed 74{,}017 & 198 & $+13.6$ $[+5.1,\,+22.2]$ & \textemdash & \textemdash \\
 &  &  &  & Run 42; seed 74{,}029 & 198 & $+10.6$ $[+1.5,\,+19.7]$ & \textemdash & \textemdash \\
 &  &  &  & Run 43; seed 74{,}047 & 198 & $+19.2$ $[+10.6,\,+27.8]$ & \textemdash & \textemdash \\
gpt-oss exact & gpt-oss-20b & GPQA Diamond & medium & Pooled (3) & 198 & $+17.4$ $[+11.3,\,+23.7]$ & $+17.4$ $[+9.6,\,+25.8]$ & $+17.8$ $[+11.1,\,+24.4]$ \\
 &  &  &  & Run 41; seed 74{,}017 & 197 & $+13.7$ $[+5.6,\,+22.3]$ & \textemdash & \textemdash \\
 &  &  &  & Run 42; seed 74{,}029 & 198 & $+17.7$ $[+9.6,\,+25.8]$ & \textemdash & \textemdash \\
 &  &  &  & Run 43; seed 74{,}047 & 198 & $+21.2$ $[+12.6,\,+29.8]$ & \textemdash & \textemdash \\
gpt-oss exact & gpt-oss-20b & MMLU-Pro & low & Pooled (3) & 500 & $+20.9$ $[+16.7,\,+25.0]$ & $+20.9$ $[+16.2,\,+25.5]$ & $+20.9$ $[+16.8,\,+25.1]$ \\
 &  &  &  & Run 41; seed 74{,}017 & 500 & $+19.4$ $[+14.2,\,+24.4]$ & \textemdash & \textemdash \\
 &  &  &  & Run 42; seed 74{,}029 & 500 & $+21.2$ $[+16.2,\,+26.0]$ & \textemdash & \textemdash \\
 &  &  &  & Run 43; seed 74{,}047 & 500 & $+22.2$ $[+17.2,\,+27.0]$ & \textemdash & \textemdash \\
gpt-oss exact & gpt-oss-20b & MMLU-Pro & medium & Pooled (3) & 500 & $+18.0$ $[+14.5,\,+21.5]$ & $+18.0$ $[+13.7,\,+22.3]$ & $+18.0$ $[+14.5,\,+21.5]$ \\
 &  &  &  & Run 41; seed 74{,}017 & 500 & $+17.2$ $[+12.4,\,+21.8]$ & \textemdash & \textemdash \\
 &  &  &  & Run 42; seed 74{,}029 & 500 & $+17.0$ $[+12.2,\,+21.8]$ & \textemdash & \textemdash \\
 &  &  &  & Run 43; seed 74{,}047 & 500 & $+19.8$ $[+14.8,\,+24.6]$ & \textemdash & \textemdash \\
gpt-oss exact & gpt-oss-120b & GPQA Diamond & low & Pooled (3) & 198 & $+23.6$ $[+16.3,\,+30.6]$ & $+23.6$ $[+15.7,\,+31.5]$ & $+23.6$ $[+16.3,\,+30.9]$ \\
 &  &  &  & Run 41; seed 74{,}017 & 198 & $+20.7$ $[+12.1,\,+29.3]$ & \textemdash & \textemdash \\
 &  &  &  & Run 42; seed 74{,}029 & 198 & $+26.3$ $[+18.2,\,+34.3]$ & \textemdash & \textemdash \\
 &  &  &  & Run 43; seed 74{,}047 & 198 & $+23.7$ $[+15.7,\,+31.8]$ & \textemdash & \textemdash \\
gpt-oss exact & gpt-oss-120b & GPQA Diamond & medium & Pooled (3) & 198 & $+23.6$ $[+16.7,\,+30.5]$ & $+23.6$ $[+16.2,\,+31.0]$ & $+23.6$ $[+16.6,\,+30.5]$ \\
 &  &  &  & Run 41; seed 74{,}017 & 198 & $+23.7$ $[+15.6,\,+31.8]$ & \textemdash & \textemdash \\
 &  &  &  & Run 42; seed 74{,}029 & 198 & $+24.2$ $[+16.2,\,+32.3]$ & \textemdash & \textemdash \\
 &  &  &  & Run 43; seed 74{,}047 & 198 & $+22.7$ $[+14.6,\,+30.8]$ & \textemdash & \textemdash \\
gpt-oss exact & gpt-oss-120b & MMLU-Pro & low & Pooled (3) & 500 & $+26.3$ $[+21.9,\,+30.5]$ & $+26.3$ $[+21.7,\,+30.7]$ & $+26.3$ $[+22.0,\,+30.6]$ \\
 &  &  &  & Run 41; seed 74{,}017 & 500 & $+27.4$ $[+22.8,\,+32.0]$ & \textemdash & \textemdash \\
 &  &  &  & Run 42; seed 74{,}029 & 500 & $+25.4$ $[+20.6,\,+30.0]$ & \textemdash & \textemdash \\
 &  &  &  & Run 43; seed 74{,}047 & 500 & $+26.0$ $[+21.0,\,+30.8]$ & \textemdash & \textemdash \\
gpt-oss exact & gpt-oss-120b & MMLU-Pro & medium & Pooled (3) & 500 & $+24.2$ $[+20.3,\,+28.2]$ & $+24.2$ $[+19.6,\,+28.9]$ & $+24.2$ $[+19.9,\,+28.5]$ \\
 &  &  &  & Run 41; seed 74{,}017 & 500 & $+21.8$ $[+17.4,\,+26.2]$ & \textemdash & \textemdash \\
 &  &  &  & Run 42; seed 74{,}029 & 500 & $+25.8$ $[+21.2,\,+30.6]$ & \textemdash & \textemdash \\
 &  &  &  & Run 43; seed 74{,}047 & 500 & $+25.0$ $[+20.2,\,+29.8]$ & \textemdash & \textemdash \\
\addlinespace
gpt-oss continuation & gpt-oss-20b & GPQA Diamond & low & Pooled (3) & 198 & $+15.5$ $[+8.8,\,+22.1]$ & $+15.5$ $[+7.6,\,+23.2]$ & $+15.5$ $[+8.8,\,+22.2]$ \\
 &  &  &  & Run 41; seed 74{,}017 & 198 & $+16.7$ $[+8.6,\,+24.7]$ & \textemdash & \textemdash \\
 &  &  &  & Run 42; seed 74{,}029 & 198 & $+12.1$ $[+3.5,\,+20.7]$ & \textemdash & \textemdash \\
 &  &  &  & Run 43; seed 74{,}047 & 198 & $+17.7$ $[+9.1,\,+26.3]$ & \textemdash & \textemdash \\
gpt-oss continuation & gpt-oss-20b & GPQA Diamond & medium & Pooled (3) & 198 & $+17.4$ $[+11.1,\,+23.8]$ & $+17.4$ $[+9.8,\,+25.3]$ & $+17.8$ $[+11.4,\,+24.1]$ \\
 &  &  &  & Run 41; seed 74{,}017 & 197 & $+14.7$ $[+6.1,\,+23.4]$ & \textemdash & \textemdash \\
 &  &  &  & Run 42; seed 74{,}029 & 198 & $+17.2$ $[+9.1,\,+25.3]$ & \textemdash & \textemdash \\
 &  &  &  & Run 43; seed 74{,}047 & 198 & $+20.7$ $[+12.1,\,+29.3]$ & \textemdash & \textemdash \\
gpt-oss continuation & gpt-oss-20b & MMLU-Pro & low & Pooled (3) & 500 & $+20.5$ $[+16.2,\,+24.5]$ & $+20.5$ $[+15.9,\,+24.9]$ & $+20.5$ $[+16.3,\,+24.7]$ \\
 &  &  &  & Run 41; seed 74{,}017 & 500 & $+19.4$ $[+14.2,\,+24.4]$ & \textemdash & \textemdash \\
 &  &  &  & Run 42; seed 74{,}029 & 500 & $+20.6$ $[+15.6,\,+25.6]$ & \textemdash & \textemdash \\
 &  &  &  & Run 43; seed 74{,}047 & 500 & $+21.4$ $[+16.4,\,+26.0]$ & \textemdash & \textemdash \\
gpt-oss continuation & gpt-oss-20b & MMLU-Pro & medium & Pooled (3) & 500 & $+18.2$ $[+14.7,\,+21.7]$ & $+18.2$ $[+13.9,\,+22.5]$ & $+18.2$ $[+14.8,\,+21.6]$ \\
 &  &  &  & Run 41; seed 74{,}017 & 500 & $+17.2$ $[+12.4,\,+21.8]$ & \textemdash & \textemdash \\
 &  &  &  & Run 42; seed 74{,}029 & 500 & $+17.4$ $[+12.6,\,+22.0]$ & \textemdash & \textemdash \\
 &  &  &  & Run 43; seed 74{,}047 & 500 & $+20.0$ $[+15.0,\,+24.8]$ & \textemdash & \textemdash \\
gpt-oss continuation & gpt-oss-120b & GPQA Diamond & low & Pooled (3) & 198 & $+23.1$ $[+15.8,\,+30.3]$ & $+23.1$ $[+15.2,\,+30.8]$ & $+23.1$ $[+15.8,\,+30.3]$ \\
 &  &  &  & Run 41; seed 74{,}017 & 198 & $+20.2$ $[+11.6,\,+28.8]$ & \textemdash & \textemdash \\
 &  &  &  & Run 42; seed 74{,}029 & 198 & $+24.2$ $[+16.2,\,+32.3]$ & \textemdash & \textemdash \\
 &  &  &  & Run 43; seed 74{,}047 & 198 & $+24.7$ $[+16.7,\,+32.8]$ & \textemdash & \textemdash \\
gpt-oss continuation & gpt-oss-120b & GPQA Diamond & medium & Pooled (3) & 198 & $+23.1$ $[+16.2,\,+30.0]$ & $+23.1$ $[+15.8,\,+30.3]$ & $+23.1$ $[+16.2,\,+29.9]$ \\
 &  &  &  & Run 41; seed 74{,}017 & 198 & $+23.2$ $[+15.2,\,+31.3]$ & \textemdash & \textemdash \\
 &  &  &  & Run 42; seed 74{,}029 & 198 & $+23.2$ $[+15.2,\,+31.3]$ & \textemdash & \textemdash \\
 &  &  &  & Run 43; seed 74{,}047 & 198 & $+22.7$ $[+14.6,\,+30.8]$ & \textemdash & \textemdash \\
gpt-oss continuation & gpt-oss-120b & MMLU-Pro & low & Pooled (3) & 500 & $+25.9$ $[+21.5,\,+30.2]$ & $+25.9$ $[+21.3,\,+30.4]$ & $+25.9$ $[+21.6,\,+30.3]$ \\
 &  &  &  & Run 41; seed 74{,}017 & 500 & $+26.2$ $[+21.4,\,+31.0]$ & \textemdash & \textemdash \\
 &  &  &  & Run 42; seed 74{,}029 & 500 & $+26.4$ $[+21.4,\,+31.2]$ & \textemdash & \textemdash \\
 &  &  &  & Run 43; seed 74{,}047 & 500 & $+25.2$ $[+20.4,\,+30.0]$ & \textemdash & \textemdash \\
gpt-oss continuation & gpt-oss-120b & MMLU-Pro & medium & Pooled (3) & 500 & $+24.1$ $[+20.1,\,+28.0]$ & $+24.1$ $[+19.5,\,+28.7]$ & $+24.1$ $[+19.9,\,+28.3]$ \\
 &  &  &  & Run 41; seed 74{,}017 & 500 & $+22.0$ $[+17.6,\,+26.4]$ & \textemdash & \textemdash \\
 &  &  &  & Run 42; seed 74{,}029 & 500 & $+26.0$ $[+21.4,\,+30.6]$ & \textemdash & \textemdash \\
 &  &  &  & Run 43; seed 74{,}047 & 500 & $+24.2$ $[+19.4,\,+29.0]$ & \textemdash & \textemdash \\
\end{longtable}
\end{center}

\begin{table}[t]
\centering
\caption{\textbf{Selected answer-context checks do not reverse the main interpretation.} Qwen neutral replay applies the same budget-free closure to every prefix, including completed traces. The second block holds the source prefix fixed and changes only the answer-time context. All rows use full-vocabulary candidate logits and paired item-bootstrap intervals. Limited-time, paraphrased, and neutral Qwen closure wording gives the same qualitative pattern (\Cref{tab:extended-context-closure}).}
\label{tab:canonical-readout}
\small
\begin{adjustbox}{max width=\textwidth}
\begin{tabular}{L{2.0cm} L{4.2cm} L{1.8cm} r r r L{2.1cm}}
\toprule
Family & Contrast & Benchmark & $B$ & $n$ & $\Delta$ (pp) & 95\% CI \\
\midrule
\multicolumn{7}{l}{\textit{Neutral replay contrasts}} \\
Qwen3-14B & prompt vs.\ surprise, neutral replay & GPQA Diamond & $512$ & $198$ & $-1.3$ & $[-5.4,+2.5]$ \\
Qwen3-14B & prompt vs.\ surprise, neutral replay & GPQA Diamond & $2{,}048$ & $198$ & $-0.3$ & $[-4.4,+3.7]$ \\
Qwen3-14B & prompt vs.\ surprise, neutral replay & GPQA Diamond & $8{,}192$ & $198$ & $-3.0$ & $[-6.1,+0.0]$ \\
Qwen3-14B & prompt vs.\ surprise, neutral replay & GPQA Diamond & $16{,}384$ & $198$ & $+1.9$ & $[-1.6,+5.6]$ \\
Qwen3-14B & prompt vs.\ surprise, neutral replay & MMLU-Pro & $512$ & $500$ & $+1.9$ & $[+0.1,+3.7]$ \\
Qwen3-14B & prompt vs.\ surprise, neutral replay & MMLU-Pro & $2{,}048$ & $500$ & $+1.1$ & $[-0.5,+2.7]$ \\
Qwen3-14B & prompt vs.\ surprise, neutral replay & MMLU-Pro & $8{,}192$ & $500$ & $+1.0$ & $[-0.6,+2.7]$ \\
Qwen3-14B & prompt vs.\ surprise, neutral replay & MMLU-Pro & $16{,}384$ & $500$ & $+0.8$ & $[-0.9,+2.5]$ \\
\addlinespace
\multicolumn{7}{l}{\textit{Direct effects of the readout context}} \\
Qwen3-14B & budget present vs.\ absent at readout & GPQA Diamond & $512$ & $198$ & $-2.9$ & $[-5.9,+0.0]$ \\
Qwen3-14B & budget present vs.\ absent at readout & GPQA Diamond & $2{,}048$ & $198$ & $-0.5$ & $[-2.9,+1.9]$ \\
Qwen3-14B & budget present vs.\ absent at readout & MMLU-Pro & $512$ & $500$ & $-0.4$ & $[-1.5,+0.7]$ \\
Qwen3-14B & budget present vs.\ absent at readout & MMLU-Pro & $2{,}048$ & $500$ & $+0.1$ & $[-0.2,+0.3]$ \\
gpt-oss-20b & low vs.\ medium readout, high source & GPQA Diamond & $512$ & $191$ & $+2.1$ & $[-1.6,+6.3]$ \\
gpt-oss-20b & low vs.\ medium readout, high source & MMLU-Pro & $512$ & $361$ & $+1.7$ & $[+0.0,+3.6]$ \\
gpt-oss-120b & low vs.\ medium readout, high source & GPQA Diamond & $512$ & $188$ & $+0.0$ & $[-2.7,+2.7]$ \\
gpt-oss-120b & low vs.\ medium readout, high source & MMLU-Pro & $512$ & $335$ & $+0.3$ & $[-2.1,+2.7]$ \\
\bottomrule
\end{tabular}
\end{adjustbox}
\end{table}

\begin{table*}[t]
\centering
\caption{\textbf{Qwen proper scores show no consistent advantage for the numeric/concision prompt at 512 or 2,048 tokens.} Visible denotes the numeric/concision condition and hidden the matched hidden-horizon condition. $\Delta$ is visible minus hidden. Higher valid-option mass is preferred; lower conditional log loss and Brier score are preferred. Rows include token-exact candidate-logit pairs from three generation replicates. Arm means and differences first average replicates within item, after which intervals bootstrap items.}
\label{tab:reviewer-qwen-scores}
\small
\begin{adjustbox}{max width=\textwidth}
\begin{tabular}{L{1.9cm} L{1.9cm} r L{2.2cm} r r r r r L{2.8cm}}
\toprule
Model & Benchmark & $B$ & Metric & Items & Item--replicate pairs & Replicates & Visible & Hidden & $\Delta$ [95\% CI] \\
\midrule
Qwen3-14B & GPQA Diamond & 512 & Valid-option mass & 197 & 580 & 3 & $99.6\%$ & $99.5\%$ & $+0.1$ $[-0.2,\,+0.3]$ \\
Qwen3-14B & GPQA Diamond & 512 & Conditional log loss & 197 & 580 & 3 & $3.310$ & $3.021$ & $+0.289$ $[+0.142,\,+0.459]$ \\
Qwen3-14B & GPQA Diamond & 512 & Conditional Brier score & 197 & 580 & 3 & $0.942$ & $0.918$ & $+0.024$ $[+0.001,\,+0.048]$ \\
\addlinespace
Qwen3-14B & GPQA Diamond & 2{,}048 & Valid-option mass & 146 & 394 & 3 & $99.1\%$ & $99.1\%$ & $+0.1$ $[-0.3,\,+0.4]$ \\
Qwen3-14B & GPQA Diamond & 2{,}048 & Conditional log loss & 146 & 394 & 3 & $2.869$ & $2.738$ & $+0.131$ $[-0.176,\,+0.470]$ \\
Qwen3-14B & GPQA Diamond & 2{,}048 & Conditional Brier score & 146 & 394 & 3 & $0.918$ & $0.923$ & $-0.005$ $[-0.076,\,+0.070]$ \\
\addlinespace
Qwen3-14B & MMLU-Pro & 512 & Valid-option mass & 406 & 1{,}085 & 3 & $96.9\%$ & $96.5\%$ & $+0.4$ $[-0.1,\,+0.9]$ \\
Qwen3-14B & MMLU-Pro & 512 & Conditional log loss & 406 & 1{,}085 & 3 & $4.234$ & $4.351$ & $-0.117$ $[-0.324,\,+0.075]$ \\
Qwen3-14B & MMLU-Pro & 512 & Conditional Brier score & 406 & 1{,}085 & 3 & $0.822$ & $0.861$ & $-0.039$ $[-0.078,\,-0.003]$ \\
\addlinespace
Qwen3-14B & MMLU-Pro & 2{,}048 & Valid-option mass & 132 & 299 & 3 & $93.9\%$ & $93.8\%$ & $+0.1$ $[-1.5,\,+1.7]$ \\
Qwen3-14B & MMLU-Pro & 2{,}048 & Conditional log loss & 132 & 299 & 3 & $3.645$ & $3.811$ & $-0.167$ $[-0.666,\,+0.341]$ \\
Qwen3-14B & MMLU-Pro & 2{,}048 & Conditional Brier score & 132 & 299 & 3 & $0.877$ & $0.898$ & $-0.021$ $[-0.112,\,+0.070]$ \\
\bottomrule
\end{tabular}
\end{adjustbox}
\end{table*}

\begin{table*}[t]
\centering
\caption{\textbf{Lower gpt-oss effort is more accurate at its stopping point under candidate-logit and generated-answer readouts.} Each row compares a completed low- or medium-effort reasoning block with high effort at the item- and replicate-specific lower-effort terminal horizon: $198$ GPQA Diamond or $500$ MMLU-Pro items across three generation replicates. If high effort finishes before that horizon, its completed reasoning block is replayed. Candidate-logit and high-active columns use a common medium-effort answer context; the latter restricts to pairs in which high effort remains unfinished. Generated-answer columns continue all replayable prefixes without option normalization. Active/all gives the high-active and replayable pair counts, and parse coverage is the share with two parsed continuations. Differences are the completed lower-effort readout minus the matched-horizon high-effort readout. Intervals first average replicates within item and then bootstrap items ($10{,}000$ resamples).}
\label{tab:reviewer-gptoss-exact}
\small
\begin{adjustbox}{max width=\textwidth}
\begin{tabular}{L{2.0cm} L{1.9cm} l L{2.7cm} L{2.7cm} r L{2.7cm} r}
\toprule
Model & Benchmark & Effort & Candidate-logit $\Delta$ [95\% CI] & High-active $\Delta$ [95\% CI] & Active / all pairs & Generated-answer $\Delta$ [95\% CI] & Parse coverage \\
\midrule
gpt-oss-20b & GPQA Diamond & low & $+14.5$ $[+7.7,\,+21.2]$ & $+14.5$ $[+7.7,\,+21.2]$ & 594/594 & $+15.5$ $[+8.8,\,+22.1]$ & $99.7\%$ \\
gpt-oss-20b & GPQA Diamond & medium & $+17.4$ $[+11.3,\,+23.7]$ & $+17.9$ $[+11.5,\,+24.3]$ & 567/593 & $+17.4$ $[+11.1,\,+23.8]$ & $98.8\%$ \\
gpt-oss-20b & MMLU-Pro & low & $+20.9$ $[+16.7,\,+25.0]$ & $+21.1$ $[+16.9,\,+25.2]$ & 1{,}496/1{,}500 & $+20.5$ $[+16.2,\,+24.5]$ & $99.3\%$ \\
gpt-oss-20b & MMLU-Pro & medium & $+18.0$ $[+14.5,\,+21.5]$ & $+19.2$ $[+15.6,\,+22.9]$ & 1{,}407/1{,}500 & $+18.2$ $[+14.7,\,+21.7]$ & $99.1\%$ \\
\addlinespace
gpt-oss-120b & GPQA Diamond & low & $+23.6$ $[+16.3,\,+30.6]$ & $+23.6$ $[+16.3,\,+30.6]$ & 594/594 & $+23.1$ $[+15.8,\,+30.3]$ & $99.8\%$ \\
gpt-oss-120b & GPQA Diamond & medium & $+23.6$ $[+16.7,\,+30.5]$ & $+23.9$ $[+17.0,\,+30.9]$ & 591/594 & $+23.1$ $[+16.2,\,+30.0]$ & $100.0\%$ \\
gpt-oss-120b & MMLU-Pro & low & $+26.3$ $[+21.9,\,+30.5]$ & $+26.3$ $[+21.9,\,+30.5]$ & 1{,}499/1{,}500 & $+25.9$ $[+21.5,\,+30.2]$ & $99.7\%$ \\
gpt-oss-120b & MMLU-Pro & medium & $+24.2$ $[+20.3,\,+28.2]$ & $+24.9$ $[+20.9,\,+28.9]$ & 1{,}473/1{,}500 & $+24.1$ $[+20.1,\,+28.0]$ & $99.5\%$ \\
\bottomrule
\end{tabular}
\end{adjustbox}
\end{table*}

\begin{table*}[p]
\centering
\caption{\textbf{High valid-option mass coexists with proper-score disagreement driven by near-zero correct-option probabilities in terminal readouts.} The table uses the all-replayable token-exact prefixes in \Cref{tab:reviewer-gptoss-exact}, comprising $198$ GPQA Diamond or $500$ MMLU-Pro items across three generation replicates. \textbf{a}, Valid-option mass. \textbf{b}, Absolute arm means and paired differences for conditional log loss and multiclass Brier score. \textbf{c}, Median conditional log loss and the share of all probes with option-normalized $q_{\mathrm{gold}}<10^{-4}$. Differences are the completed lower-effort readout minus the matched-horizon high-effort readout. Higher valid-option mass and lower proper scores are preferred. Means and differences first average replicates within item; intervals then bootstrap items.}
\label{tab:reviewer-gptoss-scores}
\scriptsize
\textbf{a} Valid-option mass\\[2pt]
\begin{adjustbox}{max width=\textwidth}
\begin{tabular}{L{2.0cm} L{1.9cm} l r r r r L{3.0cm}}
\toprule
Model & Benchmark & Effort & Items & Replicates & Terminal & High prefix & $\Delta$ [95\% CI] \\
\midrule
gpt-oss-20b & GPQA Diamond & low & 198 & 3 & $99.9\%$ & $98.5\%$ & $+1.5$ $[+1.2,\,+1.8]$ \\
gpt-oss-20b & GPQA Diamond & medium & 198 & 3 & $99.8\%$ & $98.0\%$ & $+1.8$ $[+1.3,\,+2.5]$ \\
gpt-oss-20b & MMLU-Pro & low & 500 & 3 & $99.6\%$ & $99.2\%$ & $+0.4$ $[-0.0,\,+0.8]$ \\
gpt-oss-20b & MMLU-Pro & medium & 500 & 3 & $99.9\%$ & $99.3\%$ & $+0.6$ $[+0.4,\,+0.8]$ \\
\addlinespace
gpt-oss-120b & GPQA Diamond & low & 198 & 3 & $100.0\%$ & $100.0\%$ & $+0.0$ $[+0.0,\,+0.0]$ \\
gpt-oss-120b & GPQA Diamond & medium & 198 & 3 & $100.0\%$ & $100.0\%$ & $+0.0$ $[+0.0,\,+0.0]$ \\
gpt-oss-120b & MMLU-Pro & low & 500 & 3 & $100.0\%$ & $99.9\%$ & $+0.1$ $[+0.0,\,+0.2]$ \\
gpt-oss-120b & MMLU-Pro & medium & 500 & 3 & $100.0\%$ & $99.9\%$ & $+0.1$ $[+0.0,\,+0.2]$ \\
\bottomrule
\end{tabular}
\end{adjustbox}

\vspace{3pt}
\textbf{b} Conditional proper-score means\\[2pt]
\begin{adjustbox}{max width=\textwidth}
\begin{tabular}{L{2.0cm} L{1.9cm} l L{2.8cm} r r L{3.0cm}}
\toprule
Model & Benchmark & Effort & Metric & Terminal & High prefix & $\Delta$ [95\% CI] \\
\midrule
gpt-oss-20b & GPQA Diamond & low & Conditional log loss & $5.732$ & $1.396$ & $+4.336$ $[+3.599,\,+5.068]$ \\
 &  &  & Conditional Brier score & $0.878$ & $0.723$ & $+0.155$ $[+0.051,\,+0.259]$ \\
gpt-oss-20b & GPQA Diamond & medium & Conditional log loss & $5.803$ & $1.748$ & $+4.056$ $[+3.224,\,+4.890]$ \\
 &  &  & Conditional Brier score & $0.702$ & $0.725$ & $-0.023$ $[-0.124,\,+0.079]$ \\
gpt-oss-20b & MMLU-Pro & low & Conditional log loss & $5.314$ & $1.855$ & $+3.459$ $[+2.953,\,+3.987]$ \\
 &  &  & Conditional Brier score & $0.712$ & $0.721$ & $-0.009$ $[-0.072,\,+0.056]$ \\
gpt-oss-20b & MMLU-Pro & medium & Conditional log loss & $5.075$ & $2.328$ & $+2.747$ $[+2.241,\,+3.264]$ \\
 &  &  & Conditional Brier score & $0.552$ & $0.651$ & $-0.099$ $[-0.154,\,-0.045]$ \\
\addlinespace
gpt-oss-120b & GPQA Diamond & low & Conditional log loss & $8.870$ & $3.351$ & $+5.520$ $[+4.126,\,+6.923]$ \\
 &  &  & Conditional Brier score & $0.700$ & $1.014$ & $-0.314$ $[-0.449,\,-0.179]$ \\
gpt-oss-120b & GPQA Diamond & medium & Conditional log loss & $8.457$ & $3.286$ & $+5.171$ $[+3.687,\,+6.648]$ \\
 &  &  & Conditional Brier score & $0.630$ & $0.957$ & $-0.327$ $[-0.460,\,-0.196]$ \\
gpt-oss-120b & MMLU-Pro & low & Conditional log loss & $6.219$ & $2.974$ & $+3.244$ $[+2.429,\,+4.068]$ \\
 &  &  & Conditional Brier score & $0.491$ & $0.812$ & $-0.322$ $[-0.398,\,-0.245]$ \\
gpt-oss-120b & MMLU-Pro & medium & Conditional log loss & $5.623$ & $3.196$ & $+2.427$ $[+1.655,\,+3.226]$ \\
 &  &  & Conditional Brier score & $0.419$ & $0.759$ & $-0.341$ $[-0.413,\,-0.270]$ \\
\bottomrule
\end{tabular}
\end{adjustbox}

\vspace{3pt}
\textbf{c} Conditional log-loss medians and lower tails\\[2pt]
\begin{adjustbox}{max width=\textwidth}
\begin{tabular}{L{2.0cm} L{1.9cm} l r r r r L{2.8cm}}
\toprule
Model & Benchmark & Effort & Pairs & Terminal median & High-prefix median & \multicolumn{2}{c}{$q_{\mathrm{gold}}<10^{-4}$, terminal / high prefix} \\
\midrule
gpt-oss-20b & GPQA Diamond & low & 594 & $0.000$ & $1.224$ & \multicolumn{2}{c}{$36.0\%$ / $0.2\%$} \\
gpt-oss-20b & GPQA Diamond & medium & 593 & $0.000$ & $1.155$ & \multicolumn{2}{c}{$34.6\%$ / $2.4\%$} \\
gpt-oss-20b & MMLU-Pro & low & $1{,}500$ & $0.000$ & $1.634$ & \multicolumn{2}{c}{$32.3\%$ / $0.4\%$} \\
gpt-oss-20b & MMLU-Pro & medium & $1{,}500$ & $0.000$ & $0.837$ & \multicolumn{2}{c}{$27.4\%$ / $5.9\%$} \\
\addlinespace
gpt-oss-120b & GPQA Diamond & low & 594 & $0.000$ & $2.144$ & \multicolumn{2}{c}{$34.8\%$ / $8.2\%$} \\
gpt-oss-120b & GPQA Diamond & medium & 594 & $0.000$ & $1.542$ & \multicolumn{2}{c}{$31.5\%$ / $8.6\%$} \\
gpt-oss-120b & MMLU-Pro & low & $1{,}500$ & $0.000$ & $0.946$ & \multicolumn{2}{c}{$24.5\%$ / $8.5\%$} \\
gpt-oss-120b & MMLU-Pro & medium & $1{,}500$ & $0.000$ & $0.259$ & \multicolumn{2}{c}{$20.9\%$ / $11.7\%$} \\
\bottomrule
\end{tabular}
\end{adjustbox}
\end{table*}

\FloatBarrier
\begin{center}
\scriptsize
\setlength{\tabcolsep}{3pt}
\begin{longtable}{L{1.8cm} L{1.8cm} L{1.5cm} L{2.7cm} r r r L{2.6cm} r}
\caption{\textbf{Earlier lower-effort completion accounts for most of the fixed-horizon gpt-oss gap.} Active means that reasoning had not terminated at $B=512$. Shares and contributions use item--replicate pairs from three generated replicates, and contributions sum to the overall lower-minus-high accuracy difference. Within-stratum intervals average replicate outcomes within item before bootstrapping items. Because contributions use raw pair means, they need not equal the share times the displayed item-averaged difference.}
\label{tab:reviewer-gptoss-strata}\\
\toprule
Model & Benchmark & Contrast & Termination stratum & Items & Pairs & Share & Within-stratum $\Delta$ [95\% CI] & Contribution \\
\midrule
\endfirsthead
\caption[]{\textbf{gpt-oss termination strata at $B=512$.} Continued.}\\
\toprule
Model & Benchmark & Contrast & Termination stratum & Items & Pairs & Share & Within-stratum $\Delta$ [95\% CI] & Contribution \\
\midrule
\endhead
\midrule
\multicolumn{9}{r}{Continued on next page}\\
\endfoot
\bottomrule
\endlastfoot
gpt-oss-20b & GPQA Diamond & Low--high & Both active & 95 & 178 & $30.0\%$ & $-4.2$ $[-13.3,\,+4.9]$ & $-1.0$ \\
gpt-oss-20b & GPQA Diamond & Low--high & Lower complete; high active & 171 & 403 & $67.8\%$ & $+16.2$ $[+8.1,\,+24.2]$ & $+11.8$ \\
gpt-oss-20b & GPQA Diamond & Low--high & Both complete & 7 & 13 & $2.2\%$ & $-14.3$ $[-42.9,\,+0.0]$ & $-0.2$ \\
\addlinespace
gpt-oss-20b & GPQA Diamond & Medium--high & Both active & 182 & 501 & $84.3\%$ & $+2.8$ $[-1.5,\,+7.1]$ & $+1.9$ \\
gpt-oss-20b & GPQA Diamond & Medium--high & Lower complete; high active & 49 & 80 & $13.5\%$ & $+23.1$ $[+12.2,\,+34.7]$ & $+3.4$ \\
gpt-oss-20b & GPQA Diamond & Medium--high & Both complete & 7 & 13 & $2.2\%$ & $+0.0$ $[+0.0,\,+0.0]$ & $+0.0$ \\
\addlinespace
gpt-oss-20b & MMLU-Pro & Low--high & Both active & 56 & 92 & $6.2\%$ & $+1.2$ $[-11.0,\,+13.4]$ & $-0.1$ \\
gpt-oss-20b & MMLU-Pro & Low--high & Lower complete; high active & 416 & 1{,}045 & $69.9\%$ & $+7.3$ $[+3.3,\,+11.3]$ & $+6.2$ \\
gpt-oss-20b & MMLU-Pro & Low--high & Both complete & 169 & 357 & $23.9\%$ & $-6.8$ $[-10.5,\,-3.5]$ & $-1.5$ \\
\addlinespace
gpt-oss-20b & MMLU-Pro & Medium--high & Both active & 303 & 710 & $47.4\%$ & $-1.6$ $[-5.7,\,+2.6]$ & $-0.3$ \\
gpt-oss-20b & MMLU-Pro & Medium--high & Lower complete; high active & 255 & 432 & $28.8\%$ & $+11.6$ $[+6.9,\,+16.3]$ & $+3.4$ \\
gpt-oss-20b & MMLU-Pro & Medium--high & Both complete & 163 & 344 & $22.9\%$ & $-0.4$ $[-2.5,\,+1.6]$ & $-0.1$ \\
gpt-oss-20b & MMLU-Pro & Medium--high & High complete; lower active & 12 & 13 & $0.9\%$ & $-25.0$ $[-50.0,\,+0.0]$ & $-0.2$ \\
\addlinespace
gpt-oss-120b & GPQA Diamond & Low--high & Both active & 63 & 132 & $22.2\%$ & $+10.6$ $[-1.6,\,+22.8]$ & $+3.2$ \\
gpt-oss-120b & GPQA Diamond & Low--high & Lower complete; high active & 168 & 438 & $73.7\%$ & $+22.3$ $[+15.2,\,+29.5]$ & $+15.8$ \\
gpt-oss-120b & GPQA Diamond & Low--high & Both complete & 13 & 24 & $4.0\%$ & $+0.0$ $[+0.0,\,+0.0]$ & $+0.0$ \\
\addlinespace
gpt-oss-120b & GPQA Diamond & Medium--high & Both active & 154 & 421 & $70.9\%$ & $+4.5$ $[-0.1,\,+9.4]$ & $+1.9$ \\
gpt-oss-120b & GPQA Diamond & Medium--high & Lower complete; high active & 68 & 149 & $25.1\%$ & $+20.1$ $[+9.8,\,+30.4]$ & $+5.6$ \\
gpt-oss-120b & GPQA Diamond & Medium--high & Both complete & 13 & 24 & $4.0\%$ & $+0.0$ $[+0.0,\,+0.0]$ & $+0.0$ \\
\addlinespace
gpt-oss-120b & MMLU-Pro & Low--high & Both active & 46 & 99 & $6.6\%$ & $+6.5$ $[-6.5,\,+19.6]$ & $+0.4$ \\
gpt-oss-120b & MMLU-Pro & Low--high & Lower complete; high active & 353 & 900 & $60.0\%$ & $+12.0$ $[+7.7,\,+16.3]$ & $+7.4$ \\
gpt-oss-120b & MMLU-Pro & Low--high & Both complete & 214 & 500 & $33.4\%$ & $+0.2$ $[-2.1,\,+2.6]$ & $-0.1$ \\
\addlinespace
gpt-oss-120b & MMLU-Pro & Medium--high & Both active & 164 & 379 & $25.3\%$ & $+2.0$ $[-3.5,\,+7.5]$ & $+0.0$ \\
gpt-oss-120b & MMLU-Pro & Medium--high & Lower complete; high active & 284 & 621 & $41.4\%$ & $+11.4$ $[+7.0,\,+15.9]$ & $+5.3$ \\
gpt-oss-120b & MMLU-Pro & Medium--high & Both complete & 213 & 498 & $33.2\%$ & $+0.9$ $[-0.9,\,+2.8]$ & $+0.1$ \\
gpt-oss-120b & MMLU-Pro & Medium--high & High complete; lower active & 2 & 2 & $0.1\%$ & $-50.0$ $[-100.0,\,+0.0]$ & $-0.1$ \\
\end{longtable}
\end{center}

\clearpage
\begin{table}[htbp]
\centering
\caption{\textbf{Lower gpt-oss effort makes answer evidence available earlier, while active-prefix differences are smaller.} The terminal-aware area under the curve (AUC) averages paired correct-option log-odds differences from $0$ to $512$ reasoning tokens and carries terminal states forward. Stable-correct $\Delta_{256}$ compares whether answers at $256$ tokens remain correct through $512$. The active-pair AUC retains item--replicate pairs for which both traces have $L>128$ and recomputes the AUC over $0$--$128$; conditioning on survival changes the compared population. Qwen contrasts are numeric/concision prompt minus surprise stop. gpt-oss contrasts are lower effort minus high effort. Intervals bootstrap items.}
\label{tab:extended-answer-availability}
\small
\textbf{a} Terminal-aware and stable-correct contrasts\\[2pt]
\begin{adjustbox}{max width=\textwidth}
\begin{tabular}{L{2.0cm} L{1.9cm} L{2.7cm} r L{3.0cm} L{3.2cm}}
\toprule
Model & Benchmark & Contrast & $n$ & Terminal-aware AUC $\Delta$ log-odds & Stable-correct $\Delta_{256}$ (pp) \\
\midrule
gpt-oss-20b & GPQA Diamond & low minus high & 196 & $+1.85$ $[+0.74,+2.90]$ & $+15.8$ $[+8.2,+23.5]$ \\
gpt-oss-20b & GPQA Diamond & medium minus high & 196 & $+0.83$ $[+0.49,+1.18]$ & $+4.3$ $[+0.2,+8.3]$ \\
gpt-oss-20b & MMLU-Pro & low minus high & 500 & $+1.44$ $[+0.76,+2.11]$ & $+17.7$ $[+14.1,+21.3]$ \\
gpt-oss-20b & MMLU-Pro & medium minus high & 500 & $+1.36$ $[+1.04,+1.70]$ & $+8.3$ $[+5.8,+10.7]$ \\
\addlinespace
gpt-oss-120b & GPQA Diamond & low minus high & 198 & $+4.22$ $[+2.38,+6.00]$ & $+14.1$ $[+7.6,+20.7]$ \\
gpt-oss-120b & GPQA Diamond & medium minus high & 198 & $+2.00$ $[+1.29,+2.70]$ & $+3.9$ $[-0.5,+8.4]$ \\
gpt-oss-120b & MMLU-Pro & low minus high & 500 & $+3.76$ $[+2.59,+4.91]$ & $+15.7$ $[+11.7,+19.6]$ \\
gpt-oss-120b & MMLU-Pro & medium minus high & 500 & $+3.27$ $[+2.65,+3.87]$ & $+9.2$ $[+6.5,+11.9]$ \\
\addlinespace
Qwen3-14B & GPQA Diamond & prompt minus surprise & 198 & $+0.17$ $[+0.03,+0.32]$ & $-1.2$ $[-4.2,+1.9]$ \\
Qwen3-14B & MMLU-Pro & prompt minus surprise & 500 & $+0.42$ $[+0.18,+0.65]$ & $+0.8$ $[-0.7,+2.3]$ \\
\bottomrule
\end{tabular}
\end{adjustbox}

\vspace{6pt}
\textbf{b} Jointly active early-prefix contrasts\\[2pt]
\begin{adjustbox}{max width=\textwidth}
\begin{tabular}{L{2.0cm} L{1.9cm} L{2.7cm} r L{3.2cm}}
\toprule
Model & Benchmark & Contrast & Active item--replicate pairs & Active-pair AUC$_{0:128}$ \\
\midrule
gpt-oss-20b & GPQA Diamond & low minus high & 414 & $+0.05$ $[-0.10,+0.20]$ \\
gpt-oss-20b & GPQA Diamond & medium minus high & 580 & $+0.06$ $[-0.02,+0.16]$ \\
gpt-oss-20b & MMLU-Pro & low minus high & 498 & $+0.12$ $[-0.06,+0.28]$ \\
gpt-oss-20b & MMLU-Pro & medium minus high & 1{,}314 & $+0.19$ $[+0.04,+0.36]$ \\
\addlinespace
gpt-oss-120b & GPQA Diamond & low minus high & 441 & $-0.10$ $[-0.46,+0.25]$ \\
gpt-oss-120b & GPQA Diamond & medium minus high & 580 & $-0.05$ $[-0.24,+0.15]$ \\
gpt-oss-120b & MMLU-Pro & low minus high & 534 & $+0.12$ $[-0.33,+0.55]$ \\
gpt-oss-120b & MMLU-Pro & medium minus high & 1{,}178 & $+0.53$ $[+0.29,+0.77]$ \\
\addlinespace
Qwen3-14B & GPQA Diamond & prompt minus surprise & 593 & $+0.08$ $[-0.01,+0.18]$ \\
Qwen3-14B & MMLU-Pro & prompt minus surprise & 1{,}498 & $+0.06$ $[-0.03,+0.16]$ \\
\bottomrule
\end{tabular}
\end{adjustbox}
\end{table}

\clearpage
\begin{table}[p]
\centering
\caption{\textbf{Qwen's small prompt effects are distributed across unfinished-prefix and completion strata.} Qwen3-14B numeric/concision-prompt and hidden-horizon traces are partitioned by whether each arm has completed at the imposed horizon $B$. Shares use all token-exact matched item--replicate pairs and sum to one before rounding within each benchmark--horizon cell. Within-stratum differences are visible minus hidden accuracy. Contributions use the item-normalized pair decomposition and sum before rounding to the primary pooled difference. Intervals average the three observed replicates within item before bootstrapping items. All four horizons are reported. Several late-horizon strata, especially the both-active strata, are sparse and should be read descriptively.}
\label{tab:extended-qwen-strata}
\footnotesize
\setlength{\tabcolsep}{3pt}
\begin{adjustbox}{max width=\textwidth}
\begin{tabular}{L{2.0cm} r L{3.8cm} r r r L{2.8cm} r}
\toprule
Benchmark & $B$ & Termination stratum & Items & Item--replicate pairs & Share & Within-stratum $\Delta$ [95\% CI] & Contribution \\
\midrule
GPQA Diamond & 512 & Both active & 197 & 580 & $97.6\%$ & $-0.8$ $[-2.8,+1.2]$ & $-0.7$ \\
 &  & Visible complete; hidden active & 9 & 11 & $1.9\%$ & $+11.1$ $[+0.0,+33.3]$ & $+0.2$ \\
 &  & Hidden complete; visible active & 3 & 3 & $0.5\%$ & $+0.0$ $[+0.0,+0.0]$ & $+0.0$ \\
\addlinespace
GPQA Diamond & 2{,}048 & Both active & 146 & 394 & $66.3\%$ & $-0.5$ $[-5.8,+4.9]$ & $-0.3$ \\
 &  & Visible complete; hidden active & 42 & 56 & $9.4\%$ & $+8.3$ $[+2.4,+15.5]$ & $+0.8$ \\
 &  & Hidden complete; visible active & 22 & 24 & $4.0\%$ & $+2.3$ $[-13.6,+18.2]$ & $+0.0$ \\
 &  & Both complete & 55 & 120 & $20.2\%$ & $+1.2$ $[-1.8,+5.5]$ & $+0.2$ \\
\addlinespace
GPQA Diamond & 8{,}192 & Both active & 25 & 41 & $6.9\%$ & $+7.3$ $[-9.3,+24.0]$ & $+0.7$ \\
 &  & Visible complete; hidden active & 42 & 56 & $9.4\%$ & $+0.4$ $[-15.1,+15.1]$ & $-0.2$ \\
 &  & Hidden complete; visible active & 12 & 15 & $2.5\%$ & $-33.3$ $[-58.3,-8.3]$ & $-0.8$ \\
 &  & Both complete & 180 & 482 & $81.1\%$ & $+1.0$ $[-2.2,+4.3]$ & $+1.0$ \\
\addlinespace
GPQA Diamond & 16{,}384 & Both active & 1 & 2 & $0.3\%$ & $+0.0$ $[+0.0,+0.0]$ & $+0.0$ \\
 &  & Visible complete; hidden active & 2 & 3 & $0.5\%$ & $+50.0$ $[+0.0,+100.0]$ & $+0.3$ \\
 &  & Hidden complete; visible active & 3 & 3 & $0.5\%$ & $+0.0$ $[-100.0,+100.0]$ & $+0.0$ \\
 &  & Both complete & 197 & 586 & $98.7\%$ & $+0.7$ $[-2.5,+3.9]$ & $+1.0$ \\
\addlinespace
MMLU-Pro & 512 & Both active & 406 & 1{,}085 & $72.3\%$ & $+2.0$ $[-0.1,+4.2]$ & $+1.3$ \\
 &  & Visible complete; hidden active & 99 & 118 & $7.9\%$ & $+6.6$ $[+0.5,+13.1]$ & $+0.5$ \\
 &  & Hidden complete; visible active & 54 & 61 & $4.1\%$ & $+0.9$ $[-6.5,+8.3]$ & $+0.1$ \\
 &  & Both complete & 118 & 236 & $15.7\%$ & $+0.6$ $[-2.0,+3.1]$ & $+0.1$ \\
\addlinespace
MMLU-Pro & 2{,}048 & Both active & 132 & 299 & $19.9\%$ & $+2.0$ $[-3.5,+7.6]$ & $+0.2$ \\
 &  & Visible complete; hidden active & 102 & 146 & $9.7\%$ & $+2.6$ $[-4.7,+9.8]$ & $+0.3$ \\
 &  & Hidden complete; visible active & 35 & 39 & $2.6\%$ & $-5.7$ $[-18.6,+7.1]$ & $-0.2$ \\
 &  & Both complete & 376 & 1{,}016 & $67.7\%$ & $+1.3$ $[-0.1,+2.9]$ & $+0.4$ \\
\addlinespace
MMLU-Pro & 8{,}192 & Both active & 19 & 40 & $2.7\%$ & $-11.4$ $[-28.1,+2.6]$ & $-0.3$ \\
 &  & Visible complete; hidden active & 29 & 34 & $2.3\%$ & $+1.7$ $[-17.2,+20.7]$ & $+0.1$ \\
 &  & Hidden complete; visible active & 11 & 14 & $0.9\%$ & $-22.7$ $[-54.5,+4.5]$ & $-0.2$ \\
 &  & Both complete & 481 & 1{,}412 & $94.1\%$ & $-0.8$ $[-2.5,+0.8]$ & $-0.5$ \\
\addlinespace
MMLU-Pro & 16{,}384 & Both active & 2 & 3 & $0.2\%$ & $-50.0$ $[-100.0,+0.0]$ & $-0.1$ \\
 &  & Visible complete; hidden active & 5 & 7 & $0.5\%$ & $+20.0$ $[+0.0,+40.0]$ & $+0.1$ \\
 &  & Hidden complete; visible active & 2 & 3 & $0.2\%$ & $-50.0$ $[-100.0,+0.0]$ & $-0.1$ \\
 &  & Both complete & 498 & 1{,}487 & $99.1\%$ & $+0.1$ $[-1.3,+1.5]$ & $+0.1$ \\
\bottomrule
\end{tabular}
\end{adjustbox}
\end{table}

\clearpage
\begin{table}[p]
\centering
\caption{\textbf{Qwen closure and source-context effects remain small and mixed, whereas gpt-oss source-context estimates are positive at $B=256$ and $512$.} Panel a compares three forced-answer closures for the same Qwen prompt-minus-surprise contrast. Panel a reports item-bootstrap intervals, but the number of contributing items is unavailable. Panel b reconstructs each source prompt at the branch point. Qwen rows are numeric/concision prompt minus surprise stop; gpt-oss rows are lower source effort minus high source effort. Deltas are percentage points with paired item-bootstrap intervals. The Qwen results cover all four horizons, whereas the gpt-oss results cover $B=256$ and $512$.}
\label{tab:extended-context-closure}
\footnotesize
\textbf{a} Forced-answer closure wording for Qwen3-14B\\[2pt]
\begin{adjustbox}{max width=\textwidth}
\begin{tabular}{L{2.0cm} r L{3.2cm} L{3.2cm} L{3.2cm} L{2.8cm}}
\toprule
Benchmark & $B$ & Limited-time $\Delta$ [95\% CI] & Paraphrase $\Delta$ [95\% CI] & Neutral $\Delta$ [95\% CI] & Neutral minus limited-time \\
\midrule
GPQA Diamond & 512 & $-1.3$ $[-4.5,+1.7]$ & $-1.7$ $[-4.7,+1.3]$ & $-0.8$ $[-4.9,+3.0]$ & $+0.5$ $[-4.0,+5.1]$ \\
GPQA Diamond & 2{,}048 & $-1.3$ $[-4.5,+1.7]$ & $-1.0$ $[-4.0,+2.0]$ & $-0.8$ $[-4.5,+2.9]$ & $+0.5$ $[-1.7,+2.7]$ \\
GPQA Diamond & 8{,}192 & $-2.4$ $[-5.4,+0.7]$ & $-1.9$ $[-4.7,+1.2]$ & $-2.9$ $[-5.9,+0.2]$ & $-0.5$ $[-2.2,+1.2]$ \\
GPQA Diamond & 16{,}384 & $+1.6$ $[-1.8,+5.2]$ & $+2.1$ $[-1.3,+5.7]$ & $+1.8$ $[-1.7,+5.4]$ & $+0.2$ $[+0.0,+0.5]$ \\
\addlinespace
MMLU-Pro & 512 & $+1.7$ $[+0.0,+3.5]$ & $+2.4$ $[+0.7,+4.1]$ & $+1.9$ $[+0.1,+3.7]$ & $+0.1$ $[-0.8,+1.1]$ \\
MMLU-Pro & 2{,}048 & $+0.9$ $[-0.7,+2.4]$ & $+0.8$ $[-0.8,+2.4]$ & $+1.0$ $[-0.5,+2.5]$ & $+0.1$ $[-0.3,+0.6]$ \\
MMLU-Pro & 8{,}192 & $-0.2$ $[-1.5,+1.2]$ & $-0.1$ $[-1.5,+1.3]$ & $-0.2$ $[-1.5,+1.2]$ & $+0.0$ $[-0.2,+0.2]$ \\
MMLU-Pro & 16{,}384 & $-0.7$ $[-2.2,+0.6]$ & $-0.7$ $[-2.1,+0.7]$ & $-0.6$ $[-2.1,+0.7]$ & $+0.1$ $[+0.0,+0.3]$ \\
\bottomrule
\end{tabular}
\end{adjustbox}

\vspace{6pt}
\textbf{b} Source-context branch readout\\[2pt]
\begin{adjustbox}{max width=\textwidth}
\begin{tabular}{L{1.8cm} L{1.8cm} r L{2.2cm} r r r L{2.7cm}}
\toprule
Model & Benchmark & $B$ & Contrast & $n$ & Arm A acc. & Arm B acc. & $\Delta$ [95\% CI] \\
\midrule
Qwen3-14B & GPQA Diamond & 512 & prompt--surprise & 198 & $40.4\%$ & $44.9\%$ & $-4.5$ $[-9.3,0.0]$ \\
Qwen3-14B & GPQA Diamond & 2{,}048 & prompt--surprise & 198 & $53.4\%$ & $55.6\%$ & $-2.2$ $[-5.7,+1.3]$ \\
Qwen3-14B & GPQA Diamond & 8{,}192 & prompt--surprise & 198 & $59.1\%$ & $61.8\%$ & $-2.7$ $[-5.7,+0.5]$ \\
Qwen3-14B & GPQA Diamond & 16{,}384 & prompt--surprise & 198 & $64.1\%$ & $62.3\%$ & $+1.9$ $[-1.5,+5.4]$ \\
Qwen3-14B & MMLU-Pro & 512 & prompt--surprise & 500 & $60.7\%$ & $58.7\%$ & $+1.9$ $[+0.1,+3.8]$ \\
Qwen3-14B & MMLU-Pro & 2{,}048 & prompt--surprise & 500 & $72.0\%$ & $70.9\%$ & $+1.1$ $[-0.5,+2.6]$ \\
Qwen3-14B & MMLU-Pro & 8{,}192 & prompt--surprise & 500 & $75.9\%$ & $76.1\%$ & $-0.2$ $[-1.5,+1.1]$ \\
Qwen3-14B & MMLU-Pro & 16{,}384 & prompt--surprise & 500 & $75.9\%$ & $76.5\%$ & $-0.6$ $[-2.1,+0.7]$ \\
\addlinespace
gpt-oss-20b & GPQA Diamond & 256 & low--high & 196 & $53.4\%$ & $39.6\%$ & $+13.8$ $[+6.1,+21.4]$ \\
gpt-oss-20b & GPQA Diamond & 512 & low--high & 196 & $57.0\%$ & $44.9\%$ & $+12.1$ $[+4.4,+19.7]$ \\
gpt-oss-20b & GPQA Diamond & 256 & medium--high & 196 & $42.5\%$ & $39.6\%$ & $+2.9$ $[-2.4,+8.2]$ \\
gpt-oss-20b & GPQA Diamond & 512 & medium--high & 196 & $50.0\%$ & $44.9\%$ & $+5.1$ $[-0.9,+11.2]$ \\
gpt-oss-20b & MMLU-Pro & 256 & low--high & 500 & $62.1\%$ & $49.6\%$ & $+12.5$ $[+9.1,+16.1]$ \\
gpt-oss-20b & MMLU-Pro & 512 & low--high & 500 & $63.7\%$ & $55.6\%$ & $+8.1$ $[+4.6,+11.7]$ \\
gpt-oss-20b & MMLU-Pro & 256 & medium--high & 500 & $56.1\%$ & $49.6\%$ & $+6.5$ $[+4.0,+9.0]$ \\
gpt-oss-20b & MMLU-Pro & 512 & medium--high & 500 & $62.9\%$ & $55.6\%$ & $+7.3$ $[+4.7,+9.9]$ \\
\addlinespace
gpt-oss-120b & GPQA Diamond & 256 & low--high & 198 & $51.0\%$ & $40.6\%$ & $+10.4$ $[+3.0,+17.8]$ \\
gpt-oss-120b & GPQA Diamond & 512 & low--high & 198 & $61.1\%$ & $42.9\%$ & $+18.2$ $[+10.9,+25.4]$ \\
gpt-oss-120b & GPQA Diamond & 256 & medium--high & 198 & $43.8\%$ & $40.6\%$ & $+3.2$ $[-1.2,+7.6]$ \\
gpt-oss-120b & GPQA Diamond & 512 & medium--high & 198 & $53.0\%$ & $42.9\%$ & $+10.1$ $[+4.5,+15.7]$ \\
gpt-oss-120b & MMLU-Pro & 256 & low--high & 499 & $69.9\%$ & $58.1\%$ & $+11.8$ $[+7.9,+15.8]$ \\
gpt-oss-120b & MMLU-Pro & 512 & low--high & 499 & $72.5\%$ & $65.3\%$ & $+7.1$ $[+3.3,+10.8]$ \\
gpt-oss-120b & MMLU-Pro & 256 & medium--high & 500 & $65.9\%$ & $58.1\%$ & $+7.8$ $[+5.3,+10.4]$ \\
gpt-oss-120b & MMLU-Pro & 512 & medium--high & 500 & $72.0\%$ & $65.3\%$ & $+6.7$ $[+3.9,+9.4]$ \\
\bottomrule
\end{tabular}
\end{adjustbox}
\end{table}

\clearpage
\begin{table}[p]
\centering
\caption{\textbf{Candidate-logit and generated-answer readouts both show higher accuracy for completed lower-effort reasoning, with complete or near-complete coverage.} Each row uses the primary token-exact matched gpt-oss prefixes. Panel a gives absolute candidate-logit arm means under the common medium-effort answer context; every pair eligible under the matched-horizon rule is replayable. Panel b gives absolute greedy-continuation arm means after averaging the three observed replicates within item. Complete/eligible reports pairs with both continuations relative to runnable pairs, and parse coverage requires both answers to parse.}
\label{tab:extended-gptoss-absolute}
\small
\textbf{a} Candidate-logit readout, all replayable pairs\\[2pt]
\begin{adjustbox}{max width=\textwidth}
\begin{tabular}{L{2.0cm} L{1.9cm} l r r r r}
\toprule
Model & Benchmark & Terminal effort & Terminal acc. & High-prefix acc. & Replayable / eligible & Nonreplayable \\
\midrule
gpt-oss-20b & GPQA Diamond & low & $55.9\%$ & $41.4\%$ & 594/594 & 0 \\
gpt-oss-20b & GPQA Diamond & medium & $64.9\%$ & $47.5\%$ & 593/593 & 0 \\
gpt-oss-20b & MMLU-Pro & low & $64.2\%$ & $43.3\%$ & 1{,}500/1{,}500 & 0 \\
gpt-oss-20b & MMLU-Pro & medium & $72.4\%$ & $54.4\%$ & 1{,}500/1{,}500 & 0 \\
\addlinespace
gpt-oss-120b & GPQA Diamond & low & $65.0\%$ & $41.4\%$ & 594/594 & 0 \\
gpt-oss-120b & GPQA Diamond & medium & $68.5\%$ & $44.9\%$ & 594/594 & 0 \\
gpt-oss-120b & MMLU-Pro & low & $75.5\%$ & $49.2\%$ & 1{,}500/1{,}500 & 0 \\
gpt-oss-120b & MMLU-Pro & medium & $79.1\%$ & $54.9\%$ & 1{,}500/1{,}500 & 0 \\
\bottomrule
\end{tabular}
\end{adjustbox}

\vspace{6pt}
\textbf{b} Greedy generated-answer continuation\\[2pt]
\begin{adjustbox}{max width=\textwidth}
\begin{tabular}{L{2.0cm} L{1.9cm} l r r r r}
\toprule
Model & Benchmark & Terminal effort & Terminal acc. & High-prefix acc. & Complete / eligible & Parse coverage \\
\midrule
gpt-oss-20b & GPQA Diamond & low & $55.9\%$ & $40.4\%$ & 594/594 & $99.7\%$ \\
gpt-oss-20b & GPQA Diamond & medium & $64.7\%$ & $47.3\%$ & 593/593 & $98.8\%$ \\
gpt-oss-20b & MMLU-Pro & low & $64.1\%$ & $43.7\%$ & 1{,}500/1{,}500 & $99.3\%$ \\
gpt-oss-20b & MMLU-Pro & medium & $72.3\%$ & $54.1\%$ & 1{,}500/1{,}500 & $99.1\%$ \\
\addlinespace
gpt-oss-120b & GPQA Diamond & low & $65.0\%$ & $41.9\%$ & 594/594 & $99.8\%$ \\
gpt-oss-120b & GPQA Diamond & medium & $68.5\%$ & $45.5\%$ & 594/594 & $100.0\%$ \\
gpt-oss-120b & MMLU-Pro & low & $75.5\%$ & $49.5\%$ & 1{,}500/1{,}500 & $99.7\%$ \\
gpt-oss-120b & MMLU-Pro & medium & $79.1\%$ & $55.0\%$ & 1{,}500/1{,}500 & $99.5\%$ \\
\bottomrule
\end{tabular}
\end{adjustbox}
\end{table}

\clearpage
\begin{table}[p]
\centering
\caption{\textbf{Correct-option probabilities are markedly lower among incorrect completed lower-effort predictions than among incorrect matched-horizon high-effort predictions.} Rows condition separately within each arm on an incorrect candidate-readout answer among all replayable token-exact pairs. Mean conditional log loss and the share with option-normalized $q_{\mathrm{gold}}<10^{-4}$ describe the lower tail of correct-option probability. Counts are arm-specific incorrect item--replicate probes and need not match across arms. Means first average observed replicates within item and then average items.}
\label{tab:extended-gptoss-score-tails}
\small
\begin{adjustbox}{max width=\textwidth}
\begin{tabular}{L{2.0cm} L{1.9cm} l l r r r}
\toprule
Model & Benchmark & Effort & Arm & Incorrect probes & Mean log loss & $q_{\mathrm{gold}}<10^{-4}$ \\
\midrule
gpt-oss-20b & GPQA Diamond & low & Terminal & 262 & $12.745$ & $80.1\%$ \\
 &  &  & High prefix & 348 & $1.936$ & $0.3\%$ \\
gpt-oss-20b & GPQA Diamond & medium & Terminal & 208 & $16.558$ & $98.4\%$ \\
 &  &  & High prefix & 312 & $2.952$ & $3.8\%$ \\
gpt-oss-20b & MMLU-Pro & low & Terminal & 537 & $14.657$ & $88.9\%$ \\
 &  &  & High prefix & 851 & $2.862$ & $0.8\%$ \\
gpt-oss-20b & MMLU-Pro & medium & Terminal & 414 & $18.315$ & $99.1\%$ \\
 &  &  & High prefix & 684 & $4.460$ & $11.1\%$ \\
\addlinespace
gpt-oss-120b & GPQA Diamond & low & Terminal & 208 & $25.289$ & $99.6\%$ \\
 &  &  & High prefix & 348 & $5.233$ & $12.4\%$ \\
gpt-oss-120b & GPQA Diamond & medium & Terminal & 187 & $26.860$ & $100.0\%$ \\
 &  &  & High prefix & 327 & $5.502$ & $14.5\%$ \\
gpt-oss-120b & MMLU-Pro & low & Terminal & 368 & $25.343$ & $99.4\%$ \\
 &  &  & High prefix & 762 & $5.486$ & $15.4\%$ \\
gpt-oss-120b & MMLU-Pro & medium & Terminal & 314 & $26.839$ & $100.0\%$ \\
 &  &  & High prefix & 677 & $6.658$ & $24.8\%$ \\
\bottomrule
\end{tabular}
\end{adjustbox}
\end{table}

\clearpage
\begin{table}[p]
\centering
\caption{\textbf{In single-rollout Qwen3-8B and Omni-MATH-2 comparisons, Qwen3-8B shows no broad prompt gain and every gpt-oss fixed-checkpoint difference is positive.} Panel a compares one Qwen3-8B numeric/concision-prompt rollout with one separately generated surprise-stop rollout; the accuracy denominator is the number of unique paired items. For MMLU-Pro at $B=2{,}048$, the length medians use 499 paired item--trace records although the accuracy contrast contains 498 items. Panel b starts from a difficulty-stratified 500-item Omni-MATH-2 draw; the analyzed model-specific paired cohorts contain 475 items for gpt-oss-20b and 490 for gpt-oss-120b. The available data do not record why the remaining 25 and 10 sampled items, respectively, are absent. Panel b uses one rollout per effort setting, with GPT-5 mini judging free-form answers at common source-token checkpoints. Intervals bootstrap items. These estimates characterize the sampled trajectories and do not quantify variation across generated replicates.}
\label{tab:extended-scope}
\small
\textbf{a} Qwen3-8B single-rollout comparison\\[2pt]
\begin{adjustbox}{max width=\textwidth}
\begin{tabular}{L{2.0cm} r r r r L{2.6cm} r r r}
\toprule
Benchmark & $B$ & $n$ & Prompt acc. & Surprise acc. & $\Delta$ [95\% CI] & Sign-flip $p$ & Median prompt $L$ & Median surprise $L$ \\
\midrule
GPQA Diamond & 512 & 196 & $34.2\%$ & $36.7\%$ & $-2.6$ $[-6.1,+1.0]$ & $0.269$ & 5{,}022 & 5{,}929 \\
GPQA Diamond & 2{,}048 & 198 & $43.4\%$ & $42.9\%$ & $+0.5$ $[-4.5,+5.6]$ & $1.000$ & 5{,}096 & 5{,}957 \\
MMLU-Pro & 512 & 498 & $53.4\%$ & $52.6\%$ & $+0.8$ $[-1.8,+3.2]$ & $0.638$ & 1{,}286 & 1{,}623 \\
MMLU-Pro & 2{,}048 & 498 & $67.7\%$ & $65.1\%$ & $+2.6$ $[-0.2,+5.2]$ & $0.090$ & 1{,}247 & 1{,}639 \\
\bottomrule
\end{tabular}
\end{adjustbox}

\vspace{6pt}
\textbf{b} gpt-oss fixed-checkpoint Omni-MATH-2 readout\\[2pt]
\begin{adjustbox}{max width=\textwidth}
\setlength{\tabcolsep}{3pt}
\begin{tabular}{L{1.8cm} L{1.8cm} r r r r L{3.4cm}}
\toprule
Model & Contrast & $B$ & $n$ & Lower acc. & High acc. & $\Delta$ [95\% CI] \\
\midrule
gpt-oss-20b & low--high & 512 & 475 & $29.9\%$ & $18.3\%$ & $+11.6$ $[+8.4,+14.9]$ \\
gpt-oss-20b & low--high & 1{,}024 & 475 & $38.1\%$ & $21.7\%$ & $+16.4$ $[+12.6,+20.2]$ \\
gpt-oss-20b & low--high & 2{,}048 & 475 & $45.9\%$ & $29.1\%$ & $+16.8$ $[+12.6,+20.8]$ \\
gpt-oss-20b & low--high & 4{,}096 & 475 & $48.2\%$ & $34.9\%$ & $+13.3$ $[+9.3,+17.3]$ \\
gpt-oss-20b & medium--high & 512 & 475 & $21.7\%$ & $18.3\%$ & $+3.4$ $[+0.8,+6.1]$ \\
gpt-oss-20b & medium--high & 1{,}024 & 475 & $25.9\%$ & $21.7\%$ & $+4.2$ $[+1.3,+7.4]$ \\
gpt-oss-20b & medium--high & 2{,}048 & 475 & $38.7\%$ & $29.1\%$ & $+9.7$ $[+5.9,+13.5]$ \\
gpt-oss-20b & medium--high & 4{,}096 & 475 & $46.1\%$ & $34.9\%$ & $+11.2$ $[+7.4,+14.9]$ \\
\addlinespace
gpt-oss-120b & low--high & 512 & 490 & $39.0\%$ & $26.9\%$ & $+12.0$ $[+8.2,+15.9]$ \\
gpt-oss-120b & low--high & 1{,}024 & 490 & $50.4\%$ & $30.6\%$ & $+19.8$ $[+15.9,+23.9]$ \\
gpt-oss-120b & low--high & 2{,}048 & 490 & $59.0\%$ & $41.0\%$ & $+18.0$ $[+13.7,+22.2]$ \\
gpt-oss-120b & low--high & 4{,}096 & 490 & $60.6\%$ & $52.7\%$ & $+8.0$ $[+3.9,+12.0]$ \\
gpt-oss-120b & medium--high & 512 & 490 & $30.0\%$ & $26.9\%$ & $+3.1$ $[+0.0,+6.1]$ \\
gpt-oss-120b & medium--high & 1{,}024 & 490 & $39.6\%$ & $30.6\%$ & $+9.0$ $[+5.5,+12.4]$ \\
gpt-oss-120b & medium--high & 2{,}048 & 490 & $53.9\%$ & $41.0\%$ & $+12.9$ $[+9.2,+16.5]$ \\
gpt-oss-120b & medium--high & 4{,}096 & 490 & $65.5\%$ & $52.7\%$ & $+12.9$ $[+9.0,+16.7]$ \\
\bottomrule
\end{tabular}
\end{adjustbox}
\end{table}

\clearpage
\begin{table}[p]
\centering
\caption{\textbf{The Qwen MMLU-Pro $B=512$ contrast varies by category, and a text-reconstructed four-horizon concise/early-answer grid shows no broad advantage.} Panel a pools the primary 500-item subset and an independent 500-item subset at $B=512$. Accuracies average three rollout-level candidate-logit readouts within item and then average items within category; no category-level intervals are available. In panel a, prompt denotes the numeric/concision prompt and $\Delta$ is prompt minus surprise stop. Panel b compares the concise/early-answer prompt with independently generated surprise-stop and numeric/concision-prompt cohorts; its deltas are concise/early-answer minus the named comparator. Both panels use cohorts without stored completion token IDs or verified generation seeds. Panel a describes category variation across two subsets and panel b extends the prompt comparison across four horizons, but neither contributes to the token-exact primary estimates. Panel b intervals average three nominal runs within item and bootstrap items. Deltas are reported in percentage points.}
\label{tab:extended-qwen-scope-detail}
\small
\textbf{a} MMLU-Pro category decomposition at $B=512$\\[2pt]
\begin{adjustbox}{max width=\textwidth}
\begin{tabular}{L{3.0cm} r r r r}
\toprule
Category & $n$ & Surprise acc. & Prompt acc. & $\Delta$ (pp) \\
\midrule
Biology & 60 & $77.8\%$ & $80.0\%$ & $+2.2$ \\
Business & 66 & $60.1\%$ & $63.1\%$ & $+3.0$ \\
Chemistry & 94 & $51.8\%$ & $57.4\%$ & $+5.7$ \\
Computer science & 34 & $64.7\%$ & $72.5\%$ & $+7.8$ \\
Economics & 70 & $72.4\%$ & $76.7\%$ & $+4.3$ \\
Engineering & 80 & $43.8\%$ & $42.9\%$ & $-0.8$ \\
Health & 68 & $68.6\%$ & $66.7\%$ & $-2.0$ \\
History & 32 & $69.8\%$ & $67.7\%$ & $-2.1$ \\
Law & 92 & $31.5\%$ & $37.0\%$ & $+5.4$ \\
Math & 112 & $58.9\%$ & $60.4\%$ & $+1.5$ \\
Other & 76 & $54.4\%$ & $53.9\%$ & $-0.4$ \\
Philosophy & 42 & $61.1\%$ & $54.0\%$ & $-7.1$ \\
Physics & 108 & $50.9\%$ & $55.2\%$ & $+4.3$ \\
Psychology & 66 & $72.2\%$ & $74.7\%$ & $+2.5$ \\
\addlinespace
Pooled total & 1{,}000 & $57.6\%$ & $59.8\%$ & $+2.1$ \\
\bottomrule
\end{tabular}
\end{adjustbox}

\vspace{6pt}
\textbf{b} Four-horizon concise/early-answer prompt grid\\[2pt]
\begin{adjustbox}{max width=\textwidth}
\begin{tabular}{L{1.8cm} r r r L{2.2cm} r L{2.2cm}}
\toprule
Benchmark & $B$ & Concise acc. & Surprise acc. & $\Delta$ vs.\ surprise [95\% CI] & Numeric-prompt acc. & $\Delta$ vs.\ numeric prompt [95\% CI] \\
\midrule
GPQA Diamond & 512 & $43.1\%$ & $42.9\%$ & $+0.2$ $[-4.0,+4.4]$ & $43.9\%$ & $-0.8$ $[-3.5,+1.9]$ \\
GPQA Diamond & 2{,}048 & $56.1\%$ & $54.9\%$ & $+1.2$ $[-2.5,+4.7]$ & $53.0\%$ & $+3.0$ $[+0.0,+6.2]$ \\
GPQA Diamond & 8{,}192 & $59.8\%$ & $61.8\%$ & $-2.0$ $[-5.4,+1.3]$ & $58.9\%$ & $+0.8$ $[-1.9,+3.5]$ \\
GPQA Diamond & 16{,}384 & $60.8\%$ & $62.3\%$ & $-1.5$ $[-4.9,+1.7]$ & $64.1\%$ & $-3.4$ $[-6.7,-0.2]$ \\
\addlinespace
MMLU-Pro & 512 & $63.5\%$ & $58.7\%$ & $+4.9$ $[+2.8,+7.0]$ & $60.9\%$ & $+2.7$ $[+0.9,+4.5]$ \\
MMLU-Pro & 2{,}048 & $71.8\%$ & $70.8\%$ & $+1.0$ $[-0.7,+2.7]$ & $71.9\%$ & $-0.1$ $[-1.7,+1.5]$ \\
MMLU-Pro & 8{,}192 & $74.1\%$ & $76.1\%$ & $-2.1$ $[-3.5,-0.7]$ & $75.9\%$ & $-1.9$ $[-3.3,-0.4]$ \\
MMLU-Pro & 16{,}384 & $73.9\%$ & $76.5\%$ & $-2.7$ $[-4.1,-1.3]$ & $75.9\%$ & $-2.0$ $[-3.4,-0.6]$ \\
\bottomrule
\end{tabular}
\end{adjustbox}
\end{table}

\clearpage
\begin{table}[t]
\centering
\caption{\textbf{An independent MMLU-Pro subset preserves the small numeric-prompt gain at 512 tokens but not at 2,048 tokens.} The original-subset row uses the 500-item MMLU-Pro sample corresponding to the design in \Cref{tab:reviewer-qwen-primary}; it remains a separate replication rather than part of the token-exact primary analysis. The independent row uses a second category-stratified 500-item subset that excludes the original items, and the pooled row combines both subsets. At $B=2{,}048$, the numeric/concision and concise/early-answer instructions are compared with the same independent-subset surprise-stop baseline under the same three-run procedure. Accuracy and survival use the candidate-logit readout over three nominal Qwen3-14B runs. Intervals are paired two-way item $\times$ nominal-run bootstrap intervals. Seeds and completion token IDs are unavailable for these cohorts, so the intervals do not support generalization across generation seeds. Only aggregate estimates and intervals are available for the two $B=2{,}048$ rows, so they cannot be regenerated independently and are treated as descriptive evidence.}
\label{tab:qwen-mmlu-b512-independent-subset}
\small
\begin{adjustbox}{max width=\textwidth}
\begin{tabular}{L{3.2cm} r r r r L{2.2cm} r r r r}
\toprule
Sample and instruction & $B$ & $n$ & Prompt acc. & Surprise acc. & $\Delta$ [95\% CI] & $P(L>B)$ prompt & $P(L>B)$ surprise & Median prompt $L$ & Median surprise $L$ \\
\midrule
Original 500, numeric/concision & $512$ & $500$ & $60.4\%$ & $58.7\%$ & $+1.7$ $[-0.6,+4.0]$ & $75.9\%$ & $81.3\%$ & $876$ & $1{,}184$ \\
Independent 500, numeric/concision & $512$ & $500$ & $59.1\%$ & $56.6\%$ & $+2.5$ $[+0.4,+4.8]$ & $77.1\%$ & $81.5\%$ & $1{,}048$ & $1{,}314$ \\
Pooled 1,000, numeric/concision & $512$ & $1{,}000$ & $59.8\%$ & $57.6\%$ & $+2.1$ $[+0.6,+3.6]$ & $76.5\%$ & $81.4\%$ & $955$ & $1{,}214$ \\
\addlinespace
Independent 500, numeric/concision & $2{,}048$ & $500$ & $71.5\%$ & $71.7\%$ & $-0.2$ $[-2.6,+2.2]$ & $26.3\%$ & $36.0\%$ & $1{,}043$ & $1{,}314$ \\
Independent 500, concise/early-answer & $2{,}048$ & $500$ & $74.4\%$ & $71.7\%$ & $+2.7$ $[+0.5,+5.0]$ & $19.9\%$ & $36.0\%$ & $812$ & $1{,}314$ \\
\bottomrule
\end{tabular}
\end{adjustbox}
\end{table}

\clearpage
\begin{table}[t]
\centering
\caption{\textbf{Qwen prompt controls change reasoning length more consistently than accuracy at the 512-token stop.} Each row compares a Qwen3-14B instruction with the same surprise-stop baseline at $B=512$, using three nominal runs averaged within item. The accuracy columns report terminal-aware policy accuracy, with candidate-logit scoring for unfinished prefixes, and the displayed median is the full reasoning length under the instruction. Completion token IDs and verified generation seeds are unavailable for these cohorts, which therefore characterize prompt patterns. The token-exact paired experiment in \Cref{tab:qwen-mechanism-audited} supplies the primary estimate. Accuracy differences are instruction minus surprise stop in percentage points, with paired item-cluster bootstrap intervals. The four-horizon concise/early-answer results and direct prompt-to-prompt contrasts appear in \Cref{tab:extended-qwen-scope-detail}. Full survival, survivor-accuracy, terminal-aware accuracy, and $\Delta\log L$ results are included in the companion data release.}
\label{tab:qwen-prompt-controls}
\small
\begin{adjustbox}{max width=\textwidth}
\begin{tabular}{L{3.5cm} L{1.8cm} r r r r L{2.2cm} r}
\toprule
Instruction & Benchmark & $n$ & Instruction acc. & Surprise acc. & $\Delta$ (pp) & 95\% CI & Median instruction $L$ \\
\midrule
Numeric/concision & GPQA Diamond & $198$ & $43.9\%$ & $42.9\%$ & $+1.0$ & $[-2.5,+4.5]$ & $3{,}641$ \\
Limited, no number & GPQA Diamond & $198$ & $45.3\%$ & $42.9\%$ & $+2.4$ & $[-1.7,+6.4]$ & $3{,}654$ \\
Number only & GPQA Diamond & $198$ & $43.8\%$ & $42.9\%$ & $+0.8$ & $[-3.2,+4.9]$ & $3{,}690$ \\
Advance exact-stop notice & GPQA Diamond & $198$ & $43.6\%$ & $42.9\%$ & $+0.7$ & $[-3.9,+5.1]$ & $3{,}690$ \\
Approximate budget & GPQA Diamond & $198$ & $41.4\%$ & $42.9\%$ & $-1.5$ & $[-5.4,+2.2]$ & $3{,}904$ \\
Irrelevant number & GPQA Diamond & $198$ & $42.4\%$ & $42.9\%$ & $-0.5$ & $[-4.7,+3.5]$ & $4{,}222$ \\
Concise/early-answer & GPQA Diamond & $198$ & $43.1\%$ & $42.9\%$ & $+0.2$ & $[-4.0,+4.4]$ & $3{,}080$ \\
Thorough semantic & GPQA Diamond & $198$ & $42.6\%$ & $42.9\%$ & $-0.3$ & $[-4.4,+3.7]$ & $4{,}847$ \\
\addlinespace
Numeric/concision & MMLU-Pro & $500$ & $60.9\%$ & $58.7\%$ & $+2.2$ & $[+0.3,+4.1]$ & $876$ \\
Limited, no number & MMLU-Pro & $500$ & $61.0\%$ & $58.7\%$ & $+2.3$ & $[+0.5,+4.1]$ & $892$ \\
Number only & MMLU-Pro & $500$ & $60.9\%$ & $58.7\%$ & $+2.3$ & $[+0.5,+4.0]$ & $912$ \\
Advance exact-stop notice & MMLU-Pro & $500$ & $59.9\%$ & $58.7\%$ & $+1.3$ & $[-0.5,+3.0]$ & $880$ \\
Approximate budget & MMLU-Pro & $500$ & $60.9\%$ & $58.7\%$ & $+2.2$ & $[+0.5,+4.0]$ & $938$ \\
Irrelevant number & MMLU-Pro & $500$ & $59.9\%$ & $58.7\%$ & $+1.2$ & $[-0.5,+3.0]$ & $1{,}065$ \\
Concise/early-answer & MMLU-Pro & $500$ & $63.5\%$ & $58.7\%$ & $+4.9$ & $[+2.8,+7.0]$ & $710$ \\
Thorough semantic & MMLU-Pro & $500$ & $59.7\%$ & $58.7\%$ & $+1.0$ & $[-0.7,+2.7]$ & $1{,}299$ \\
\bottomrule
\end{tabular}
\end{adjustbox}
\end{table}

\clearpage
\begin{table}[t]
\centering
\caption{\textbf{Continuing an incorrect Qwen prefix can rescue its answer, but anchoring remains substantial.} Each cell reports the rescue rate and anchoring rate in percent after replaying the first $B$ tokens of an incorrect Qwen3-14B natural trace. The analysis uses the returned-top-$20$ forced-answer readout rather than full-vocabulary candidate-logit scoring. ``Answer now'' closes reasoning immediately. The continuation conditions request $R=32{,}768$ additional reasoning tokens before the forced answer, with the remaining budget either hidden or stated in the prompt. The effective maximum is limited by the context window in every displayed continuation cell, with medians from approximately $24{,}100$ to $31{,}900$ tokens.}
\label{tab:rescue}
\small
\begin{adjustbox}{max width=\textwidth}
\begin{tabular}{L{2.5cm} r r r r r}
\toprule
Benchmark & $B$ & $n$ & \multicolumn{3}{c}{Rescue / anchoring rate (\%)} \\
\cmidrule(lr){4-6}
 & & & Answer now & Continue, hidden $R$ & Continue, visible $R$ \\
\midrule
GPQA Diamond & $512$ & $123$ & $30.1 / 40.7$ & $38.2 / 37.4$ & $37.4 / 35.8$ \\
GPQA Diamond & $2{,}048$ & $112$ & $24.1 / 45.5$ & $33.9 / 42.9$ & $27.7 / 50.0$ \\
GPQA Diamond & $8{,}192$ & $34$ & $20.6 / 55.9$ & $26.5 / 47.1$ & $26.5 / 47.1$ \\
\addlinespace
MMLU-Pro & $512$ & $176$ & $26.1 / 43.2$ & $31.2 / 51.1$ & $31.2 / 51.7$ \\
MMLU-Pro & $2{,}048$ & $95$ & $20.0 / 55.8$ & $29.5 / 48.4$ & $29.5 / 55.8$ \\
MMLU-Pro & $8{,}192$ & $30$ & $26.7 / 46.7$ & $33.3 / 50.0$ & $30.0 / 50.0$ \\
\bottomrule
\end{tabular}
\end{adjustbox}
\end{table}

\clearpage
\begin{table}[t]
\centering
\caption{\textbf{The prompted budget is decodable at the prompt boundary but mostly falls to label-shuffle levels after reasoning begins.} The table reports the best-layer activation readouts for Qwen3-14B at representative token positions $k$. ``Bundled prompt?'' is a binary readout over surprise-stop and numeric/concision-prompt traces. ``Which budget?'' is a four-class readout over numeric/concision-prompt traces only, with $B\in\{512,2{,}048,8{,}192,16{,}384\}$. Balanced accuracy uses item-grouped cross-validation. The shuffle column is the 95th percentile over 20 label shuffles.}
\label{tab:activation-readout}
\small
\begin{adjustbox}{max width=\textwidth}
\begin{tabular}{L{2.7cm} L{2.2cm} r r r r r}
\toprule
Readout & Benchmark & $k$ & Best layer & Items & Balanced acc. & Shuffle q95 \\
\midrule
Bundled prompt? & GPQA Diamond & $0$ & $10$ & $197$ & $1.000$ & $0.546$ \\
Bundled prompt? & GPQA Diamond & $64$ & $39$ & $197$ & $0.621$ & $0.533$ \\
Bundled prompt? & GPQA Diamond & $512$ & $39$ & $197$ & $0.700$ & $0.526$ \\
Bundled prompt? & GPQA Diamond & $2{,}048$ & $39$ & $163$ & $0.623$ & $0.539$ \\
Bundled prompt? & MMLU-Pro & $0$ & $10$ & $200$ & $1.000$ & $0.537$ \\
Bundled prompt? & MMLU-Pro & $64$ & $39$ & $200$ & $0.627$ & $0.540$ \\
Bundled prompt? & MMLU-Pro & $512$ & $39$ & $178$ & $0.689$ & $0.548$ \\
Bundled prompt? & MMLU-Pro & $2{,}048$ & $39$ & $75$ & $0.588$ & $0.580$ \\
\addlinespace
Which budget? & GPQA Diamond & $0$ & $39$ & $197$ & $0.938$ & $0.283$ \\
Which budget? & GPQA Diamond & $64$ & $10$ & $197$ & $0.258$ & $0.268$ \\
Which budget? & GPQA Diamond & $512$ & $30$ & $197$ & $0.278$ & $0.269$ \\
Which budget? & GPQA Diamond & $2{,}048$ & $39$ & $156$ & $0.262$ & $0.286$ \\
Which budget? & MMLU-Pro & $0$ & $39$ & $200$ & $0.905$ & $0.278$ \\
Which budget? & MMLU-Pro & $64$ & $0$ & $200$ & $0.261$ & $0.279$ \\
Which budget? & MMLU-Pro & $512$ & $20$ & $171$ & $0.259$ & $0.278$ \\
Which budget? & MMLU-Pro & $2{,}048$ & $0$ & $69$ & $0.343$ & $0.301$ \\
\bottomrule
\end{tabular}
\end{adjustbox}
\end{table}

\clearpage
\begin{table}[t]
\centering
\caption{\textbf{Adding numeric wording to gpt-oss effort produces no broad advantage at the 512-token stop.} Each row compares a numeric-budget instruction crossed with the indicated effort setting against the effort-only fixed-checkpoint baseline at the same $B=512$ stop. Differences are numeric-instruction minus effort-only accuracy in percentage points, with paired bootstrap intervals over items. Longer announced-budget contrasts lack a directly matched effort-only baseline because the effort-only comparison covers only $B=512$.}
\label{tab:gptoss-numeric-effort}
\small
\begin{adjustbox}{max width=\textwidth}
\begin{tabular}{L{2.0cm} L{1.8cm} L{1.4cm} r r r r L{2.0cm}}
\toprule
Model & Benchmark & Effort & $n$ & Numeric acc. & Effort-only acc. & $\Delta$ (pp) & 95\% CI \\
\midrule
gpt-oss-20b & GPQA Diamond & low & $198$ & $55.6\%$ & $56.1\%$ & $-0.5$ & $[-7.6,+6.1]$ \\
gpt-oss-20b & GPQA Diamond & medium & $198$ & $44.4\%$ & $50.0\%$ & $-5.6$ & $[-10.6,-0.5]$ \\
gpt-oss-20b & GPQA Diamond & high & $196$ & $44.4\%$ & $44.4\%$ & $+0.0$ & $[-6.1,+6.1]$ \\
gpt-oss-20b & MMLU-Pro & low & $498$ & $66.5\%$ & $63.7\%$ & $+2.9$ & $[-0.5,+6.5]$ \\
gpt-oss-20b & MMLU-Pro & medium & $500$ & $63.0\%$ & $63.0\%$ & $+0.0$ & $[-2.7,+2.7]$ \\
gpt-oss-20b & MMLU-Pro & high & $499$ & $58.9\%$ & $56.2\%$ & $+2.7$ & $[+0.0,+5.5]$ \\
\addlinespace
gpt-oss-120b & GPQA Diamond & low & $198$ & $64.1\%$ & $61.1\%$ & $+3.0$ & $[-3.5,+9.6]$ \\
gpt-oss-120b & GPQA Diamond & medium & $198$ & $53.5\%$ & $53.0\%$ & $+0.5$ & $[-4.2,+5.2]$ \\
gpt-oss-120b & GPQA Diamond & high & $198$ & $41.4\%$ & $43.9\%$ & $-2.5$ & $[-7.1,+2.0]$ \\
gpt-oss-120b & MMLU-Pro & low & $499$ & $73.7\%$ & $73.1\%$ & $+0.6$ & $[-2.2,+3.4]$ \\
gpt-oss-120b & MMLU-Pro & medium & $500$ & $73.2\%$ & $71.8\%$ & $+1.4$ & $[-0.8,+3.7]$ \\
gpt-oss-120b & MMLU-Pro & high & $500$ & $66.0\%$ & $66.7\%$ & $-0.7$ & $[-3.0,+1.6]$ \\
\bottomrule
\end{tabular}
\end{adjustbox}
\end{table}

\clearpage
\begin{table}[t]
\centering
\caption{\textbf{The scripted positive control detects a known early-answer policy.} We construct 96 balanced synthetic four-choice items whose questions contain no answer information. The long policy uses a neutral reasoning prefix whose candidate answer appears after 64 tokens, whereas the short policy states the candidate answer immediately. Both policies reveal the candidate answer by $B=512$. Accuracy uses the Qwen3-14B candidate-logit readout. Differences are short minus long in percentage points, with paired bootstrap intervals over items. The reported $p$ values use a two-sided paired item-level sign-flip randomization test with 10,000 draws.}
\label{tab:positive-control}
\small
\begin{adjustbox}{max width=\textwidth}
\begin{tabular}{r r r r r L{2.2cm} r r r}
\toprule
$B$ & $n$ & Long acc. & Short acc. & $\Delta$ (pp) & 95\% CI & Long reveal & Short reveal & Sign-flip $p$ \\
\midrule
$64$ & $96$ & $25.0\%$ & $100.0\%$ & $+75.0$ & $[+65.6,+83.3]$ & $0.0\%$ & $100.0\%$ & $0.0001$ \\
$512$ & $96$ & $100.0\%$ & $100.0\%$ & $+0.0$ & $[+0.0,+0.0]$ & $100.0\%$ & $100.0\%$ & $1.0000$ \\
\bottomrule
\end{tabular}
\end{adjustbox}
\end{table}

\clearpage
\begin{table}[t]
\centering
\caption{\textbf{None of the 2,788 terminal gpt-oss replay prefixes in a separate single-run cohort contains final-channel answer content.} This terminal-replay validation does not exhaustively cover the three-replicate primary cohort. It covers terminal low- and medium-effort prefixes in the single-run common-context readout. Answer assertions inside the reasoning text are part of the measured behavior and are not counted as leaks. A final-channel leak is a replay prefix containing a final-channel delimiter. A final-reply suffix leak is the parsed natural final reply appended to the replay prefix when the reply contains at least three non-whitespace characters.}
\label{tab:terminal-prefix-leak-audit}
\small
\begin{adjustbox}{max width=\textwidth}
\begin{tabular}{L{2.1cm} L{2.1cm} L{2.7cm} r r r r}
\toprule
Model & Benchmark & Prefix source & Prefixes & Final-channel leaks & Reply checks & Final-reply suffix leaks \\
\midrule
gpt-oss-120b & GPQA Diamond & low terminal & 198 & 0 & 0 & 0 \\
gpt-oss-120b & GPQA Diamond & medium terminal & 198 & 0 & 0 & 0 \\
gpt-oss-120b & MMLU-Pro & low terminal & 500 & 0 & 0 & 0 \\
gpt-oss-120b & MMLU-Pro & medium terminal & 500 & 0 & 0 & 0 \\
gpt-oss-20b & GPQA Diamond & low terminal & 196 & 0 & 0 & 0 \\
gpt-oss-20b & GPQA Diamond & medium terminal & 196 & 0 & 0 & 0 \\
gpt-oss-20b & MMLU-Pro & low terminal & 500 & 0 & 1 & 0 \\
gpt-oss-20b & MMLU-Pro & medium terminal & 500 & 0 & 1 & 0 \\
\addlinespace
Total & & & 2{,}788 & 0 & 2 & 0 \\
\bottomrule
\end{tabular}
\end{adjustbox}
\end{table}

\clearpage

\end{document}